%% file: main.tex
\documentclass[10pt]{article} 
\usepackage[preprint]{tmlr}

\input{math_commands.tex}

\usepackage{hyperref}
\usepackage{url}
\usepackage{graphicx}
\usepackage{booktabs}
\usepackage{multirow}
\usepackage{wrapfig}
\usepackage{float}
\usepackage{changepage}

\renewcommand{\arraystretch}{0.9}


\usepackage{subcaption}
\usepackage{caption}
\title{Adaptive Depth in Looped Transformers: \\ Diagnosing Learned Halting Gates and Trajectory Readouts
}

\author{
\\
\name Andrei Cristian Popescu \\
\addr University of Cambridge
\AND
\name Haitz Sáez de Ocáriz Borde \\
\addr University of Cambridge
\AND
\name Pietro Liò \\
\addr University of Cambridge
}

\def\month{MM}  
\def\year{YYYY} 
\def\openreview{\url{https://openreview.net/forum?id=XXXX}} 

\begin{document}

\maketitle
\vspace{-28pt}
\begin{abstract}
Looped Transformers increase test-time computation by repeatedly applying a shared recurrent block. Learned halting objectives in looped Transformers typically use a single exit distribution both as the inference-time stopping rule and as the training-time weighting of per-depth losses. This entangles exit selection with trajectory formation: the gate not only chooses which recurrent state to use, but also determines how strongly each intermediate state is supervised. Consequently, poor adaptive-compute performance can arise from the readout, the induced trajectory, or their interaction. We study adaptive depth in looped Transformers through this trajectory--readout lens, across controlled synthetic tasks (modular arithmetic and binary parity) and large-scale Ouro-1.4B and 2.6B checkpoints. We find that fixed-prior depth supervision, which shapes the trajectory without an input-dependent halting policy, produces difficulty-aware trajectories whose intermediate states expose useful stopping signals, and that simple post-hoc confidence readouts often match or outperform learned linear and MLP gates. Fitting gates on frozen trajectories localizes the failure: it appears to stem mainly from the trajectory induced by joint gate training rather than from limited gate expressivity. The same pattern is present in Ouro evaluations, where pretrained ponder gates are competitive but not uniformly Pareto-optimal, and measured latency confirms that the resulting reductions in average exit depth translate into practical inference-time savings. Our systematic diagnostic evaluation reframes adaptive depth in looped Transformers as a joint problem of trajectory formation and exit readout, rather than gate learning alone, highlighting a distinction that prior learned-halting work has often left implicit.
\end{abstract}


\begin{figure}[H]
  \centering
  \begin{adjustwidth}{-0.9in}{-0.9in}
    \centering
    \includegraphics[height=2.2in]{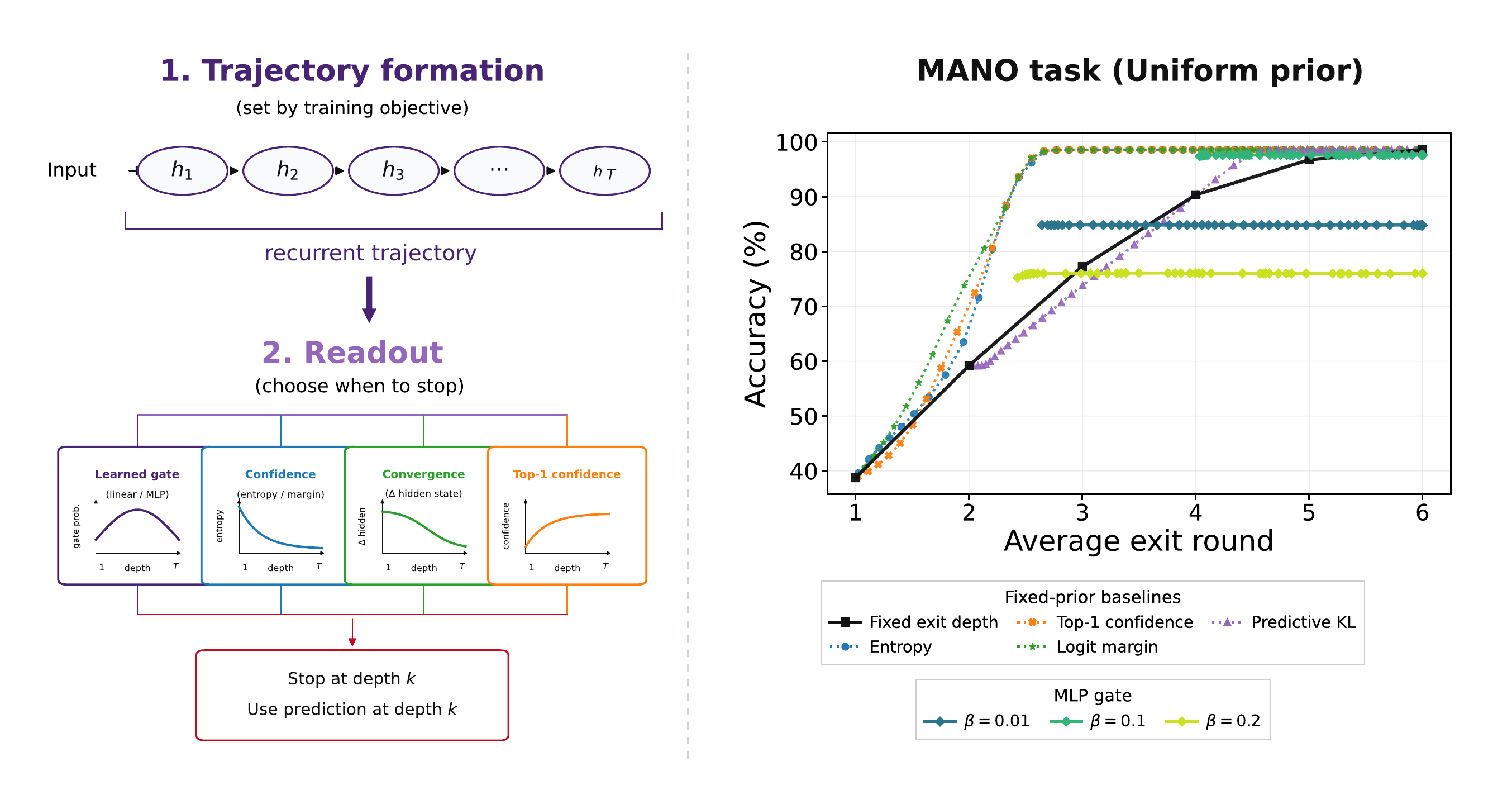}
  \end{adjustwidth}
  \caption{\small \textbf{Adaptive depth is a trajectory--readout problem, not just a gate-learning problem.} \emph{Left:} a looped Transformer's recurrent trajectory $h_1,\ldots,h_T$ is shaped by the training objective, while a separate readout (a learned gate, or a simple heuristic) decides when to stop. \emph{Right:} on MANO, post-hoc readouts on a \textbf{fixed-prior} trajectory reach near-ceiling accuracy at lower exit depth than a fixed-depth baseline, matching or outperforming an MLP gate, whose halting distribution both selects exits and shapes the trajectory it is trained on.}
  \label{fig:teaser}
\end{figure}

\vspace{-10pt}
\section{Introduction}

This paper provides a systematic diagnostic study of adaptive depth in looped Transformers. Learned halting mechanisms promise adaptive computation by allocating recurrent depth to the input and stopping once further refinement is unnecessary. However, current learned halting objectives in looped Transformers couple two roles that should be distinguished. The learned exit distribution is used as an inference-time stopping rule, but it also weights the training losses across recurrent depths. Thus, a learned gate does not merely decide when to stop; it also changes the recurrent trajectory from which exits are selected.

This coupling makes adaptive depth difficult to diagnose: poor compute--quality tradeoffs may arise from the stopping rule, the induced trajectory, or their interaction. We therefore study adaptive depth as a trajectory--readout problem, varying training objectives, learned gates, post-hoc readouts, forced exits, and evaluating both controlled models trained on MANO (a synthetic modular-arithmetic task) and parity tasks, as well as pretrained Ouro checkpoints.

Our results show that adaptive depth is not only a question of learning a better halting gate. Fixed-prior depth supervision, which shapes the trajectory without learning an input-dependent halting policy, can produce difficulty-aware recurrent trajectories whose intermediate states support strong post-hoc early exits. In several settings, simple confidence readouts match or outperform learned gates in compute--accuracy tradeoff. We do not interpret this as evidence against learned halting in principle. Rather, our results suggest that current learned-gate objectives can fail because they entangle trajectory formation and exit selection, and that future adaptive-depth methods should treat both components as first-class design choices.

Our contributions are:
(i) proposing a trajectory--readout view of adaptive depth in looped Transformers;
(ii) systematically evaluating training objectives, gate parameterizations, post-hoc readouts, forced-exit trajectories, model scales, and latency;
(iii) introducing fixed-prior depth supervision to decouple trajectory formation from exit selection;
(iv) showing that fixed-prior training produces difficulty-aware recurrent trajectories on MANO;
(v) demonstrating that simple post-hoc confidence readouts can match or outperform learned linear and MLP gates;
(vi) showing, on Ouro-1.4B and Ouro-2.6B, that pretrained ponder gates are not uniformly Pareto-optimal; and
(vii) providing latency measurements that connect reductions in average exit depth to practical inference-time speedups.

\section{Related Work}

Next, we review prior work along the two axes most relevant to our study: recurrent-depth Transformer architectures and adaptive computation methods for selecting how much computation to use.

\subsection{Looped Transformers}
Looped Transformers replace stacks of distinct layers with repeated applications of a shared block of layers, decoupling effective computational depth from parameter count. Early work in the field of looped Transformers includes the Universal Transformer \citep{UT}, implicit-depth models such as Deep Equilibrium Models \citep{bai2019deepequilibriummodels}, and cross-layer parameter sharing in ALBERT \citep{albert}. More recent work has reframed recurrence as latent computation, in which each recurrence step refines the hidden representation before performing the prediction \citep{reasoninglatentthoughtspower,Huginn,Ouro,loopformer}. This depth-wise recurrence differs from approaches that extend the sequence with continuous hidden states or latent tokens, since computation remains on the model's depth axis rather than in context \citep{coconut,zelikman2024quietstarlanguagemodelsteach,huang2025fastquietstarthinkingthought}.

Looped models have been proven to have promising performance on iterative and algorithmic computation, including programmable computation, length generalization, and multi-step refinement on synthetic tasks \citep{loopedtransformersprogrammablecomputers,anil2022exploringlengthgeneralizationlarge,yang2024loopedtransformersbetterlearning,kohli2026loopthinkgeneralize}. More recent language-modeling work studies whether this recurrent-depth scaling transfers to larger models, where the additional computation can be spent in latent space rather than through longer generated reasoning traces \citep{Huginn,Ouro,loopformer}. Our work builds on this view of looped Transformers as a means to increase test-time computation, but focuses on a different question: how training objectives shape the recurrent trajectories, and how exit readouts convert them into compute--quality tradeoffs.

\subsection{Adaptive Computation}

Adaptive computation methods aim to improve performance--compute tradeoffs by allocating different amounts of computation to different tokens. Early approaches such as ACT \citep{adaptivecomputationtimerecurrent} and PonderNet \citep{pondernet} learn probabilistic halting policies, while later work applies adaptive depth to Transformers and language models \citep{dynabertdynamicbertadaptive,depthadaptivetransformer,SwitchTS,bae-etal-2023-fast,Elhoushi_2024,mixtureofdepthsdynamicallyallocatingcompute}. This idea adapts naturally to looped Transformers. Their recurrent structure produces intermediate states, thus allowing us either to exit early and commit to one of those hidden states or to continue refining them through additional iterations. Recent looped language models therefore use learned gates or routers to allocate recurrent depth \citep{UT,Ouro,adaponderlm,MoR}.

Learned gates are not the only possible stopping signal. Early-exit feedforward Transformer methods have also used confidence, entropy, or prediction stability as inexpensive criteria for terminating computation \citep{confidentadaptivelanguagemodeling,deebert,zhou2020bertlosespatiencefast}, and entropy dynamics over chain-of-thought traces have similarly been linked to reasoning correctness \citep{sia}. In looped Transformers, analogous signals can be computed across recurrent iterations, including changes in hidden states, logits, or predictive distributions \citep{Huginn}. Our work builds on this observation, but separates two roles that are often implicitly coupled: training objectives shape the recurrent trajectory, while readouts select exits from the induced trajectory.

Prior work asks how to make recurrent depth dynamic, through learned halting policies, routers, or other input-dependent allocation rules \citep{Huginn,Ouro,adaponderlm,MoR}. We ask a complementary question: when such a method succeeds or fails, what part of the system is responsible? In looped Transformers a halting mechanism is not only a stopping rule; during training it also determines which recurrent depths receive supervision, and therefore shapes the trajectory from which exits are later selected. We isolate this distinction experimentally, separating trajectory formation from exit readout.

\section{Formalizing Adaptive Depth as Trajectory and Readout}

In this section, we introduce the notation and mechanisms used throughout the paper. We first formalize looped Transformer trajectories as the sequence of hidden states produced by repeatedly applying a shared Transformer block. We then describe learned halting gates, which both define an exit distribution at inference time and weight losses across recurrent depths during training. Finally, we introduce fixed-prior depth supervision and post-hoc trajectory readouts, which allow us to separate trajectory formation from exit selection.

\subsection{Looped Transformer Trajectories} 
Let \(x=(x_1,\ldots,x_M)\) be an input sequence of length $M$ and let 
\( h_0=\text{E}(x) \in \mathbb{R}^{M \times d} \) 
denote its token embeddings, for an embedding function E$: \mathbb{R}^{|V|} \rightarrow \mathbb{R}^d$ with vocabulary size $|V|$ and hidden size $d$. A looped Transformer repeatedly applies a shared block $\mathcal{F}_\theta$ of $L$ Transformer layers to update the hidden state, where $\theta$ represents the parameters of the model:
\( h_{t}= \mathcal{F}_\theta(h_{t-1}), \text{ for } t=1,...,T \)
where $T$ is the maximum possible number of loops. The generated hidden states can be unembedded using a layer $\text{lmhead}:\mathbb{R}^d \rightarrow \mathbb{R}^{|V|}$. Looped Transformers naturally induce a trajectory in the latent space: \( \tau_\theta(x) = \{ h_i \}_{i=1}^{T}\).

In this work, we study how this induced trajectory is shaped by the adaptive-computation training objective and whether trajectory proxies are competitive with trained gates at predicting when computation should halt. Additional architectural background and notation for looped Transformers are provided in Appendix~\ref{app:looped_architecture_details}.

\subsection{Gating Mechanism for Adaptive Computation} 
\label{sec:gating-mechanism}
A common approach to adaptive computation in looped Transformers is to attach a PonderNet inspired halting module that predicts an exit distribution over recurrent depths \citep{pondernet, Ouro, adaponderlm}. 

Let \( e_t(x)=\sigma(g_\phi(h_t)) \in (0,1) \) denote the conditional probability of exiting at loop \(t\), given that the model has not exited earlier. In this work, we experiment with the gate as both a linear projection and an MLP. The predicted conditional halt probabilities induce a distribution over exit depths:
\[ q_t(x)=e_t(x)\prod_{j<t}\left(1-e_j(x)\right)\quad \text{for }t=1,...,T-1. \qquad q_T(x)=\prod_{j<T}\left(1-e_j(x)\right). \]
Note that we assign all remaining probability mass to the final recurrent depth \(T\), ensuring that \(q(\cdot\mid x)\) is a valid distribution over exits. At inference time, the learned distribution can be used as a stopping rule, by halting computation when the cumulative exit distribution exceeds a threshold $\alpha$. We can define the early exit depth $d_\alpha$ as follows: \( d_\alpha(x) = \min\left\{ t: \sum_{s=1}^t q_s(x)\ge \alpha \right\}. \)

During training, the exit distribution is used to weight the losses at different recurrent depths. Given per depth Cross-Entropy losses \(\ell_t\), the model is jointly trained with the gate using the following objective:
\[ \mathcal{L}(\theta,\phi) = \mathbb{E}_{(x,y)} \left[ \sum_{t=1}^T q_t(x)\ell_t(x,y) + \beta KL(q(x) || \pi) \right], \]
where $\pi$ is a target prior over depths, and \(\beta\) controls the strength of the regularization. We note that since the learned weights \(q_t(x)\) appear inside the loss, the gate is not only an inference-time stopping rule. It also determines how much prediction loss each recurrent depth contributes during training. Therefore, the learned trajectory $\tau_{(\theta,\phi)}(x)=\{h_t\}_{t=1}^T$ depends on the halting objective and on the gate parameters \(\phi\). This implies that training a gate entangles two effects: learning a readout \(q(\cdot\mid x)\) for selecting exits, and changing the depth supervision that shapes the trajectory itself. Thus, different gate design choices affect how the model spreads the computation across loops. A central goal of our study is to separate these two effects.
\subsection{Fixed-Prior Depth Supervision}
\label{sec:fixed_prior}
The training objective above uses an input-dependent distribution \(q(\cdot\mid x)\) to define an inference-time readout and to weight losses across the loops during training. To decouple these roles, we also study objectives with fixed loss weights over recurrent depth. Given a fixed distribution \(\pi\) over loops, fixed-prior training minimizes the following loss function:
\[ \mathcal{L}_{\mathrm{fixed}}(\theta) = \mathbb{E}_{(x,y)} \left[ \sum_{t=1}^T \pi_t \ell_t(x,y) \right]. \]
In contrast to \(q(\cdot\mid x)\), the weights \(\pi_t\) do not depend on the input and do not define a learned stopping policy. They only specify how strongly each recurrent depth is supervised during training. Fixed-prior training provides a controlled way to shape the trajectory without simultaneously learning an adaptive readout. 

In this study, we consider two families of fixed priors: the uniform and geometric priors. The uniform prior assigns equal loss weight to every recurrent depth, \( \pi_t=\frac{1}{T}. \)
The truncated geometric prior places more weight on earlier depths. To obtain a valid probability distribution, we assign all remaining probability mass to the final recurrent depth:
\[ \pi_t=\lambda(1-\lambda)^{t-1}, \quad \text{for }t= 1,...,T-1. \qquad \pi_T=(1-\lambda)^{T-1}. \]
Since fixed-prior trained models do not learn a halting distribution, they need a separate readout at inference time. This makes them useful for our decomposition. If post-hoc trajectory readouts can stop fixed-prior trajectories effectively, then adaptive exit does not necessarily require a learned gate, but a trajectory whose intermediate states contain informative stopping signals.
\subsection{Trajectory Readouts}
\label{sec:trajectory_readout}
After training, an adaptive-compute method must choose an exit depth from the recurrent trajectory. We use the term \emph{readout} for any rule that maps the trajectory to an exit depth. A fixed-depth readout exits at the same loop \(d(x)=t_0\) for every input. A learned gate readout uses the learned exit distribution \(q(\cdot\mid x)\), for example through the CDF threshold rule \(d_\alpha(x)\) defined above.

We also study post-hoc trajectory readouts that do not require a trained gate. Confidence readouts decide to exit when the model appears sufficiently certain, using quantities such as entropy, top-1 confidence, or logit margin. Convergence readouts exit when the trajectory appears to have stabilized, using quantities such as predictive KL between consecutive predictions, logit change, or hidden-state movement. All readouts are implemented as threshold rules over these scalar signals. Formal definitions are given in Appendix~\ref{app:readout_definitions}. These readouts are applied only after training, so they can select exits from a trajectory but cannot influence how the trajectory is shaped during training. To separate trajectory quality from readout quality, we also use forced-exit evaluations. A forced exit ignores any adaptive rule and evaluates the prediction at each loop \(t\). This gives the per-depth quality curve of the trajectory itself.

\section{Methodology}
We design experiments to separate trajectory formation from exit selection. In the controlled MANO experiments, we train looped Transformers under fixed-prior and learned-gate objectives, then evaluate multiple readouts on the resulting trajectories. We use Ouro 1.4B and Ouro 2.6B to validate our claims on large-scale public foundation models. In particular, we evaluate whether the pretrained gate in these public looped language model checkpoints is competitive with simple post-hoc trajectory readouts. Additional implementation details for MANO, Ouro, threshold selection, and optimization are provided in Appendix~\ref{app:experimental_details}.

\subsection{Controlled MANO Experiments}
Our primary controlled setting is MANO \citep{allenzhu2025physicslanguagemodels41}, a synthetic modular arithmetic task over integers modulo 23. Each example is an arithmetic expression written in the prefix notation, and the model must predict the final answer. We use expressions with up to 10 operations and group examples by operation count when analyzing difficulty. Unlike setups that provide the expression length as an input token, our implementation does not include a length token, so that the model must infer difficulty from the expression itself, rather than finding a shortcut.

We focus on this setting because it exposes a nontrivial accuracy--compute tradeoff. Early recurrent states are not always sufficient, while later recurrent states remain relevant for solving harder expressions. Moreover, this dataset allows us to have a clear difficulty ordering, through the number of operations, a fact used in the analysis. All MANO models use the same looped Transformer backbone with 4 layers, 4 attention heads, hidden embedding size 512, and maximum recurrent depth \(T=6\). We evaluate models both by adaptive exits and by forced exits. In a forced-exit evaluation, the model is evaluated independently at each recurrent depth \(t\), without using any learned or post-hoc stopping rule. This gives the per-loop trajectory of accuracy and loss, and lets us analyze whether training produced useful intermediate states.

For MANO, we generate three disjoint datasets of 50k examples using different random seeds. Models are trained on a split generated with seed 42, post-hoc readout thresholds are selected on a validation set generated with seed 43, and all reported compute--quality curves are evaluated on a test set generated with seed 44. The same validation and test splits are used for all readouts, and the test split is used only after the threshold grid has been fixed.

\subsection{Training Objectives and Gate Parameterizations}

We train three model families. Fixed-prior models use the objective in Section~\ref{sec:fixed_prior} with a uniform prior or truncated geometric priors (larger \(\lambda\) places more loss weight on earlier loops). Learned linear-gate models use the objective in Section~\ref{sec:gating-mechanism} with \(g_\phi\) a linear projection to a halt logit, trained under both priors while sweeping \(\beta\). MLP-gate models replace \(g_\phi\) with a two-layer feedforward network, testing whether gate expressivity changes the comparison against simple trajectory readouts.

\subsection{Trajectory Readouts and Threshold Selection}

For each trained trajectory, we evaluate readouts that select an exit depth. Learned-gate models use their native cumulative-threshold rule \(d_\alpha\) (Section~\ref{sec:trajectory_readout}), swept over \(\alpha\) to trace a Pareto curve. We also evaluate post-hoc readouts that do not affect the trajectory: confidence readouts (entropy, top-1 confidence, logit margin) and convergence readouts (predictive KL, logit change, hidden-state displacement norm), each swept over a validation-calibrated grid. Fixed-depth baselines require no calibration. Grid and held-out details are in Section~\ref{sec:evaluation_metrics} and Appendix~\ref{app:experimental_details}.

\subsection{Evaluation Protocol and Metrics}
\label{sec:evaluation_metrics}

Our evaluation separates two questions: whether a training objective produces useful intermediate states, and whether a readout selects good exits from them. We measure trajectory quality with \emph{forced exits}, evaluating every example at each recurrent depth without any stopping rule. We measure adaptive stopping by sweeping thresholds into compute--quality curves; each threshold gives one operating point of task quality versus average exit depth. Validation data are used only for threshold selection (MANO: select on seed 43, evaluate on seed 44; Ouro: select on the validation split or subset of Section~\ref{sec:large_scale_methodology}, evaluate on the held-out split). Computation is the mean selected depth \(\mathbb{E}_x[d(x)]\), the \emph{average loops}. For MANO we summarize frontiers with D@X, the minimum average depth to reach $X\%$ test accuracy; for Ouro we report held-out accuracies at the validation-selected operating points.

\subsection{Large-Scale Looped Language-Model Evaluation}
\label{sec:large_scale_methodology}

To test whether the same readout pattern appears in large-scale looped language-model checkpoints, we evaluate Ouro-1.4B and Ouro-2.6B \citep{Ouro}, a family of pretrained Looped Language Models that build iterative latent computation into pretraining, applying a shared Transformer block for up to four recurrent steps and using a PonderNet-style gate trained in two stages (joint entropy-regularized training, then frozen-backbone fine-tuning on a marginal-utility signal). Unlike the MANO experiments, we do not train or modify these models: we evaluate the released checkpoints as-is and compare the pretrained ponder gate against simple post-hoc readouts applied to the recurrent trajectory. Full architectural and training details are given in Appendix~\ref{app:ouro-background}.

We evaluate on six standard benchmarks: MMLU \citep{hendrycks2021measuringmassivemultitasklanguage}, ARC-Easy \citep{clark2018thinksolvedquestionanswering}, ARC-Challenge \citep{clark2018thinksolvedquestionanswering}, OpenBookQA \citep{mihaylov2018suitarmorconductelectricity}, HellaSwag \citep{zellers2019hellaswagmachinereallyfinish}, and CommonsenseQA \citep{talmor2019commonsenseqaquestionansweringchallenge}. We evaluate MMLU with 5-shot prompting, ARC-Challenge with 25-shot prompting, ARC-Easy with 8-shot prompting, OpenBookQA with 0-shot prompting, and HellaSwag and CommonsenseQA with 10-shot prompting. For benchmarks with labeled test sets, thresholds are selected on the standard validation split and evaluated on the standard test split. For HellaSwag and CommonsenseQA, where labeled test sets are not available, we split the labeled validation data into a 20\% threshold-selection subset and an 80\% held-out evaluation subset using seed 42. The same held-out protocol and compute metrics from Section~\ref{sec:evaluation_metrics} are used for all Ouro readouts.

\section{Results}

The results proceed in three stages. First, we use MANO as a controlled setting where we can change the training objective and directly inspect the recurrent trajectory. Section~\ref{sec:results_fixed_prior} asks whether adaptive exits require a learned halting policy and whether fixed-prior trajectories develop difficulty-aware dynamics. Section~\ref{sec:results_fixed_vs_gate} compares fixed-prior trajectories against learned-gate trajectories, and Section~\ref{sec:training-objective-shapes-trajectories} uses forced exits to diagnose how the objective changes intermediate states. Second, Section~\ref{sec:posthoc-gate-fitting} freezes pretrained MANO trajectories and fits new post-hoc gates, separating gate expressivity from trajectory quality. Finally, Section~\ref{sec:large-scale} tests the same readout pattern in pretrained Ouro-1.4B and Ouro-2.6B checkpoints, and Section~\ref{sec:latency} measures whether reductions in average recurrent depth correspond to latency savings.

\subsection{Fixed-prior training produces difficulty-aware trajectories}
\label{sec:results_fixed_prior}

We first ask whether adaptive computation requires a learned halting distribution. Fixed-prior-trained models provide a controlled test. Their objective shapes a recurrent trajectory, but it does not learn an input-dependent exit policy.

Throughout the paper, Pareto plots report task quality (y-axis) against average exit depth (x-axis), the mean number of recurrent loops executed under a given readout. Points closer to the upper left are preferable, achieving higher accuracy with less computation. Each curve is obtained by sweeping the stopping threshold of a single readout, so it represents the full compute--quality tradeoff for that readout.

\begin{figure}[!t]
    \centering
    \begin{subfigure}{0.46\linewidth}
        \centering
        \includegraphics[width=\linewidth,trim={0.1in 0.05in 0.1in 0.08in},clip]{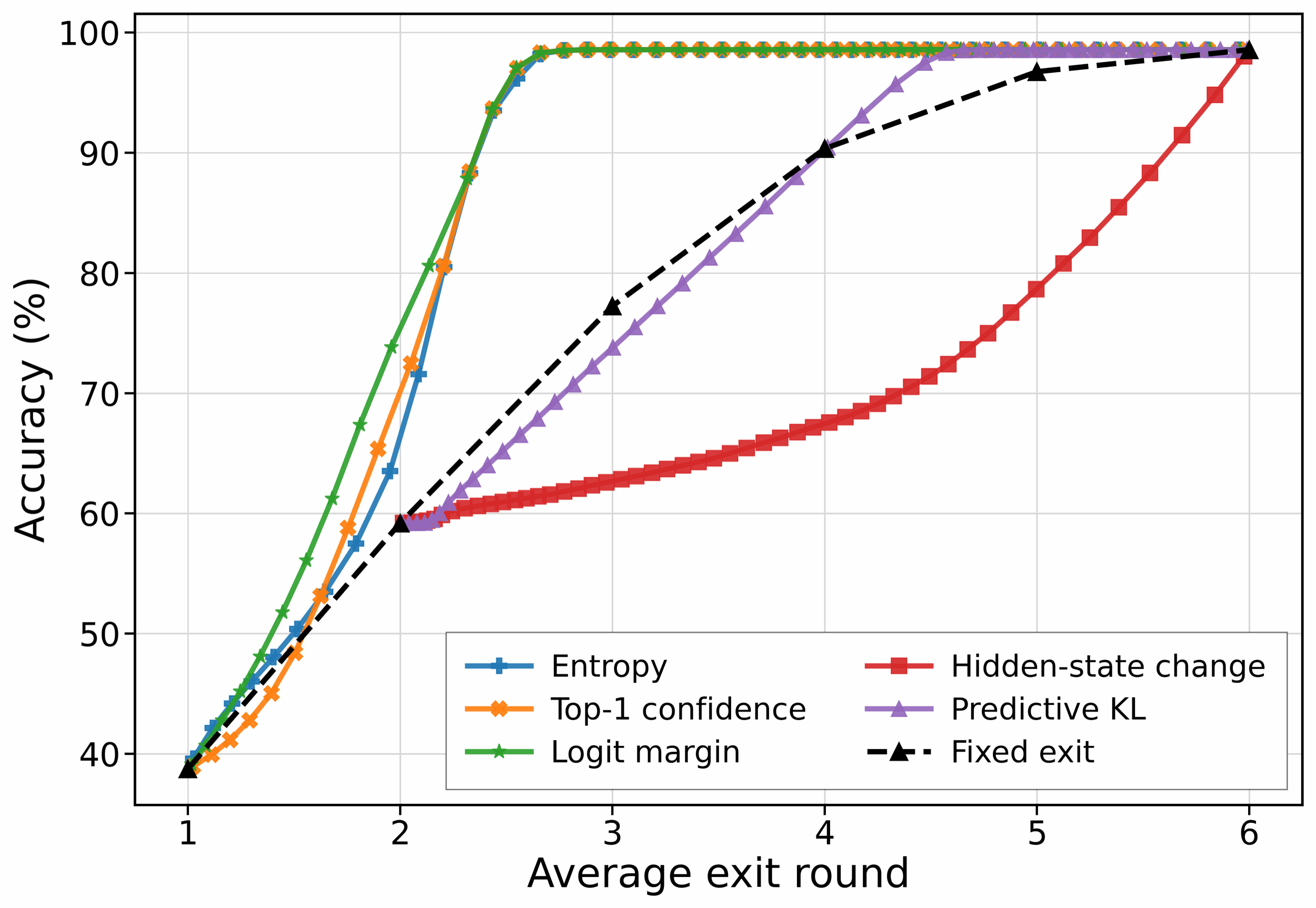}
        \caption{Uniform fixed prior}
        \label{fig:fixed_prior_pareto_uniform}
    \end{subfigure}
    \hfill
    \begin{subfigure}{0.46\linewidth}
        \centering
        \includegraphics[width=\linewidth,trim={0.1in 0.05in 0.1in 0.08in},clip]{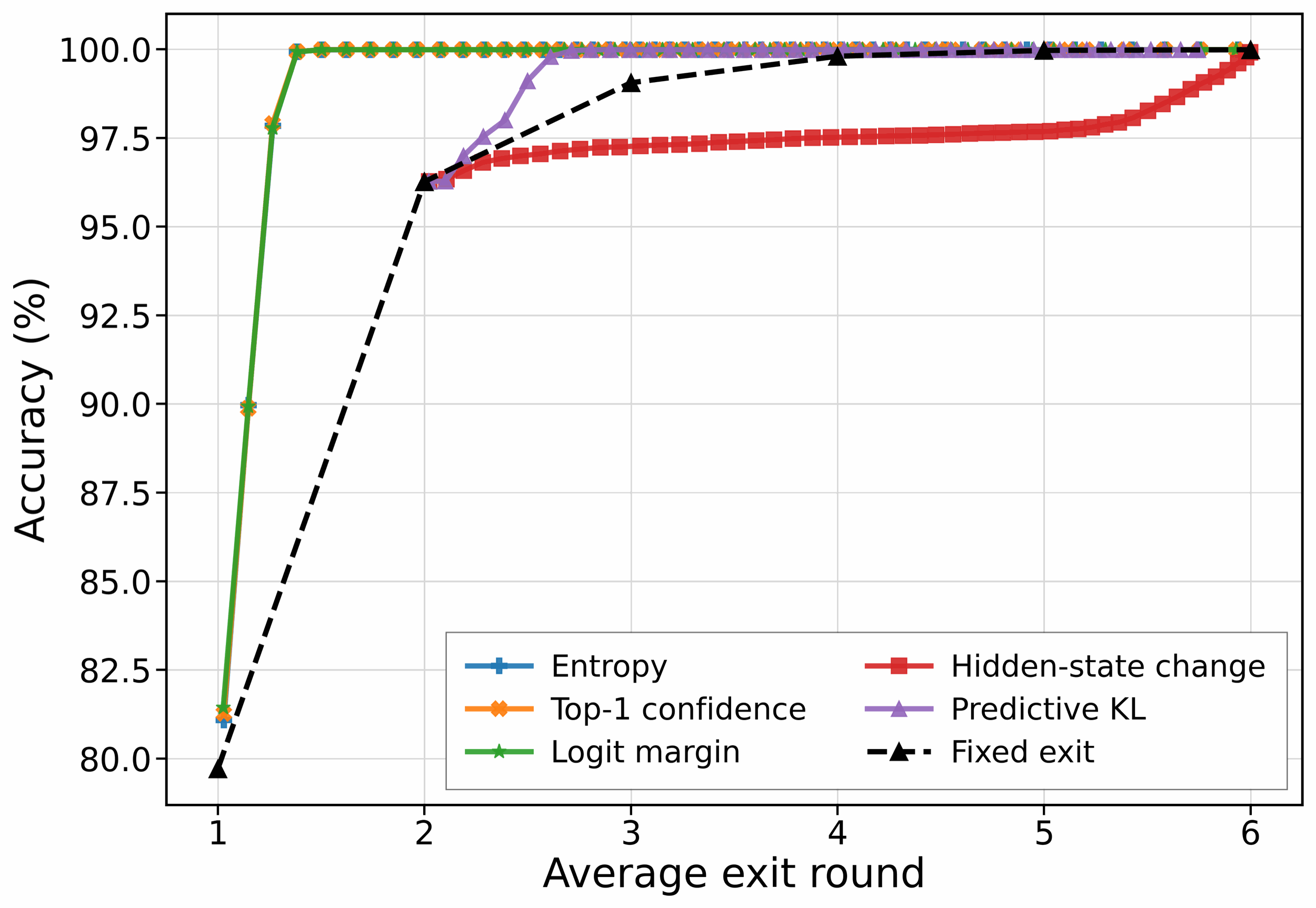}
        \caption{Geometric fixed prior, $\lambda=0.3$}
        \label{fig:fixed_prior_pareto_geom03}
    \end{subfigure}
    \caption{
    Post-hoc readout curves for fixed-prior trajectories.
    We plot test accuracy against average exit depth for the uniform prior and the geometric prior with \(\lambda=0.3\).
    }
    \label{fig:fixed_prior_pareto}
\end{figure}

\begin{table}[t]
\centering
\scriptsize
\setlength{\tabcolsep}{4pt}
\renewcommand{\arraystretch}{0.95}
\caption{
Fixed-prior frontier summary over three seeds.
D@X denotes the minimum average exit depth required to reach X\% test accuracy.
$N/A$ indicates that the corresponding accuracy threshold was unattainable, so no D@X value could be computed.
}
\label{tab:fixed_prior_frontier_summary}

\begin{tabular}{llccc}
\toprule
Training & Readout & D@95 & D@98 & D@99 \\
\midrule
\multirow{4}{*}{Uniform}
& Fixed depth & $4.50\pm0.71$ & $6.00\pm0.00$ & N/A \\
& Entropy     & $2.19\pm0.51$ & $2.66\pm0.44$ & N/A \\
& Top-1 conf. & $2.19\pm0.51$ & $2.66\pm0.44$ & N/A \\
& Margin      & $2.19\pm0.51$ & $2.66\pm0.44$ & N/A \\
\midrule
\multirow{4}{*}{Geom. ($\lambda=.3$)}
& Fixed depth & $2.50\pm0.71$ & $3.50\pm0.71$ & $3.50\pm0.71$ \\
& Entropy     & $1.38\pm0.16$ & $1.50\pm0.16$ & $1.55\pm0.24$ \\
& Top-1 conf. & $1.38\pm0.17$ & $1.50\pm0.16$ & $1.50\pm0.16$ \\
& Margin      & $1.38\pm0.16$ & $1.50\pm0.16$ & $1.55\pm0.24$ \\
\bottomrule
\end{tabular}
\end{table}

Table~\ref{tab:fixed_prior_frontier_summary} confirms the representative curves across three seeds. The geometric prior \(\lambda=0.3\) yields a favorable frontier, with confidence readouts reaching $99\%$ accuracy at roughly 1.5 average loops. Figure~\ref{fig:fixed_prior_pareto} shows representative compute--quality curves for the uniform fixed prior and the geometric fixed prior with \(\lambda=0.3\). In both cases, simple confidence readouts such as entropy, top-1 confidence, and logit margin recover efficient early-exit behavior. Under the uniform prior, these readouts already reach high accuracy using fewer than the maximum number of loops. Under the geometric prior, useful predictions appear earlier in the trajectory, producing stronger tradeoffs at low average exit depth.

\begin{figure}[!t]
    \centering
    \begin{subfigure}{0.46\linewidth}
        \centering
        \includegraphics[width=\linewidth,trim={0.1in 0.05in 0.1in 0.08in},clip]{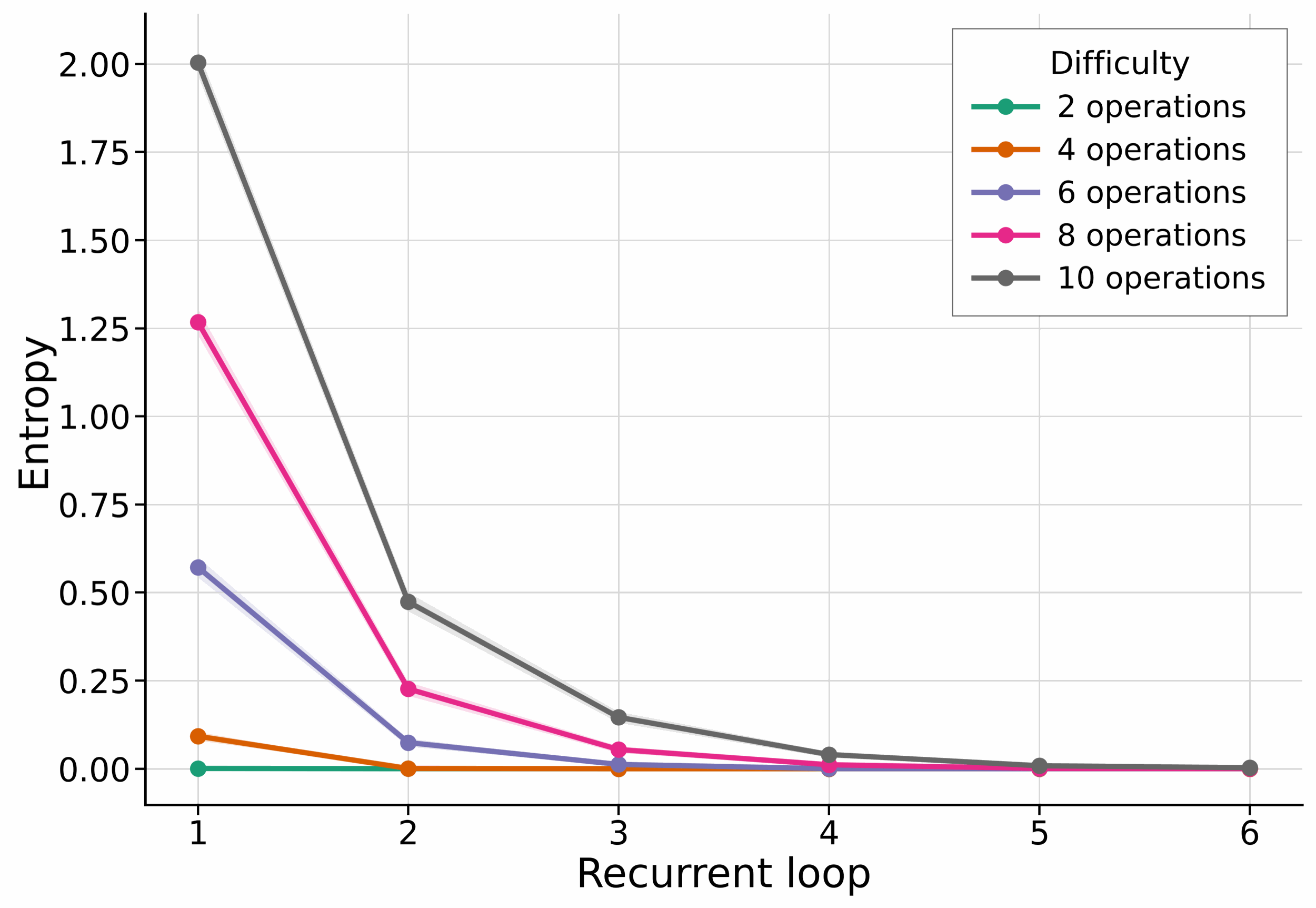}
        \caption{Entropy}
        \label{fig:fixed_prior_entropy_geom03}
    \end{subfigure}
    \hfill
    \begin{subfigure}{0.46\linewidth}
        \centering
        \includegraphics[width=\linewidth,trim={0.1in 0.05in 0.1in 0.08in},clip]{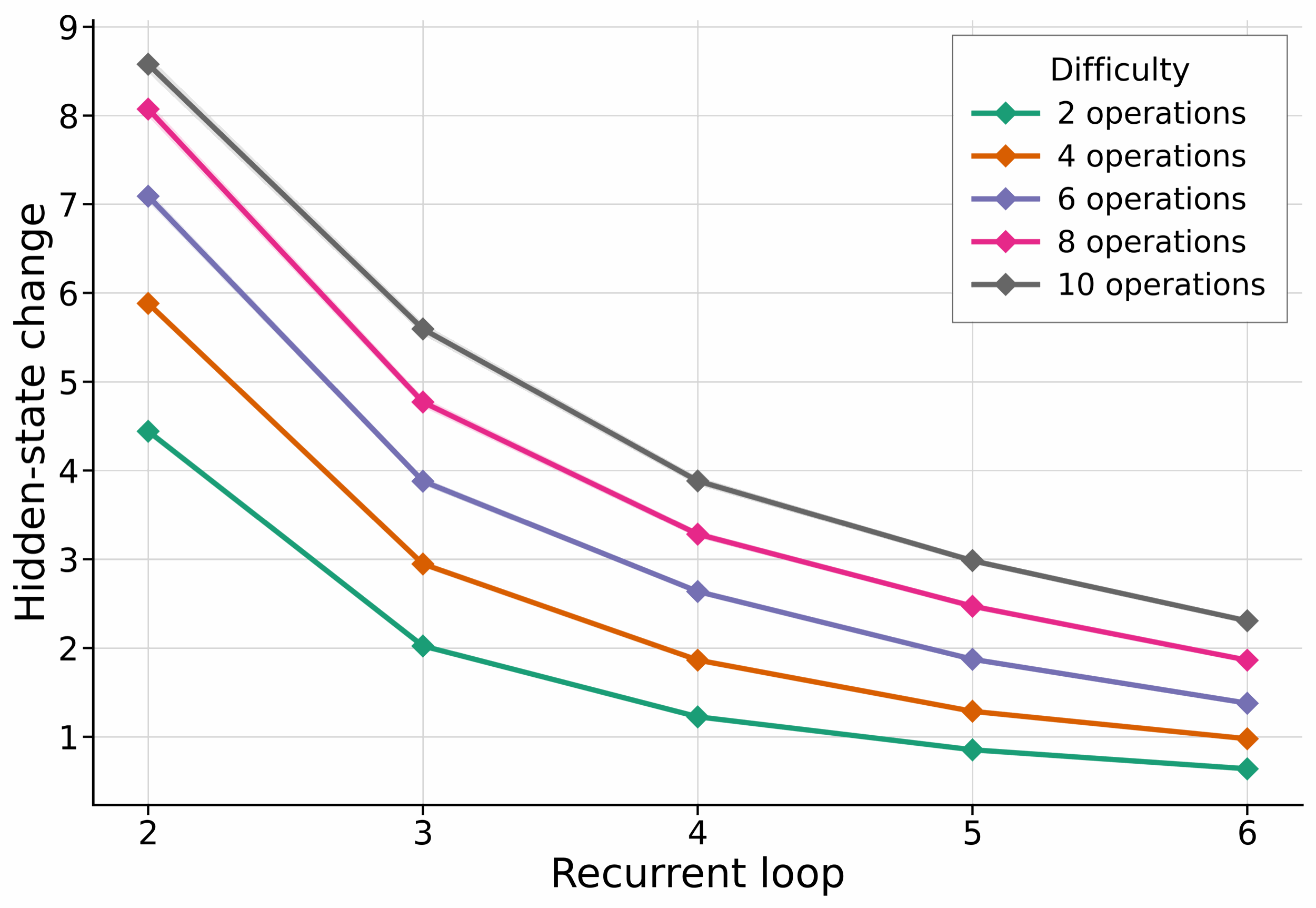}
        \caption{Hidden-state displacement}
        \label{fig:fixed_prior_hidden_geom03}
    \end{subfigure}
    \caption{
    Trajectory signals for the geometric fixed-prior model with $\lambda=0.3$ grouped by difficulty, which is represented as the number of operations in the input example.
    Entropy and hidden-state displacement values are plotted as a function of recurrent depth.
    }
    \label{fig:fixed_prior_convergence}
\end{figure}

The same difficulty-aware structure persists beyond the training range: harder out-of-distribution expressions stay uncertain for more loops and converge more slowly in hidden-state space (Appendix~\ref{app:mano_difficulty_extrapolation}). Full sweeps and diagnostics are in Appendix~\ref{app:fixed_prior_difficulty}, and the same pattern holds on a second synthetic benchmark, binary parity over bit strings of length 1--40 (Appendix~\ref{app:parity_diagnostic}), where fixed-prior trajectories support strong post-hoc readouts and diagnostics stay ordered by bit length.

\subsection{Fixed-prior readouts can match or outperform learned gates}
\label{sec:results_fixed_vs_gate}

\begin{figure}[t]
    \centering
    \begin{subfigure}{0.44\linewidth}
        \centering
        \includegraphics[width=\linewidth,trim={0.1in 0.05in 0.1in 0.08in},clip]{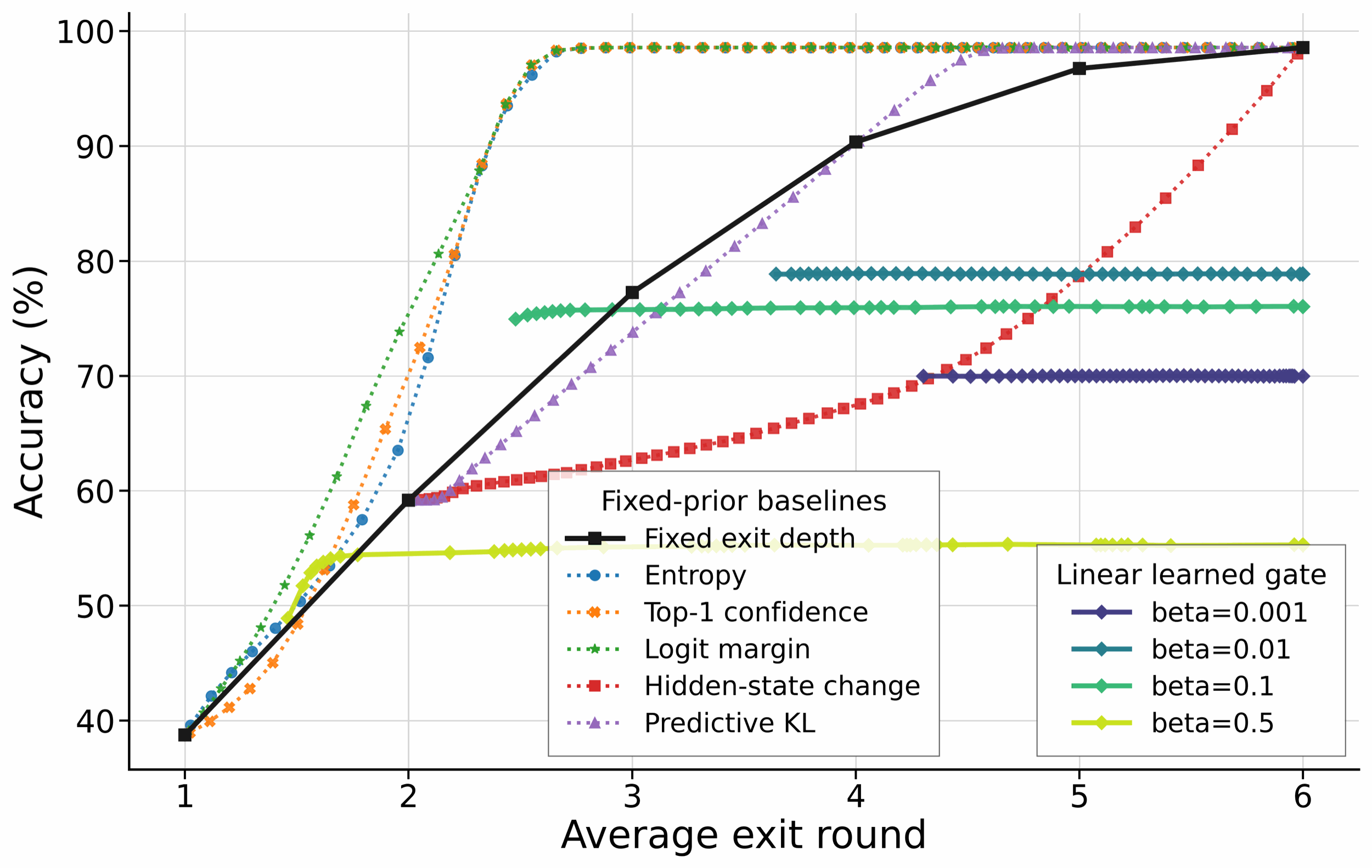}
        \caption{Linear gate, uniform prior}
        \label{fig:linear_gate_vs_fixed_uniform}
    \end{subfigure}
    \hfill
    \begin{subfigure}{0.44\linewidth}
        \centering
        \includegraphics[width=\linewidth,trim={0.1in 0.05in 0.1in 0.08in},clip]{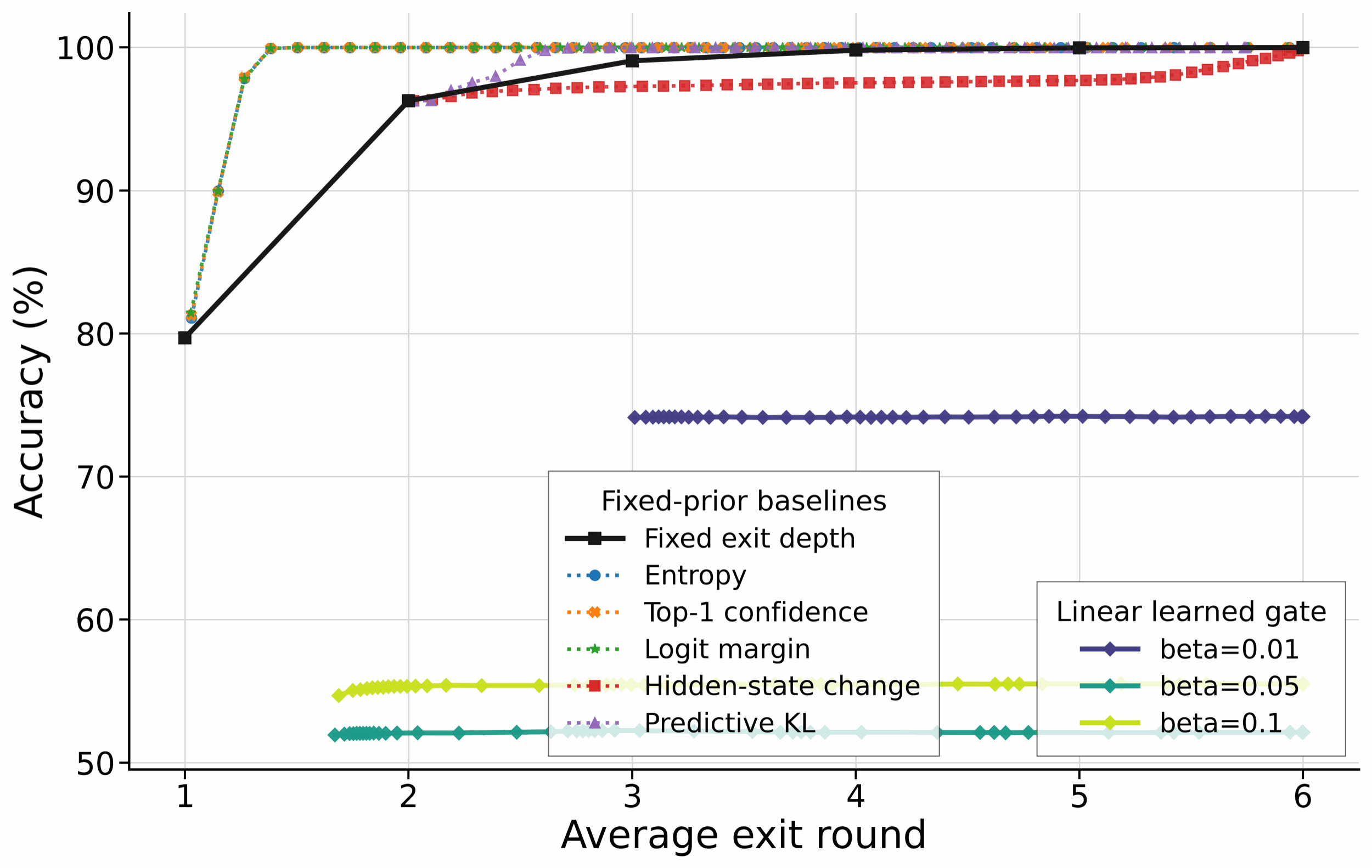}
        \caption{Linear gate, geometric prior ($\lambda=0.3$)}
        \label{fig:linear_gate_vs_fixed_geom03}
    \end{subfigure}
    \vspace{2pt}
    \begin{subfigure}{0.44\linewidth}
        \centering
        \includegraphics[width=\linewidth,trim={0.1in 0.05in 0.1in 0.08in},clip]{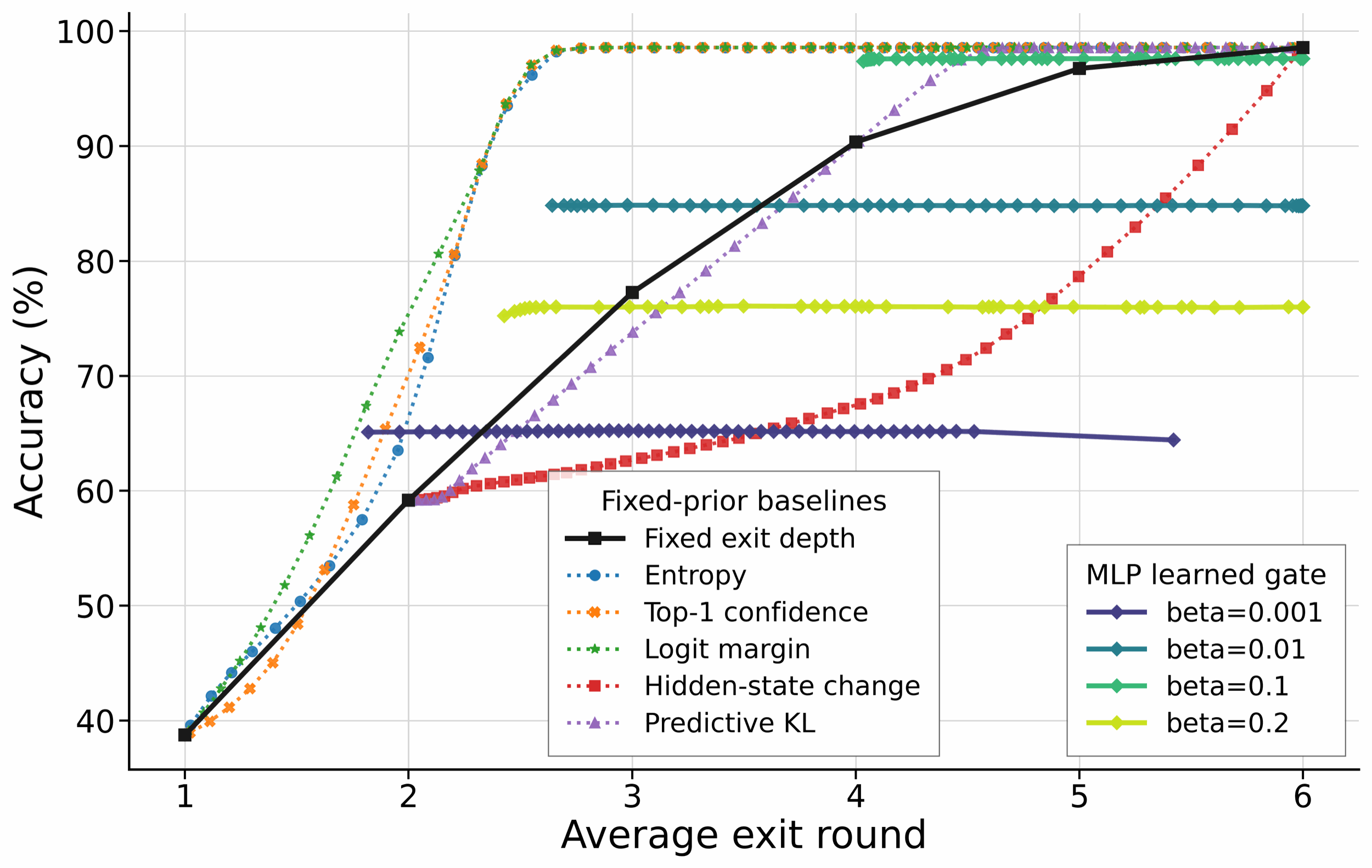}
        \caption{MLP gate, uniform prior}
        \label{fig:mlp_gate_vs_fixed_uniform}
    \end{subfigure}
    \hfill
    \begin{subfigure}{0.44\linewidth}
        \centering
        \includegraphics[width=\linewidth,trim={0.1in 0.05in 0.1in 0.08in},clip]{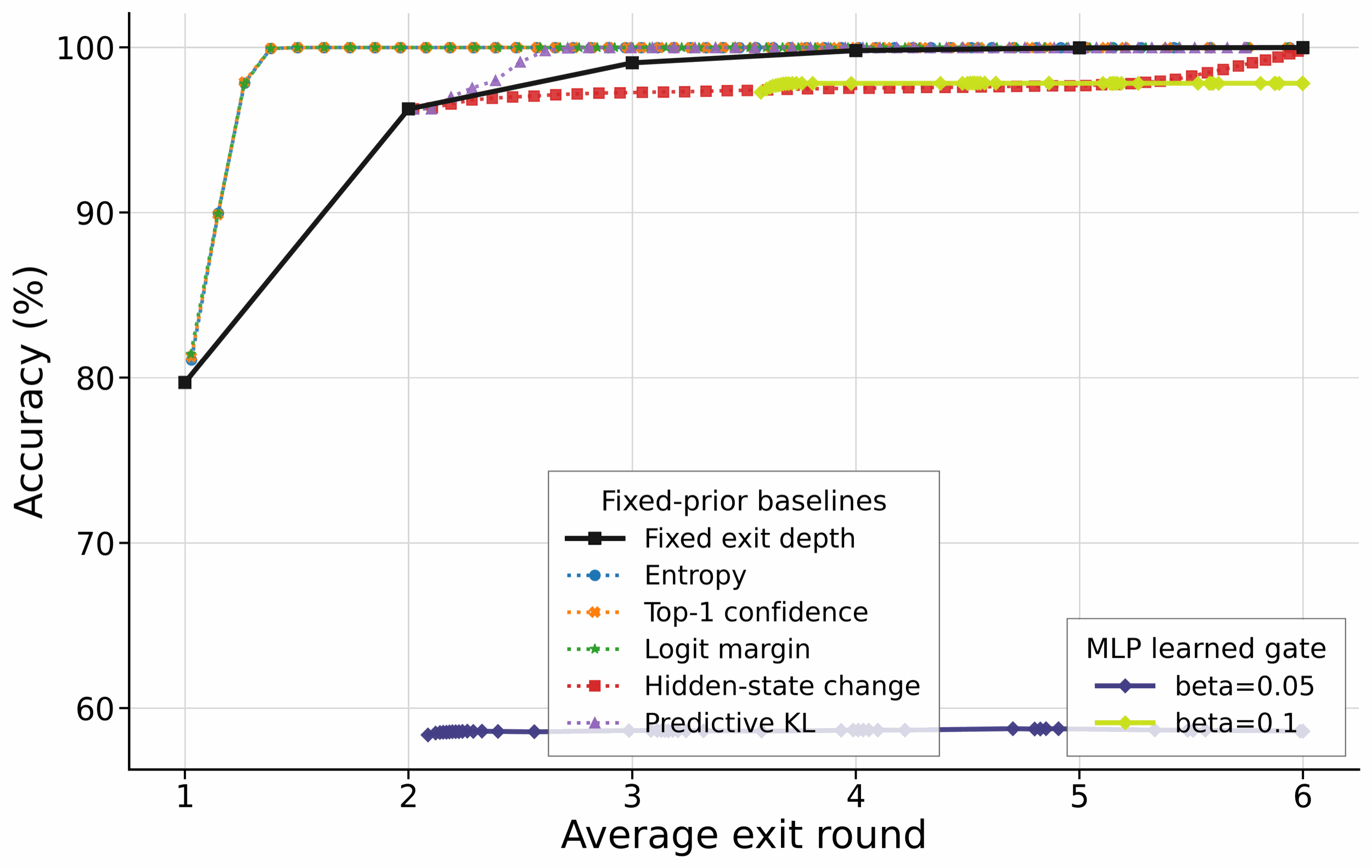}
        \caption{MLP gate, geometric prior ($\lambda=0.3$)}
        \label{fig:mlp_gate_vs_fixed_geom03}
    \end{subfigure}
    \caption{
    Fixed-prior readouts compared with learned linear and MLP gates under uniform and geometric priors. 
    }
    \label{fig:gate_vs_fixed}
\end{figure}

\begin{table}[t]
\centering
\scriptsize
\setlength{\tabcolsep}{4pt}
\renewcommand{\arraystretch}{0.95}
\caption{
Comparison of fixed-prior readouts and learned gates over three seeds.
D@X denotes the minimum average exit depth required to reach X\% test accuracy. $N/A$ indicates that the target accuracy was not reached by any model across the three seeds.
}
\label{tab:fixed_vs_gate_multiseed_summary}

\begin{tabular}{llccc}
\toprule
Training & Readout & D@95 & D@98 & D@99 \\
\midrule
Fixed (Uniform)
    & Margin      & $2.19\pm0.51$ & $2.66\pm0.44$ & N/A \\
Linear (Uniform)
    & Native gate & $4.49\pm0.28$ & $4.49\pm0.28$ & N/A \\
MLP (Uniform)
    & Native gate & $4.30\pm0.34$ & N/A & N/A \\
\midrule
Fixed (Geom., $\lambda=.3$)
    & Margin      & $1.38\pm0.16$ & $1.50\pm0.16$ & $1.55\pm0.24$ \\
Linear (Geom., $\lambda=.3$)
    & Native gate & $2.80\pm0.30$ & N/A & N/A \\
MLP (Geom., $\lambda=.3$)
    & Native gate & $3.58\pm0.36$ & N/A & N/A \\
\bottomrule
\end{tabular}
\end{table}

We next compare fixed-prior trajectories with jointly trained learned-gate trajectories. Fixed-prior models use post-hoc readouts, whereas learned-gate models use their native CDF-threshold readout.

Figure~\ref{fig:gate_vs_fixed} shows that fixed-prior trajectories paired with simple confidence readouts consistently outperform learned linear gates. Increasing gate capacity improves the learned baseline, but MLP gates still fail to match the best fixed-prior frontiers.

Table~\ref{tab:fixed_vs_gate_multiseed_summary} summarizes the same comparison numerically across three seeds. To avoid giving the fixed-prior models an adaptive choice of readout in this table, we use a single explicit confidence readout, logit margin, while learned-gate models use their native gate readout with the highest-performing \(\beta\) for each prior and gate class. Thus, the comparison favors the learned-gate side by allowing \(\beta\) selection, while using a fixed post-hoc readout for the fixed-prior models.

Across three seeds, the geometric fixed-prior trajectory paired with logit margin achieves a favorable frontier, requiring only 1.4–1.6 average loops to reach $95\%--99\%$ accuracy. Learned gates consistently require greater depth, indicating that trajectory quality, not gate parameterization alone, determines adaptive-compute performance. Additional gate comparisons across all prior settings are reported in Appendix~\ref{app:gate_vs_fixed}, with separate full sweeps for the linear gate in Appendix~\ref{app:linear-gate} and the MLP gate in Appendix~\ref{app:mlp-gate}.

\subsection{The training objective shapes trajectories through depth-weighted losses}
\label{sec:training-objective-shapes-trajectories}

\begin{figure}[!t]
    \centering

    \begin{subfigure}{0.85\linewidth}
        \centering
        \includegraphics[width=\linewidth]{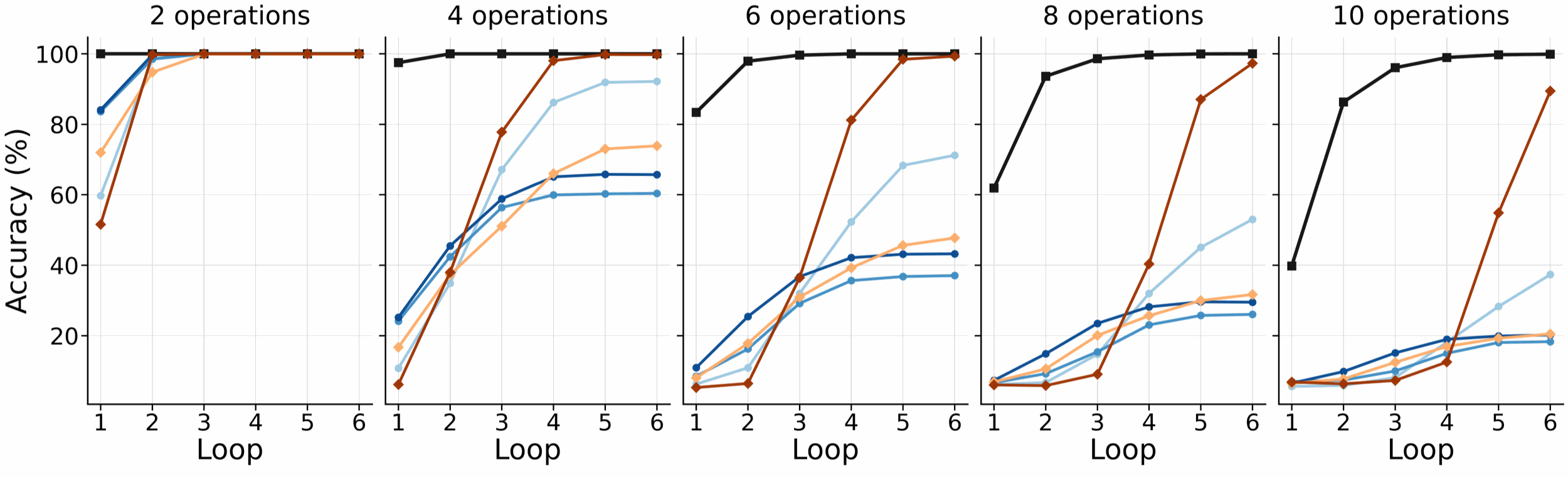}
        \caption{Forced-exit accuracy}
        \label{fig:geom03_forced_exit_performance_by_difficulty}
    \end{subfigure}

    \vspace{0.5em}

    \begin{subfigure}{0.85\linewidth}
        \centering
        \includegraphics[width=\linewidth]{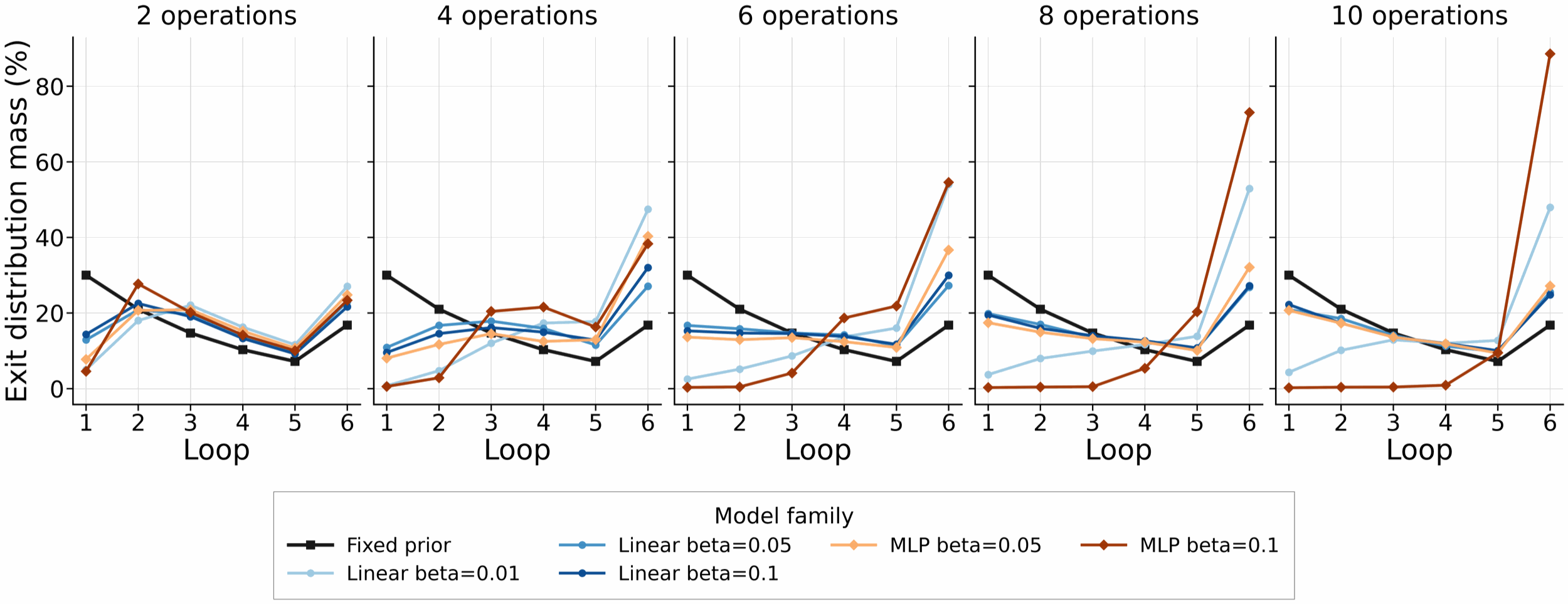}
        \caption{Learned exit distribution}
        \label{fig:geom03_exit_distribution_by_difficulty}
    \end{subfigure}

    \caption{
    Forced-exit trajectories and learned exit distributions for the geometric prior with $\lambda=0.3$.
    Panel (a) evaluates accuracy at each recurrent loop without using an adaptive readout.
    Panel (b) shows the average learned exit distribution over recurrent loops, stratified by task difficulty.
    }
    \label{fig:gate_shapes_trajectory}
\end{figure}

We now examine why these frontiers differ. Because the learned exit distribution also weights per-depth training losses, it shapes the recurrent trajectory in addition to defining the inference-time stopping rule. We therefore compare forced exits, which evaluate every recurrent loop without adaptive stopping.

Figure~\ref{fig:gate_shapes_trajectory} compares forced-exit trajectories and learned exit distributions for the geometric prior with \(\lambda=0.3\). The forced-exit curves show that the training objective changes the per-depth accuracy distribution. Fixed-prior training produces useful intermediate predictions early in the trajectory, including on harder examples. The trained gate trajectories show different depth distributions, indicating that the gate objective changes the trajectory rather than only changing the exit rule. The learned exit distributions are difficulty-aware, assigning harder examples more probability mass at later loops. However, the forced-exit curves show that allocating later exits alone is insufficient: the trajectory must also produce accurate intermediate states.

These results explain why adaptive-compute performance depends on both trajectory formation and exit selection. Learned gates entangle these two roles: the same distribution \(q(\cdot\mid x)\) both weights the losses that shape the trajectory and defines the native stopping rule used at inference time. Fixed-prior training separates these roles by shaping the trajectory with input-independent depth supervision, after which simple readouts can select exits. The multiseed frontier summaries in Tables~\ref{tab:fixed_prior_frontier_summary} and~\ref{tab:fixed_vs_gate_multiseed_summary} are consistent with this trajectory-level diagnosis. Objectives that produce stronger intermediate forced exits also support stronger adaptive frontiers. Full forced-exit trajectories and learned exit-distribution diagnostics across prior settings are provided in Appendix~\ref{app:forced_exit_all}.

\subsection{Post-hoc fitting of a learned gate over a pretrained trajectory}
\label{sec:posthoc-gate-fitting}

A remaining ambiguity in the gate results is whether the gate readout itself is weak, or whether the problem arises from jointly training the gate with the recurrent trajectory. To separate these effects, we freeze pretrained MANO trajectories and train a new exit gate on top of the frozen hidden states. In this setting, the gate can only act as a readout: it has no path to modify the recurrent states, the LM head, or the per-depth predictions.

We consider two post-hoc gate objectives. The first is the same PonderNet-style objective used for learned gates, but with the model fixed. The new gate defines an exit distribution over depths and is trained to minimize the expected per-depth cross-entropy plus the same prior regularizer. The second is a marginal-utility objective inspired by Ouro Stage-II training. For each adjacent pair of recurrent depths, we compute the realized loss improvement from continuing one more step and train the gate to predict whether continuation is worthwhile. In both cases, we train either a linear gate or a two-layer MLP gate, evaluate with the same validation-selected threshold protocol as the other readouts, and report the full linear/MLP results across fixed-prior and learned-gate trajectories in Appendix~\ref{app:posthoc-gates}.

\begin{figure}[t]
\centering
\begin{subfigure}[t]{0.40\linewidth}
    \centering
    \includegraphics[width=\linewidth,trim={0.1in 0.05in 0.1in 0.08in},clip]{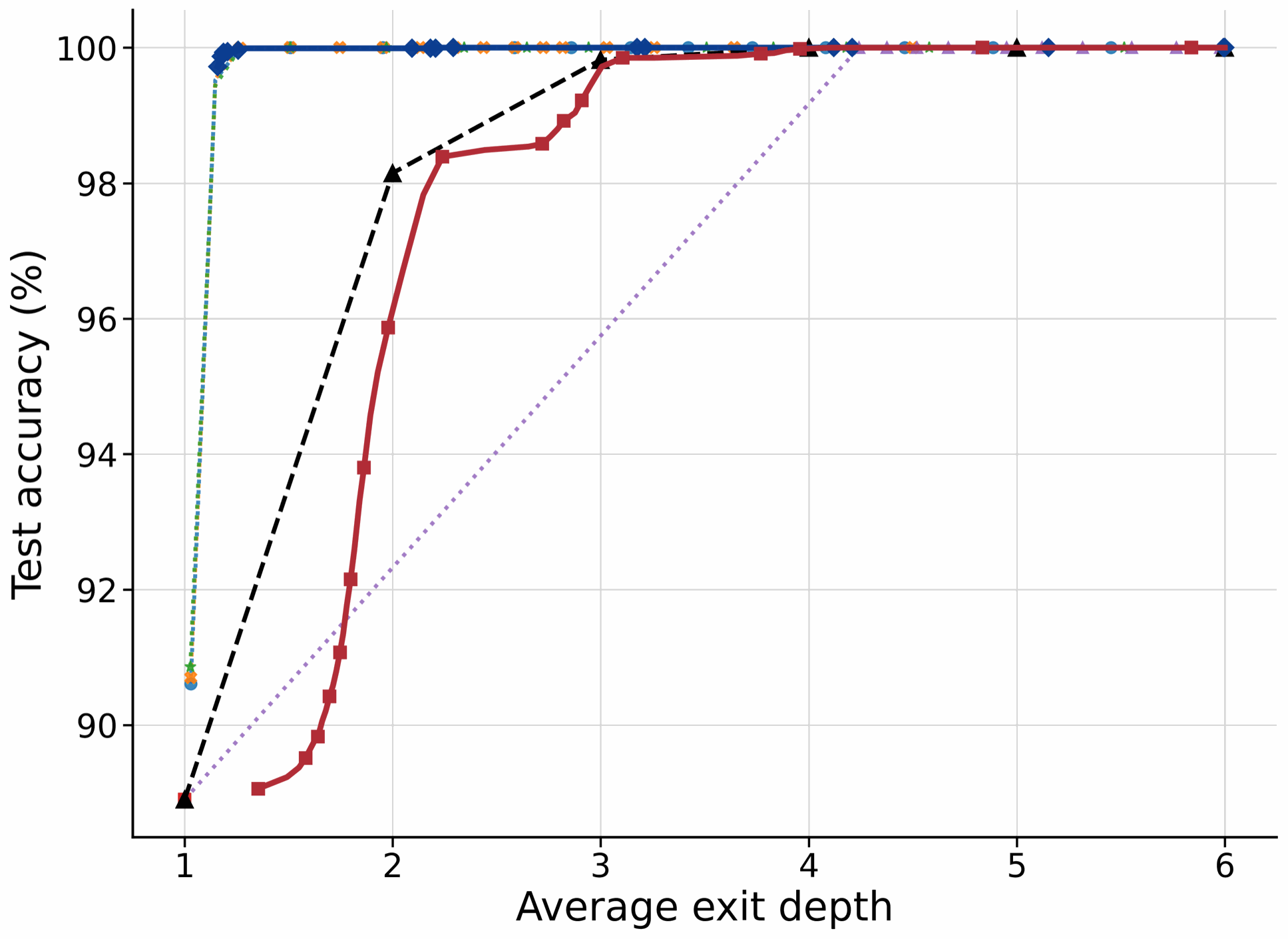}
    \caption{Frozen fixed-prior trajectory ($\lambda=0.3$).}
    \label{fig:posthoc_gate_fixed_geom03}
\end{subfigure}
\hfill
\begin{subfigure}[t]{0.52\linewidth}
    \centering
    \includegraphics[
        width=\linewidth,
        trim={0in 0.05in 0.0in 0.08in}, 
        clip
    ]{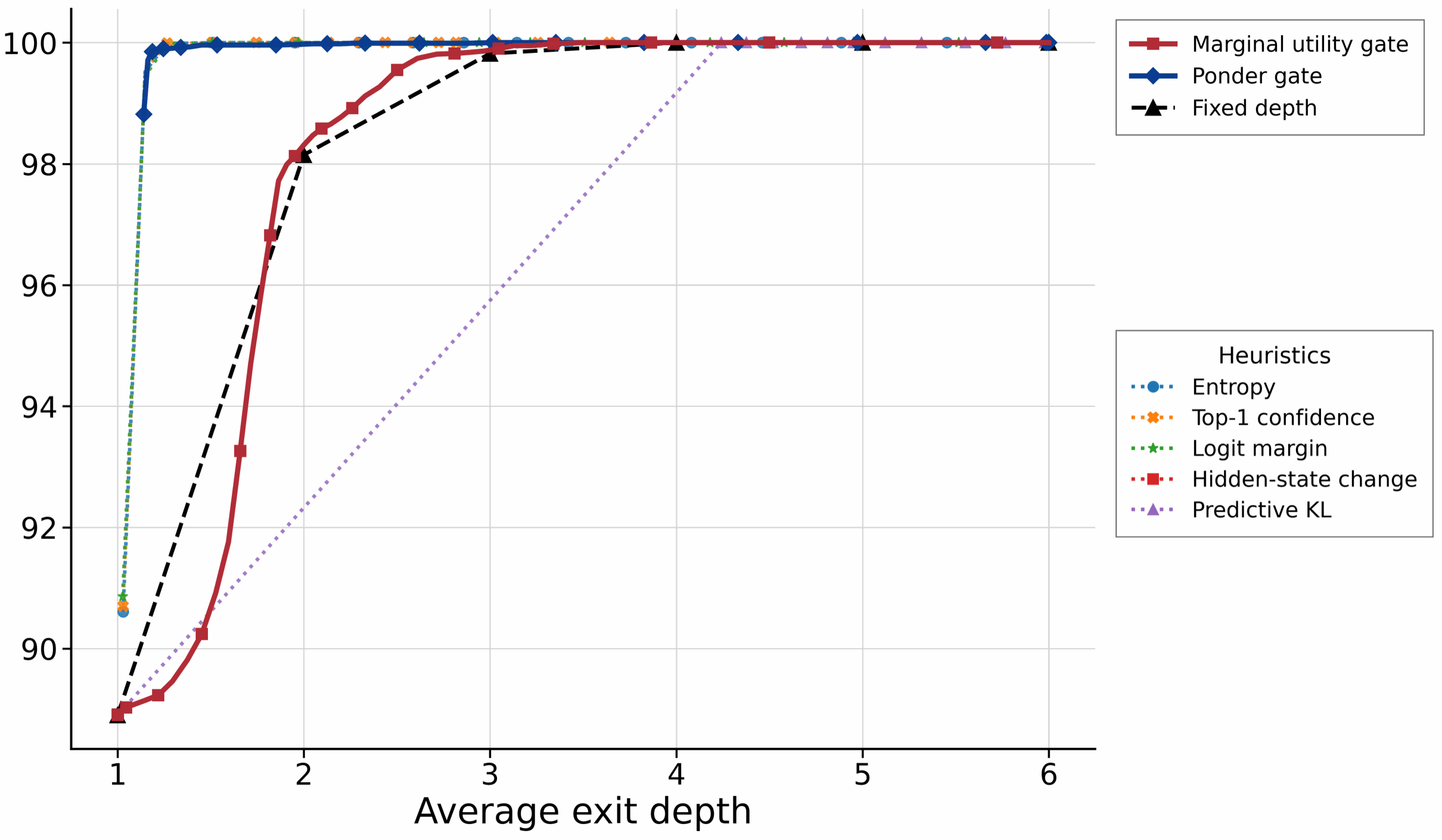}
    \caption{Frozen learned-gate trajectory ($\lambda=0.3$, $\beta=0.1$).}
\end{subfigure}
\caption{
Post-hoc gate training on frozen MANO trajectories: we freeze the backbone and LM head, fit new linear or MLP exit gates, and compare against simple trajectory readouts.
}
\label{fig:posthoc_gate_main}
\end{figure}

Figure~\ref{fig:posthoc_gate_main} shows two representative frozen trajectories. On the fixed-prior geometric trajectory, post-hoc gates recover strong early-exit behavior, reaching near-final-loop accuracy at much lower average depth. This shows that the gate class is capable of representing useful stopping policies when trajectory formation is separated from readout learning. However, the simple readouts remain competitive, indicating that much of the stopping signal is already exposed in the prediction trajectory.

On the frozen trajectory produced by joint learned-gate training, fitting a new post-hoc gate does not improve performance. The newly trained gates largely track the same frontier as the native gate and remain limited by the quality of the underlying trajectory. Across these frozen-trajectory experiments, the PonderNet-style post-hoc objective also gives a stronger readout than the marginal-utility objective, which tends to exit later and does not improve the frontier. This suggests that the weak gate result is not simply due to an under-trained or underexpressive native gate. Instead, the main issue is the trajectory induced by the joint objective. Thus, learned gates are best understood not only as exit readouts, but also as training mechanisms that shape which recurrent states become useful.

\subsection{From controlled tasks to large-scale checkpoints}
\label{sec:large-scale}

\begin{figure}[t]
    \centering
    \begin{subfigure}{0.42\linewidth}
        \centering
        \includegraphics[width=\linewidth,trim={0.1in 0.05in 0.1in 0.08in},clip]{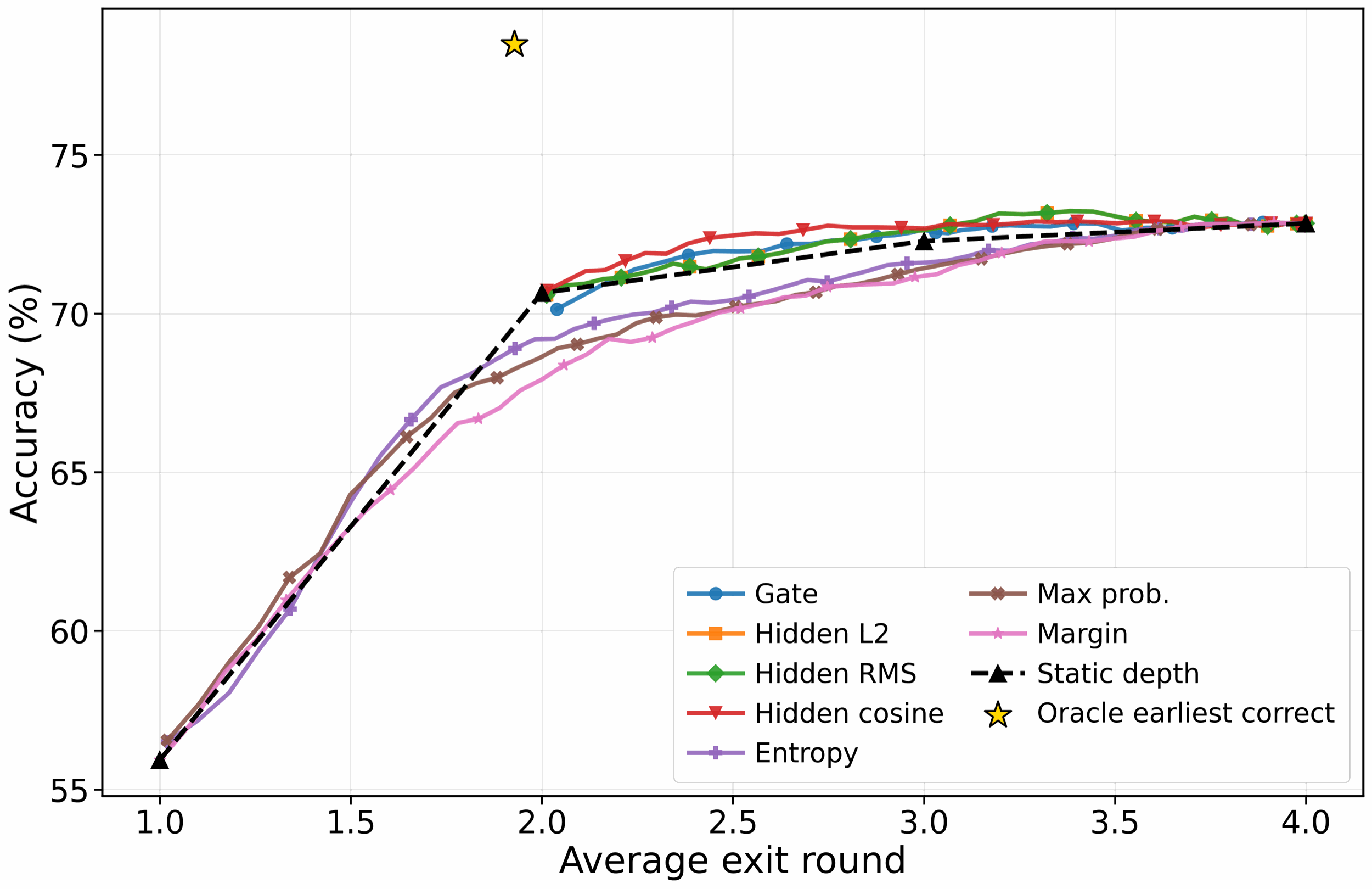}
        \caption{Ouro-1.4B, HellaSwag}
        \label{fig:ouro_14b_hellaswag}
    \end{subfigure}
    \hfill
    \begin{subfigure}{0.42\linewidth}
        \centering
        \includegraphics[width=\linewidth,trim={0.1in 0.05in 0.1in 0.08in},clip]{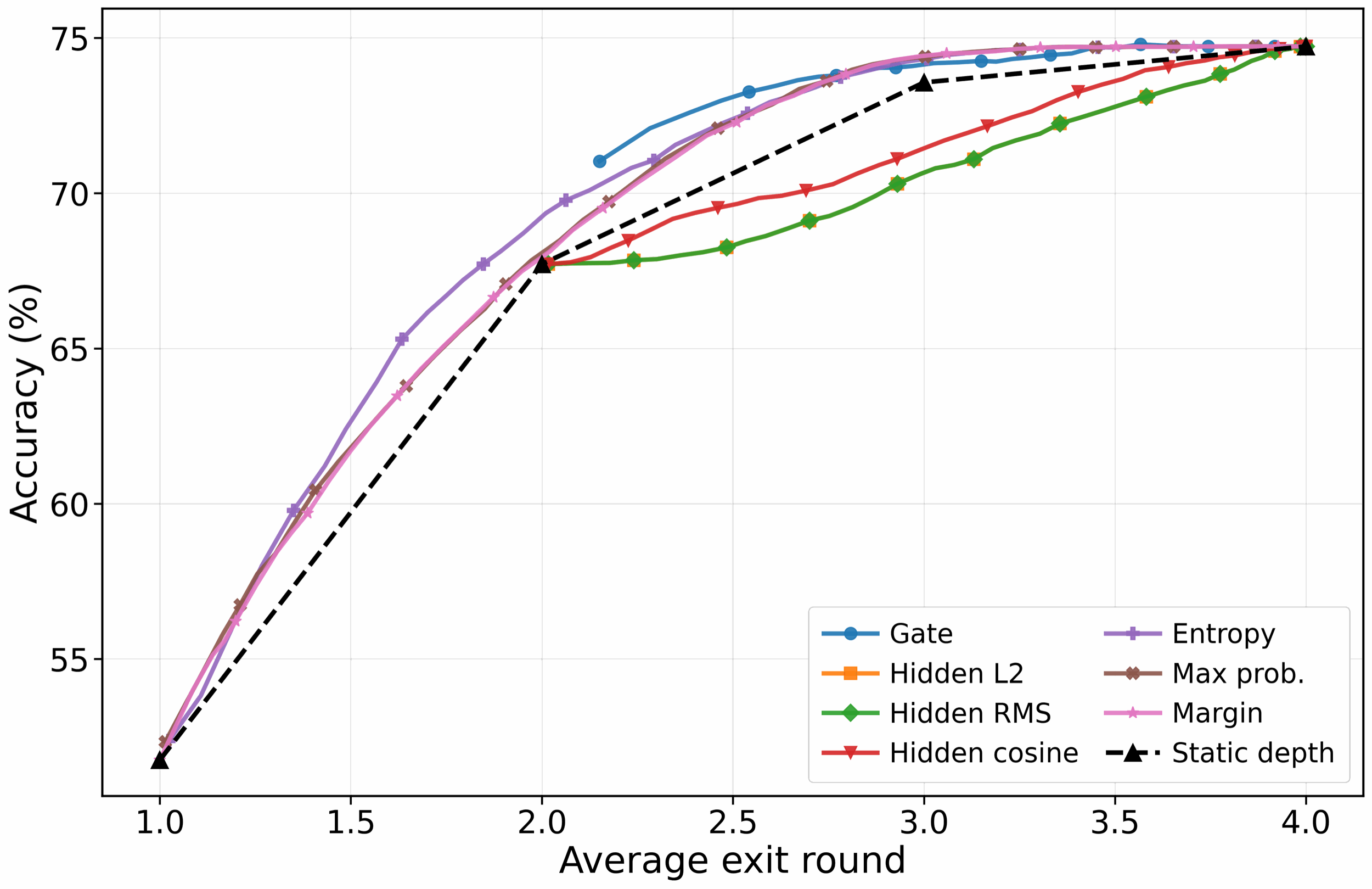}
        \caption{Ouro-2.6B, MMLU}
        \label{fig:ouro_26b_mmlu}
    \end{subfigure}
    \caption{
    Large-scale Ouro readout comparisons.
    Each panel plots benchmark accuracy against average loops.
    The pretrained ponder gate is compared with confidence and convergence readouts applied to the same recurrent trajectories.
    }
    \label{fig:ouro_pareto_main}
\end{figure}

\begin{table}[t]
\centering
\footnotesize
\caption{
Ouro readout comparison across benchmarks.
For each model and benchmark, we compare the pretrained gate against the best post-hoc readout, both validation-selected, on the held-out split.
}
\label{tab:ouro_readout_summary}
\resizebox{\linewidth}{!}{
\begin{tabular}{llcc lcc}
\toprule
Model & Benchmark & Gate accuracy & Gate avg. loops & Post-hoc readout & Post-hoc accuracy & Post-hoc avg. loops \\
\midrule
Ouro-1.4B & MMLU & 67.6 & 3.75 & Hidden cosine & 67.3 & 3.86 \\
Ouro-1.4B & ARC-Easy & 82.9 & 2.72 & Hidden cosine & 83.2 & 2.69 \\
Ouro-1.4B & ARC-Challenge & 59.8 & 2.65 & Max prob. & 59.5 & 3.12 \\
Ouro-1.4B & OpenBookQA & 42.8 & 3.13 & Hidden cosine & 41.6 & 2.95 \\
Ouro-1.4B & HellaSwag & 72.8 & 3.94 & Hidden L2 & 72.9 & 3.13 \\
Ouro-1.4B & CommonsenseQA & 74.7 & 3.43 & Hidden cosine & 74.9 & 3.93 \\
\midrule
Ouro-2.6B & MMLU & 74.4 & 3.33 & Max prob. & 74.7 & 3.30 \\
Ouro-2.6B & ARC-Easy & 87.2 & 3.68 & Hidden cosine & 87.2 & 3.92 \\
Ouro-2.6B & ARC-Challenge & 66.6 & 3.38 & Margin & 65.8 & 2.90 \\
Ouro-2.6B & OpenBookQA & 43.4 & 3.78 & Hidden cosine & 41.6 & 2.63 \\
Ouro-2.6B & HellaSwag & 77.5 & 2.93 & Hidden cosine & 77.6 & 3.69 \\
Ouro-2.6B & CommonsenseQA & 81.1 & 2.49 & Entropy & 80.5 & 2.37 \\
\bottomrule
\end{tabular}
}
\end{table}

We ask whether the same trajectory--readout pattern appears in larger looped language models. Figure~\ref{fig:ouro_pareto_main} shows representative Pareto curves for the two model sizes. Because these checkpoints are fixed, we vary only the readout, not the trajectory. Unlike MANO, we cannot retrain under a different prior, so the trajectory is whatever pretraining produced. On Ouro-1.4B HellaSwag, post-hoc readouts can match or improve the gate's compute--accuracy tradeoff; on Ouro-2.6B MMLU, the ponder gate remains competitive, but post-hoc readouts still recover similar or better tradeoffs. The best readout varies across model size and benchmark, indicating that the learned gate is not necessarily learning a criterion superior to classical heuristics.

Table~\ref{tab:ouro_readout_summary} summarizes the full benchmark grid numerically. The post-hoc readouts are highly competitive with the pretrained ponder gate across both model sizes. In five of the twelve model--benchmark pairs, the selected post-hoc readout achieves higher held-out accuracy than the gate, though several of these differences are small and reported on single runs. In seven pairs, it uses fewer average recurrent loops, and in three pairs, it improves both accuracy and compute. Several of the more promising post-hoc readouts are simple confidence or convergence criteria, such as hidden-state cosine similarity, hidden-state displacement, maximum probability, entropy, and logit margin.

Across the full evaluation, the qualitative pattern is consistent with the controlled experiments: recurrent trajectories contain stopping signals that simple confidence or convergence readouts can expose. In the controlled setting we change the training objective and show that trajectory formation matters; in the large-scale setting the trajectory is fixed by pretraining, yet alternative readouts still recover competitive early-exit behavior. This supports treating adaptive computation as a trajectory--readout problem rather than a property of the learned gate alone. Full per-benchmark Pareto curves and reasoning-tuned Ouro-Thinking diagnostics are in Appendices~\ref{app:ouro_pareto} and~\ref{app:ouro_thinking}.

\subsection{Average exit depth corresponds to latency savings}
\label{sec:latency}

Our main compute metric is average exit depth, which directly counts applications of the shared recurrent block. We additionally evaluate whether this proxy corresponds to latency savings once readout computation and adaptive-exit overhead are included. Tables~\ref{tab:mano_latency} and \ref{tab:ouro_latency} report measured end-to-end latency on representative models trained on the MANO task, as well as on the Ouro model. For these experiments, we use an NVIDIA RTX 5070 Ti GPU.

\begin{table}[t]
\centering
\footnotesize
\caption{
Latency measurements on MANO and Ouro-1.4B.
(a) Ouro-1.4B across evaluated benchmarks; for each dataset we report
the final-loop baseline, the pretrained gate, and the best heuristic readout.
(b) Representative fixed-prior and learned-gate models on MANO;
the fixed-prior model uses a geometric prior with $\lambda=0.3$,
the learned-gate model uses the same prior with $\beta=0.1$.
}
\label{tab:latency_combined}

\begin{subtable}[t]{0.48\linewidth}
\centering
\caption{Ouro-1.4B}
\label{tab:ouro_latency}
\resizebox{\linewidth}{!}{
\begin{tabular}{llrrrr}
\toprule
Dataset & Readout & Acc.\ (\%) & Avg.\ loops & ms/ex. & Speedup \\
\midrule
OpenBookQA & Final loop & 43.0 & 4.00 & 99.1 & 1.00$\times$ \\
OpenBookQA & Gate & 42.0 & 2.80 & 69.3 & 1.43$\times$ \\
OpenBookQA & Hidden cos. & 39.4 & 2.01 & 49.8 & 1.99$\times$ \\
\midrule
ARC-Easy & Final loop & 84.0 & 4.00 & 170.9 & 1.00$\times$ \\
ARC-Easy & Gate & 82.4 & 2.53 & 108.0 & 1.58$\times$ \\
ARC-Easy & Margin & 82.8 & 2.98 & 127.3 & 1.34$\times$ \\
\midrule
ARC-Challenge & Final loop & 60.2 & 4.00 & 616.2 & 1.00$\times$ \\
ARC-Challenge & Gate & 59.5 & 2.65 & 408.8 & 1.51$\times$ \\
ARC-Challenge & Hidden L2 & 59.7 & 2.86 & 440.3 & 1.40$\times$ \\
\midrule
CommonsenseQA & Final loop & 75.6 & 4.00 & 79.9 & 1.00$\times$ \\
CommonsenseQA & Gate & 75.1 & 3.23 & 64.4 & 1.24$\times$ \\
CommonsenseQA & Margin & 75.4 & 3.31 & 66.1 & 1.21$\times$ \\
\bottomrule
\end{tabular}
}
\end{subtable}
\hfill
\begin{subtable}[t]{0.48\linewidth}
\centering
\caption{MANO}
\label{tab:mano_latency}
\resizebox{\linewidth}{!}{
\begin{tabular}{llrrrr}
\toprule
Model & Readout & Acc.\ (\%) & Avg.\ loops & ms/ex. & Speedup \\
\midrule
Fixed prior & Final loop & 100.0 & 6.00 & 0.066 & 1.00$\times$ \\
Fixed prior & Top-1 conf. & 100.0 & 1.26 & 0.049 & 1.36$\times$ \\
Fixed prior & Logit margin & 100.0 & 1.26 & 0.051 & 1.29$\times$ \\
Fixed prior & Pred.\ KL & 99.9 & 2.32 & 0.060 & 1.11$\times$ \\
Fixed prior & Hidden disp. & 99.5 & 5.80 & 0.074 & 0.90$\times$ \\
\midrule
Learned gate & Final loop & 76.1 & 6.00 & 0.066 & 1.00$\times$ \\
Learned gate & Gate & 75.4 & 1.59 & 0.053 & 1.24$\times$ \\
\bottomrule
\end{tabular}
}
\end{subtable}

\end{table}

In the controlled MANO setting, post-hoc readouts on the fixed-prior trajectory reduce measured latency while preserving accuracy. For the geometric fixed-prior model with \(\lambda=0.3\), top-1 confidence and logit margin maintain \(100.0\%\) accuracy while reducing average depth from 6.0 to 1.26 loops, giving \(1.36\times\) and \(1.29\times\) speedups; predictive KL preserves near-perfect accuracy at \(1.11\times\). These are smaller than the raw loop-count reduction, as expected when the small recurrent block is not the only runtime cost, but they confirm that early exits yield real savings; the hidden-displacement readout shows the opposite case, exiting late and adding overhead. The learned gate reduces average depth from 6.0 to 1.59 loops (\(1.24\times\)) but operates at lower accuracy (around \(75\%\)), so here the fixed-prior trajectory with simple confidence readouts gives a more favorable accuracy--latency tradeoff.

The Ouro estimates provide the same latency-side sanity check at larger scale. Across all evaluated benchmarks, reducing the average loops produces substantial estimated latency reductions relative to the final-loop baseline. The pretrained gate consistently reduces estimated latency, with speedups ranging from \(1.24\times\) on CommonsenseQA to \(1.58\times\) on ARC-Easy. Post-hoc readouts produce similar estimated latency reductions when they exit at comparable average loops. Thus, the average exit-loops metric used in the Pareto curves corresponds to a practical latency proxy for Ouro as well.

\section{Conclusion}

In this systematic evaluation, we have shown that adaptive depth in looped Transformers is best understood as a trajectory--readout problem, not a gate-learning problem. The training objective shapes the recurrent trajectory, while a separate readout decides when to stop. Learned halting gates conflate these roles, using one exit distribution to both weight training losses and stop at inference. Across controlled MANO and parity experiments, and large-scale Ouro-1.4B and 2.6B checkpoints, we find that simple post-hoc readouts on well-shaped trajectories match or outperform learned gates, and that gate failures trace to the induced trajectory rather than gate expressivity. Where prior work focuses on making recurrent depth dynamic, our contribution is diagnostic: isolating the roles of trajectory formation and exit selection in determining adaptive-compute performance. We hope this trajectory--readout perspective motivates future work on alternative pretraining objectives and looped Transformer architectures.




\bibliography{main}
\bibliographystyle{tmlr}

\appendix

\section{Model Background}
This appendix provides additional background on the architectures studied in this work. We first summarize the looped Transformer formulation used throughout the paper, then review the Ouro architecture and training procedure.

\subsection{Looped Transformer Architecture}
\label{app:looped_architecture_details}

A looped Transformer replaces a stack of independently parameterized Transformer layers with repeated applications of a shared recurrent block. Let \(h_0=E(x)\in\mathbb{R}^{M\times d}\) denote the embedded input sequence. A shared block \(\mathcal{F}_\theta:\mathbb{R}^{M\times d}\rightarrow\mathbb{R}^{M\times d}\), typically consisting of one or more Transformer layers, is then applied recurrently:

\[
h_t = \mathcal{F}_\theta(h_{t-1}), \qquad t=1,\ldots,T.
\]

Here \(T\) is the maximum number of recurrent loops, and the same parameters \(\theta\) are reused at every depth. Thus increasing \(T\) increases the amount of test-time computation without increasing the number of learned parameters. This parameter sharing distinguishes looped Transformers from standard depth scaling, where each additional layer introduces a new set of parameters.

The shared block can be written abstractly as a residual Transformer update,
\[
z_t = h_{t-1} + \mathrm{Attn}_\theta(\mathrm{LN}(h_{t-1})), 
\qquad
h_t = z_t + \mathrm{MLP}_\theta(\mathrm{LN}(z_t)),
\]

or as a short stack of such layers. In either case, the important property for this paper is that the same update map is iterated multiple times. The resulting sequence

\[
\tau_\theta(x) = (h_1,\ldots,h_T)
\]

is the recurrent trajectory. Applying the same prediction head to every recurrent state gives per-depth logits

\[
o_t = W_{\mathrm{out}} h_t,
\]

and corresponding per-depth losses \(\ell_t(x,y)\). These intermediate logits make it possible to evaluate the model at any recurrent loop, rather than only at the final loop.

Adaptive depth in this architecture amounts to choosing a stopping depth
\[
d(x)\in\{1,\ldots,T\}
\]

and returning the prediction from \(h_{d(x)}\). A fixed-depth baseline chooses the same \(d\) for every input. A learned gate predicts a distribution over stopping depths from the recurrent states. A post-hoc readout instead computes a stopping signal from quantities already present in the trajectory, such as entropy, maximum probability, logit margin, predictive change, or hidden-state displacement. All of these methods operate on the same sequence of states \((h_1,\ldots,h_T)\). They differ only in how the stopping depth is selected.

This architecture is especially useful for separating trajectory formation from exit selection. The training objective determines which intermediate states receive supervision and therefore which parts of the trajectory become predictive. The readout determines which of those states is used at inference time. Learned halting objectives couple these two roles because the gate distribution both weights the per-depth training losses and defines the inference-time stopping rule. Fixed-prior training removes the learned input-dependent gate from the training objective, allowing us to study how the trajectory is shaped independently of the adaptive readout.

In our MANO and parity experiments, we train looped Transformers from scratch with \(T=6\) recurrent loops and apply the same language-model head at every loop. In the Ouro experiments, we use pretrained looped language-model checkpoints and keep the recurrent trajectory fixed, varying only the readout used to select an exit. This common architectural view lets us compare learned gates, fixed-depth exits, and post-hoc trajectory readouts within the same trajectory--readout framework.

\subsection{The Ouro Model Family}
\label{app:ouro-background}

Ouro \citep{Ouro} implements a Looped Language Model by repeatedly applying a shared stack of $L$ standard decoder-only Transformer layers (multi-head attention with RoPE, SwiGLU feed-forward blocks, sandwich RMSNorm) up to $T_{\max}$ times, rather than stacking $L$ \emph{distinct} layers as in a conventional Transformer. The released checkpoints are Ouro-1.4B (24 layers) and Ouro-2.6B (48 layers, obtained by doubling and further training the 24-layer stack), both with hidden size 2048 and $T_{\max}=4$ trained recurrent steps, pre-trained on 7.7T tokens spanning web, math, code, and long-context data across four stages.

At each recurrent step $t$, a linear exit gate produces a halt probability $\lambda_t(x) = \sigma(\mathrm{Linear}_\phi(h^{(t)}))$, which induces a discrete exit distribution $p_\phi(t\mid x)$ over $\{1,\dots,T_{\max}\}$ in the same way as the gating mechanism in Section~\ref{sec:gating-mechanism} of this paper. Ouro's gate is trained in two stages. \textbf{Stage~I} jointly optimizes the gate and the backbone with an entropy-regularized objective,

\[
\mathcal{L} = \sum_{t=1}^{T_{\max}} p_\phi(t\mid x)\,\mathcal{L}^{(t)} - \beta\, H\big(p_\phi(\cdot\mid x)\big),
\]

which is equivalent to a KL penalty toward a \emph{uniform} prior over exit depths (chosen to avoid biasing the gate toward any particular depth before depth-dependent difficulty is learned). \textbf{Stage~II} then freezes the backbone and fine-tunes only the gate using a marginal-utility signal. For each adjacent pair of recurrent steps, a detached loss-improvement is calculated as follows:

\[ I_i^{(t)} = \max(0,\, \mathcal{L}^{(t-1)}_{i,\mathrm{stop}} - \mathcal{L}^{(t)}_{i,\mathrm{stop}}) 
\]

This loss improvement term is converted into a soft continue/exit label and used as a binary cross-entropy target for the gate. This Stage-II objective is the direct inspiration for the marginal-utility post-hoc gate we fit on frozen trajectories in Appendix~\ref{app:posthoc-gates}.

Ouro's authors report that the resulting 1.4B and 2.6B checkpoints match dense Transformer baselines 2--3$\times$ larger across general, math, and coding benchmarks, and that gains stem primarily from improved knowledge \emph{manipulation} rather than increased knowledge capacity. They also report that post-training RLVR attempts on Ouro provided limited gains, attributing this in part to the mismatch between fast fixed-depth rollout infrastructure (e.g., vLLM/SGLang) and the model's variable-depth, gate-driven execution path. This difficulty in optimizing around the gate at the RL stage is part of our motivation for asking, in the present work, whether the gate itself is the right object to optimize versus the trajectory it induces. We use the publicly released 1.4B and 2.6B base checkpoints without modifying or retraining them.

\section{Experimental Details}
\label{app:experimental_details}

This appendix summarizes dataset construction, model architecture, optimization settings, threshold selection, and benchmark protocols used in the experiments.

\subsection{MANO Task}
MANO consists of modular arithmetic expressions over integers modulo 23. Expressions are represented in prefix notation and contain up to 10 operations. We balance examples across operation-count bins and use operation count as the difficulty variable in trajectory diagnostics. Unlike MANO setups that provide the expression length as an input token, our implementation does not include a length token, in order to avoid giving the model a shortcut towards compute allocation.

\subsection{MANO Architecture}
All MANO experiments use the same looped Transformer backbone with 4 layers, 4 attention heads, hidden embedding size 512, block size 32, no dropout, and bias terms enabled. The model uses tied input embeddings and language-model head weights. The maximum recurrent depth is T=6. Learned exit gates map the recurrent hidden state to a scalar halt logit.

\subsection{MANO Training}
All MANO variants are trained with AdamW using learning rate $5\times10^{-4}$, $\beta=(0.9,0.98)$, weight decay 0.1, batch size 1024, and gradient clipping at 1.0. We use a warmup-cosine learning-rate schedule with 2000 warmup steps and minimum learning rate $5\times10^{-5}$. Models are trained for $40{,}000$ steps using bf16 mixed precision.

\subsection{MANO Variants}
Fixed-prior models use either a uniform prior or truncated geometric priors with $\lambda\in\{0.2,0.3,0.5,0.7\}$. Learned linear-gate models use either a uniform prior or the same geometric priors, and sweep the KL regularization strength $\beta$. For the uniform prior, we sweep $\beta\in\{0.001,0.005,0.01,0.02,0.05,0.1,0.2,0.5\}$. For geometric priors, we sweep $\beta\in\{0.01,0.05,0.1\}$. MLP-gate models replace the linear halt head with a two-layer feedforward gate. In the case of the MLP gate, we sweep $\beta\in\{0.05, 0.1\}$ for the geometric prior, while for the uniform prior we use $\beta\in\{0.001, 0.01, 0.1, 0.2\}$.

\subsection{Threshold Selection}
For each post-hoc readout, we construct a 51-point threshold grid from validation-set readout values using empirical quantiles between the 0.5th and 99.5th percentiles. The resulting grid is fixed before evaluation and then applied to the test set. Learned-gate readouts use a fixed data-independent grid of cumulative probability thresholds $\alpha\in[0.01,1]$. Fixed-depth baselines require no threshold selection.

\subsection{Post-Hoc Readout Definitions}
\label{app:readout_definitions}

Let \(z_t(x)\) denote the task logits at recurrent loop \(t\),
\(p_t(x)=\mathrm{softmax}(z_t(x))\), and let \(r_t(x)\) denote the pooled hidden
representation used for hidden-state readouts. For confidence readouts, we use
\[
\begin{aligned}
H_t(x) &= -\sum_k p_t(k\mid x)\log p_t(k\mid x),\\
C_t(x) &= \max_k p_t(k\mid x),\\
M_t(x) &= z_{t,(1)}(x)-z_{t,(2)}(x),
\end{aligned}
\]
where \(z_{t,(1)}(x)\) and \(z_{t,(2)}(x)\) are the largest and second-largest
logits at loop \(t\). For convergence readouts, we use
\[
\begin{aligned}
\Delta_{\mathrm{KL}}(t;x)
&=D_{\mathrm{KL}}\!\left(p_t(x)\,\|\,p_{t-1}(x)\right),\\
\Delta_{\mathrm{logit}}(t;x)
&=\|z_t(x)-z_{t-1}(x)\|_2,\\
\Delta_{\mathrm{hid}}(t;x)
&=\|r_t(x)-r_{t-1}(x)\|_2.
\end{aligned}
\]
A confidence readout exits once the model is sufficiently confident. For top-1
confidence and logit margin, we use
\(d_\gamma(x)=\min\{t:S_t(x)\ge\gamma\}\), where
\(S_t\in\{C_t,M_t\}\). For entropy, where lower values indicate higher
confidence, we instead use
\(d_\gamma^{H}(x)=\min\{t:H_t(x)\le\gamma\}\).

In contrast to confidence readouts, a convergence readout exits once the corresponding change signal falls below a
threshold: \(d_\epsilon(x)=\min\{t\ge 2:\Delta(t;x)\le\epsilon\}\). All readouts
default to \(T\) if the threshold is never reached.

\subsection{Ouro Models and Benchmarks}
We evaluate ByteDance Ouro-1.4B and Ouro-2.6B without further training. We use six standard benchmarks: MMLU with 5-shot prompting, ARC-Challenge with 25-shot prompting, ARC-Easy with 8-shot prompting, OpenBookQA with 0-shot prompting, CommonsenseQA with 10-shot prompting, and HellaSwag with 10-shot prompting. 

\subsection{Ouro Threshold Selection}
For benchmarks with labeled test sets, heuristic threshold grids are selected on the validation split and evaluated on the test split. For CommonsenseQA and HellaSwag, where labeled test sets are unavailable, we split the labeled validation set into a $20\%$ validation subset for threshold selection and an $80\%$ held-out test subset for evaluation using a fixed random seed. As in MANO, gate thresholds use a fixed data-independent grid over cumulative gate probability, and fixed-depth baselines require no threshold selection.

\section{Fixed-Prior Trajectory Analysis}
\label{app:fixed_prior_difficulty}

This appendix collects the full fixed-prior analyses supporting Section~\ref{sec:results_fixed_prior}. We report the complete geometric-prior Pareto sweep across \(\lambda\) in Figure~\ref{fig:appendix_fixed_prior_pareto_sweep}, per-difficulty trajectory diagnostics for the uniform and geometric priors in Appendices~\ref{app:uniform-convergence} and~\ref{app:geometric-convergence}, and the difficulty extrapolation study in Appendix~\ref{app:mano_difficulty_extrapolation}. We use the number of operations as the difficulty variable and evaluate every recurrent loop with forced exits, so the reported signals reflect the trajectory itself rather than any adaptive stopping rule.

\subsection{\texorpdfstring
    {Pareto Curves for the Geometric Prior over Multiple $\lambda$ Values}
    {Pareto Curves for the Geometric Prior over Multiple Lambda Values}}
    
\begin{figure}[htbp]
    \centering
    \begin{subfigure}{0.45\linewidth}
        \centering
        \includegraphics[width=\linewidth]{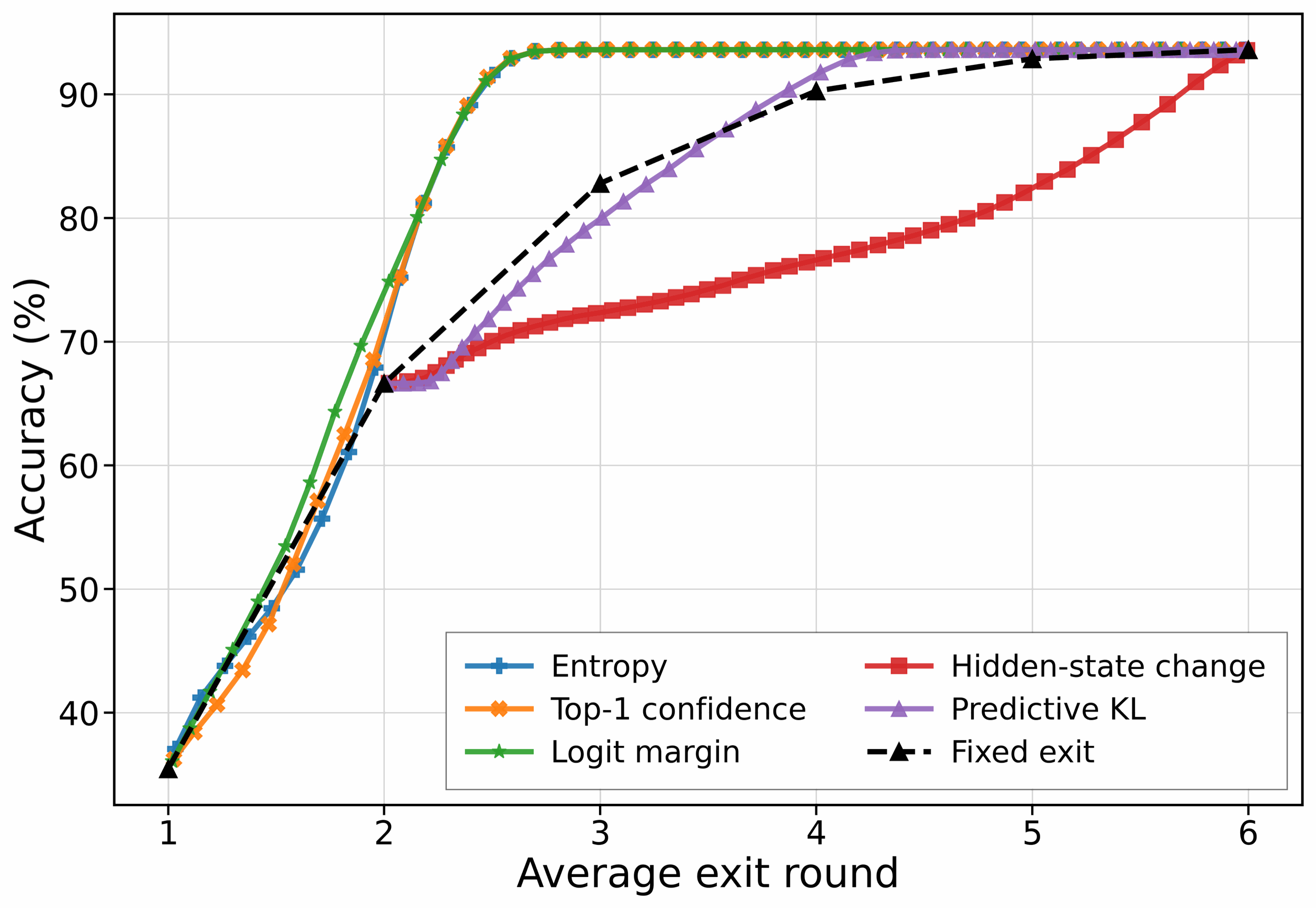}
        \caption{$\lambda=0.2$}
    \end{subfigure}
    \hfill
    \begin{subfigure}{0.45\linewidth}
        \centering
        \includegraphics[width=\linewidth]{images/fixed_prior/pareto/geom_fixed_lambd03_threshold_pareto_accuracy.png}
        \caption{$\lambda=0.3$}
    \end{subfigure}

    \vspace{0.5em}

    \begin{subfigure}{0.45\linewidth}
        \centering
        \includegraphics[width=\linewidth]{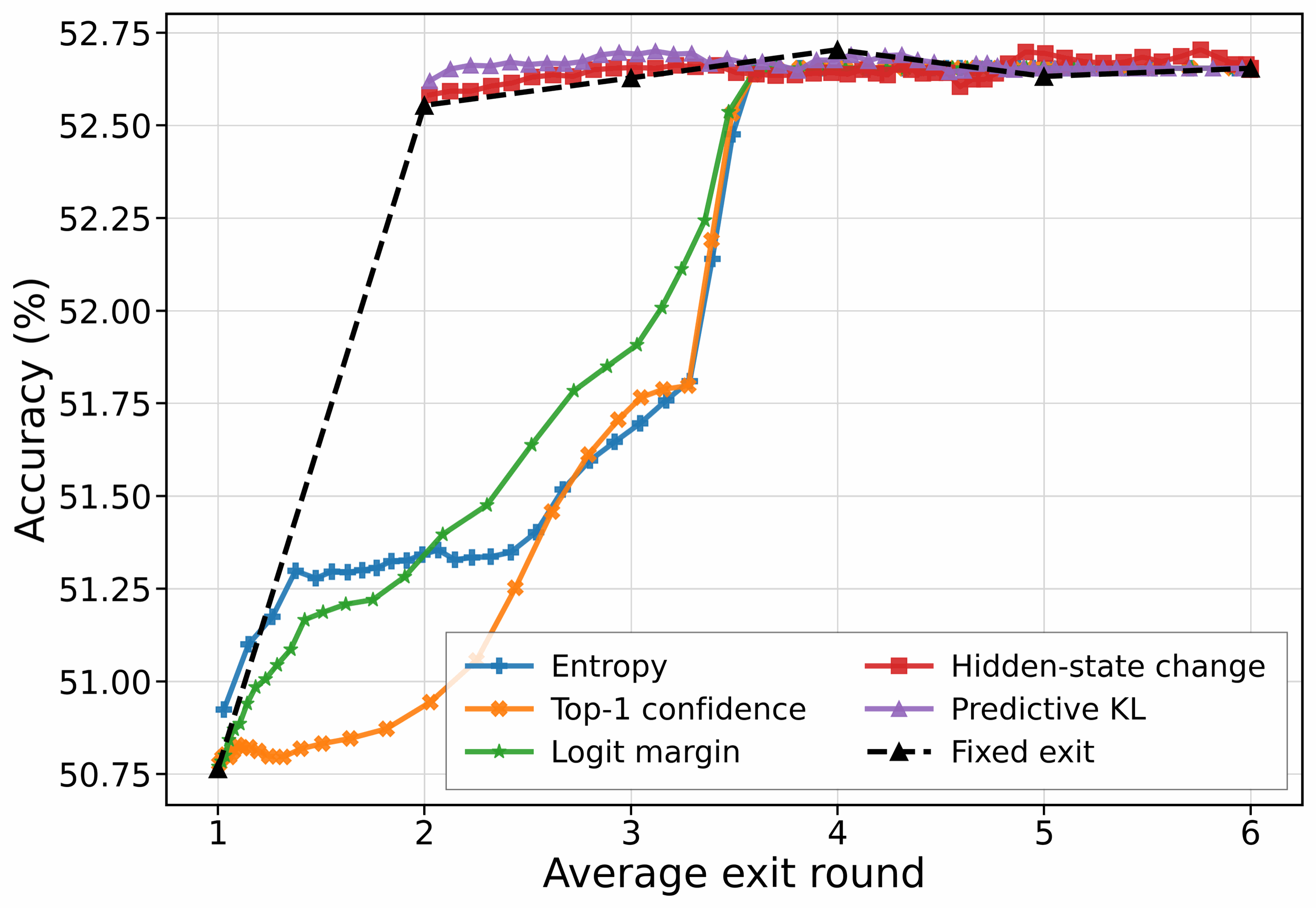}
        \caption{$\lambda=0.5$}
    \end{subfigure}
    \hfill
    \begin{subfigure}{0.45\linewidth}
        \centering
        \includegraphics[width=\linewidth]{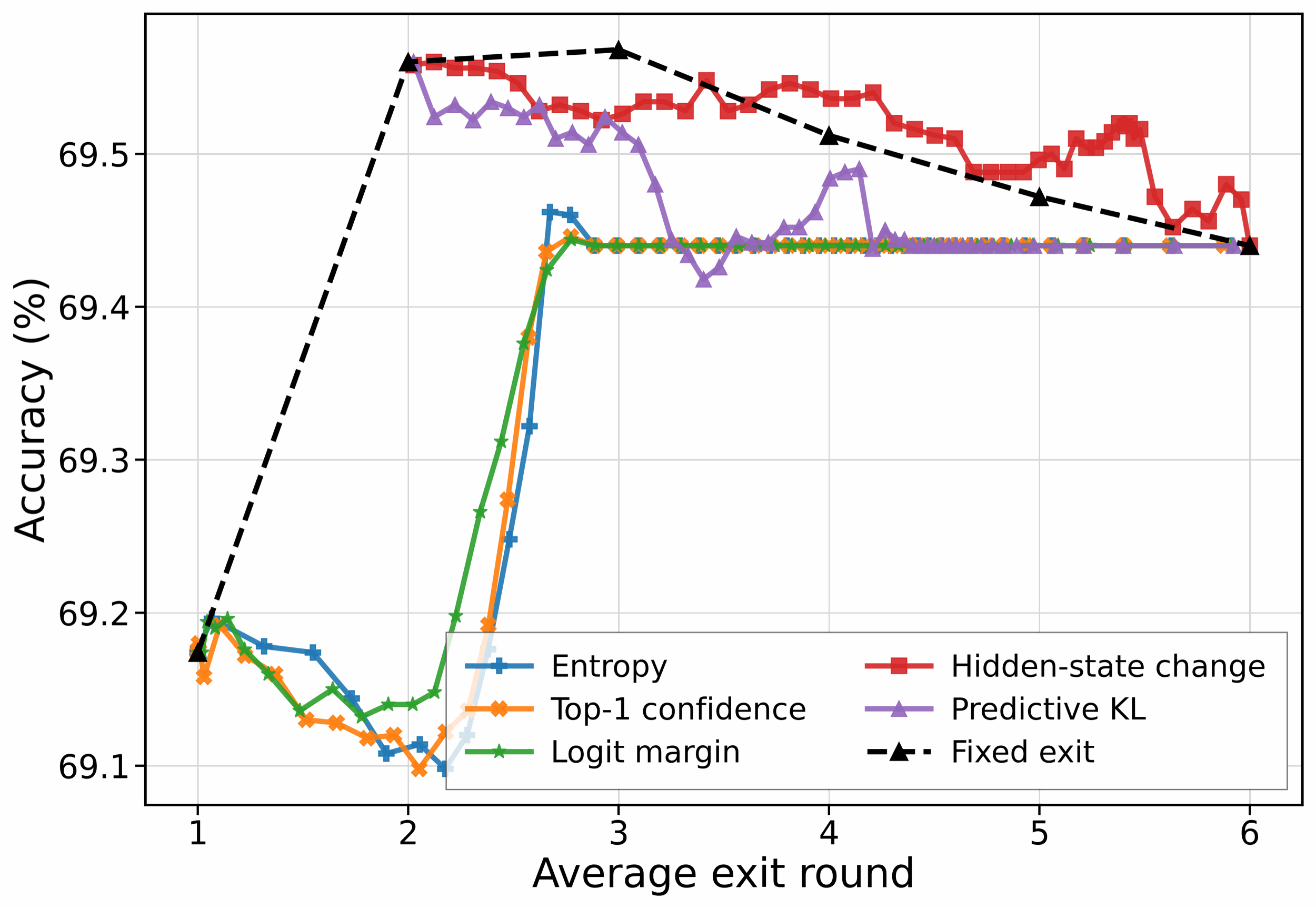}
        \caption{$\lambda=0.7$}
    \end{subfigure}

    \caption{
    Full geometric fixed-prior sweep.
    Each panel shows threshold-swept post-hoc readouts for a different truncated geometric prior.
    Moderate early-depth weighting yields useful early exits, while overly aggressive early weighting can degrade the trajectory.
    }
    \label{fig:appendix_fixed_prior_pareto_sweep}
\end{figure}

Figure~\ref{fig:appendix_fixed_prior_pareto_sweep} sweeps the truncated geometric prior across $\lambda \in \{0.2, 0.3, 0.5, 0.7\}$, plotting threshold-swept post-hoc readouts for each prior. The sweep reveals a clear optimum in early-depth weighting. Moderate values, and $\lambda = 0.3$ in particular, push useful predictions early in the trajectory and show a strong low-depth frontier. As $\lambda$ grows, the prior concentrates almost all loss weight on the first loop, which degrades the later recurrent states and flattens the accuracy ceiling, so the trajectory no longer benefits from additional computation. The uniform prior sits at the opposite extreme, supervising all depths equally and producing useful early exits only at higher average depth. This is consistent with the main-text finding that trajectory formation, not just the readout, shapes the achievable compute--quality tradeoff.

\subsection{Uniform Prior Difficulty Convergence}
\label{app:uniform-convergence}
\begin{figure}[!htbp]
    \centering
    \begin{subfigure}{0.45\linewidth}
        \centering
        \includegraphics[width=\linewidth]{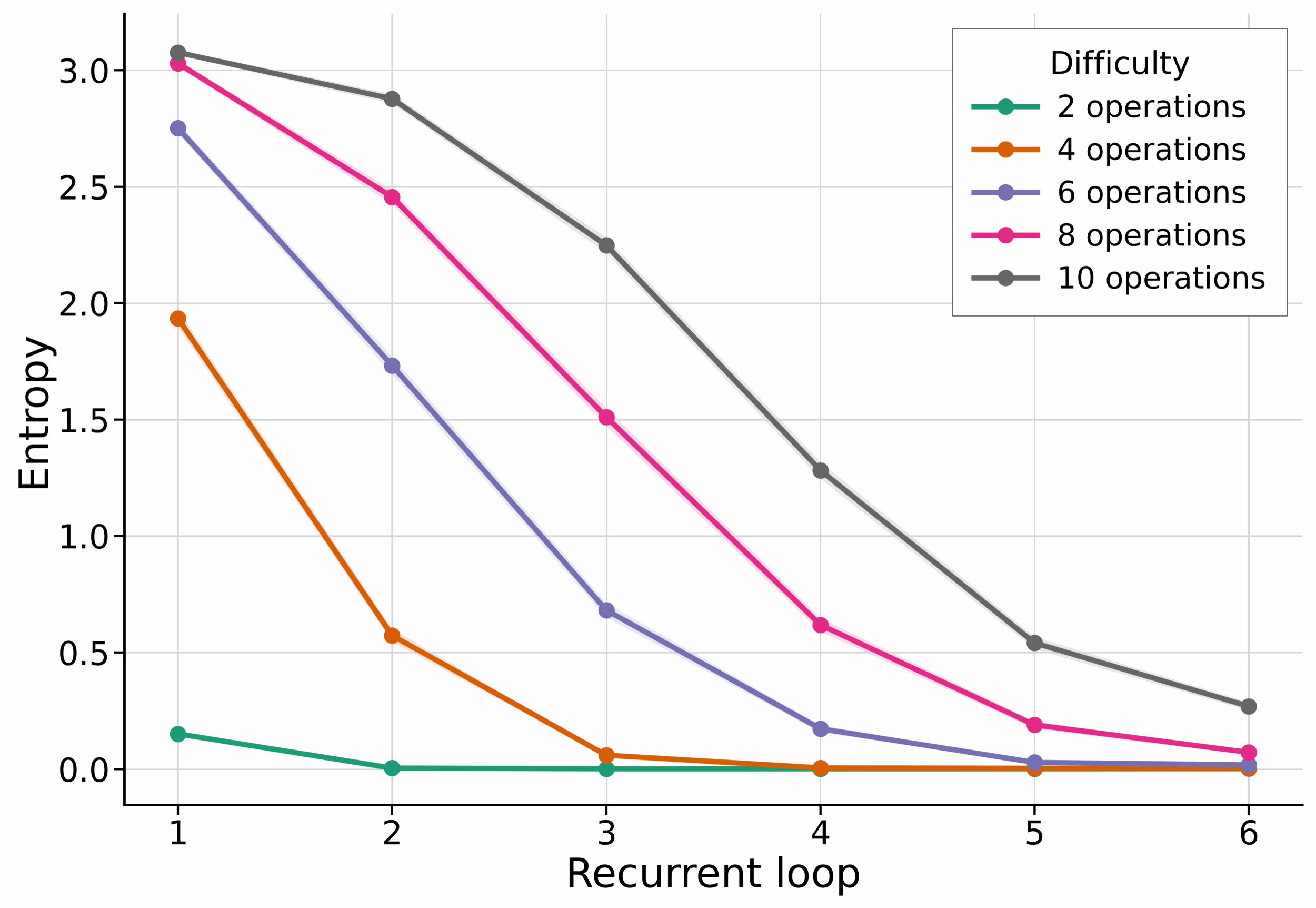}
        \caption{Entropy}
        \label{fig:app_uniform_entropy_by_ops}
    \end{subfigure}
    \hfill
    \begin{subfigure}{0.45\linewidth}
        \centering
        \includegraphics[width=\linewidth]{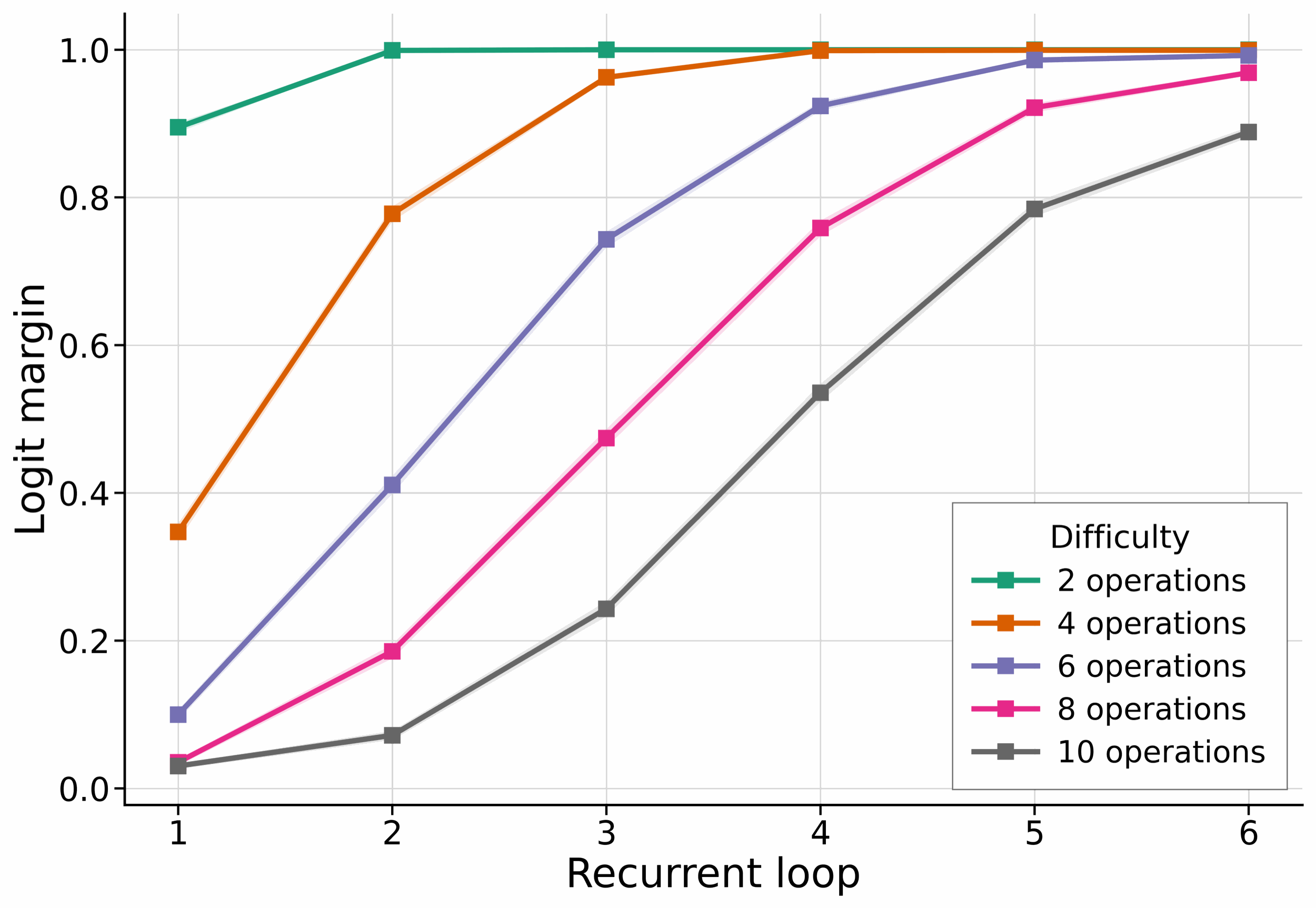}
        \caption{Logit margin}
        \label{fig:app_uniform_logit_margin_by_ops}
    \end{subfigure}

    \vspace{0.5em}

    \begin{subfigure}{0.45\linewidth}
        \centering
        \includegraphics[width=\linewidth]{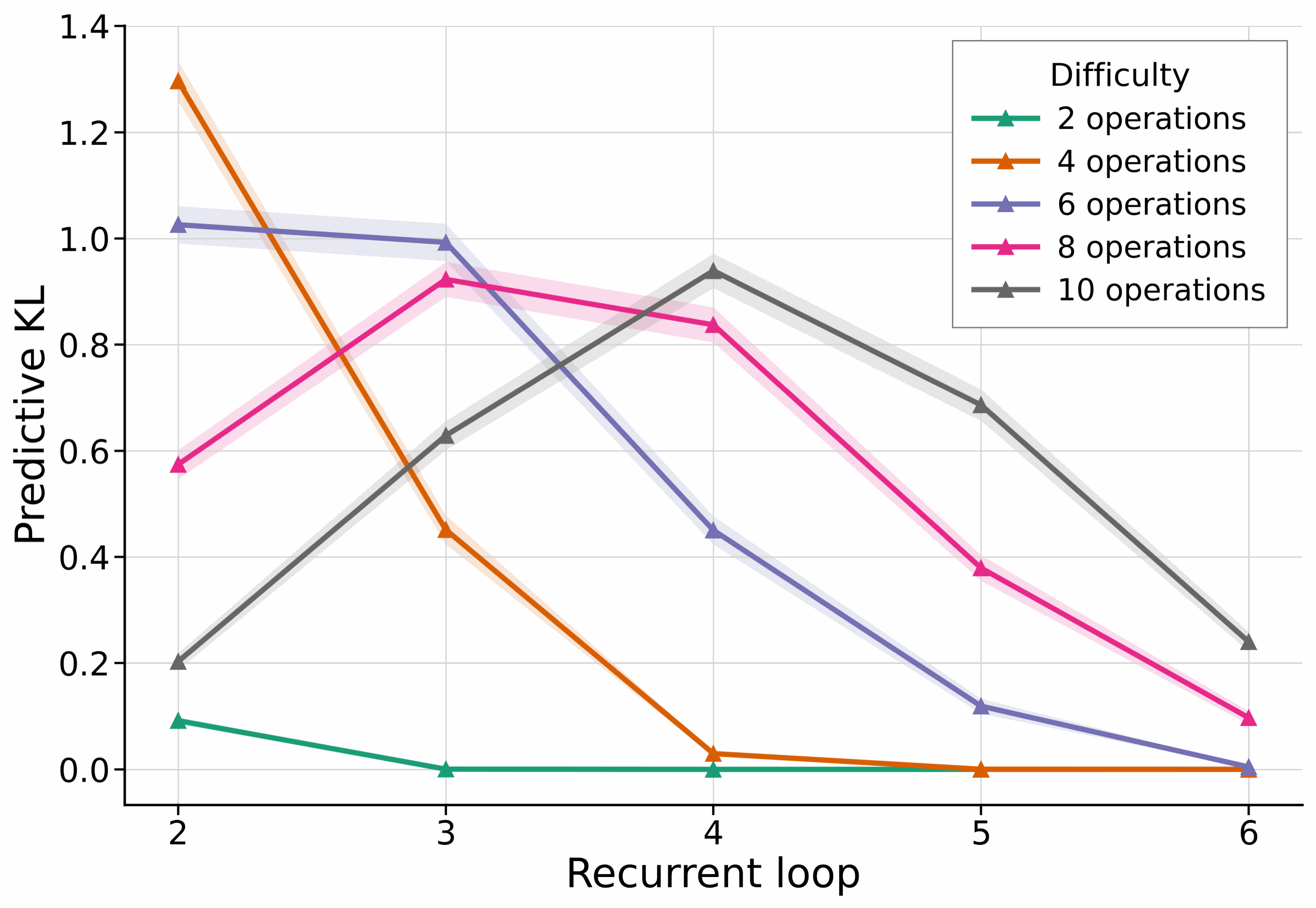}
        \caption{Predictive KL}
        \label{fig:app_uniform_predictive_kl_by_ops}
    \end{subfigure}
    \hfill
    \begin{subfigure}{0.45\linewidth}
        \centering
        \includegraphics[width=\linewidth]{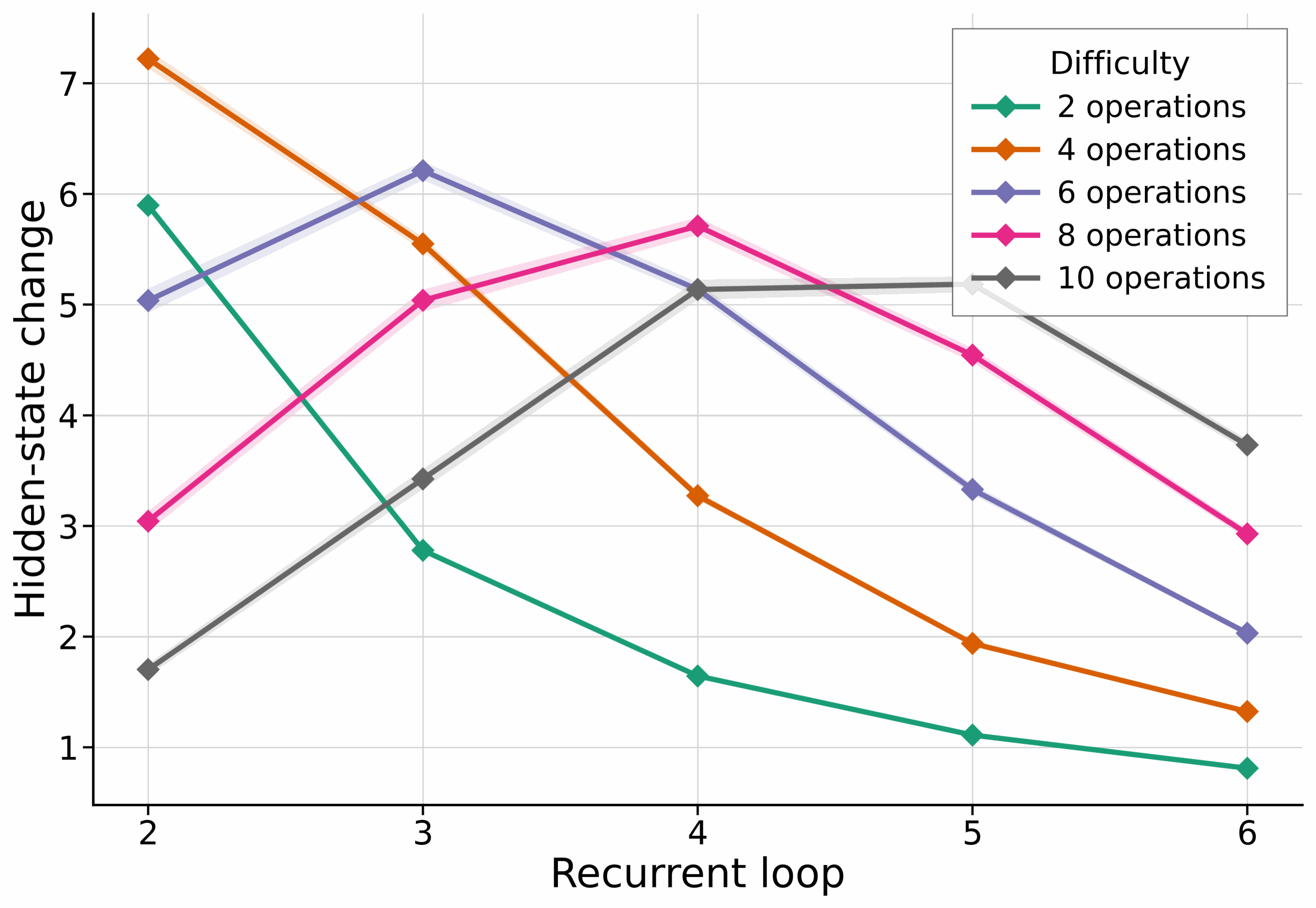}
        \caption{Hidden-state displacement}
        \label{fig:app_uniform_hidden_delta_by_ops}
    \end{subfigure}

    \caption{
    Difficulty-stratified trajectory diagnostics for the uniform fixed-prior model.
    Examples are grouped by the number of operations required by the task, and each panel plots a trajectory signal across recurrent loops.
    Prediction-space signals such as entropy, logit margin, and predictive KL show more legible difficulty structure than hidden-state displacement.
    }
    \label{fig:app_uniform_convergence_by_ops}
\end{figure}

Figure~\ref{fig:app_uniform_convergence_by_ops} shows per-difficulty trajectory diagnostics for the uniform fixed-prior model, grouping examples by the number of operations in the input sequence. Prediction-space signals carry a clean difficulty ordering. Entropy decreases and logit margin increases with recurrent depth, and at every loop both quantities separate by operation count, with harder expressions remaining higher-entropy and lower-margin for more loops. Predictive KL between consecutive predictions follows the same trend, though with more overlap between adjacent difficulty bins. Hidden-state displacement is noisier. Harder examples tend to keep moving in representation space for longer, but the separation by operation count is weaker and less monotonic than in prediction space.

\begin{table}[!htbp]
\centering
\caption{
Difficulty ordering of the uniform fixed-prior MANO trajectory across three seeds.
For each recurrent loop, we compute a signed Spearman correlation between operation count and the trajectory signal, then average across loops and seeds.
The sign is chosen so that positive values mean harder examples appear harder: higher entropy, lower logit margin, larger predictive KL, or larger hidden-state displacement.
}
\label{tab:uniform_difficulty_ordering}
\begin{tabular}{lc}
\toprule
Signal & Signed Spearman $\rho$ \\
\midrule
Entropy & $1.00 \pm 0.00$ \\
Logit margin & $1.00 \pm 0.00$ \\
Predictive KL & $0.80 \pm 0.11$ \\
Hidden-state displacement & $0.35 \pm 0.48$ \\
\bottomrule
\end{tabular}
\end{table}

Table~\ref{tab:uniform_difficulty_ordering} summarizes this ordering across seeds using signed Spearman correlations.

\subsection{Geometric Prior Difficulty Convergence}
\label{app:geometric-convergence}
\begin{figure}[htbp]
    \centering
    \begin{subfigure}{0.48\linewidth}
        \centering
        \includegraphics[width=\linewidth]{images/fixed_prior/convergence/geom_fixed_lambd03/entropy_by_num_ops.png}
        \caption{Entropy}
        \label{fig:app_geom03_entropy_by_ops}
    \end{subfigure}
    \hfill
    \begin{subfigure}{0.48\linewidth}
        \centering
        \includegraphics[width=\linewidth]{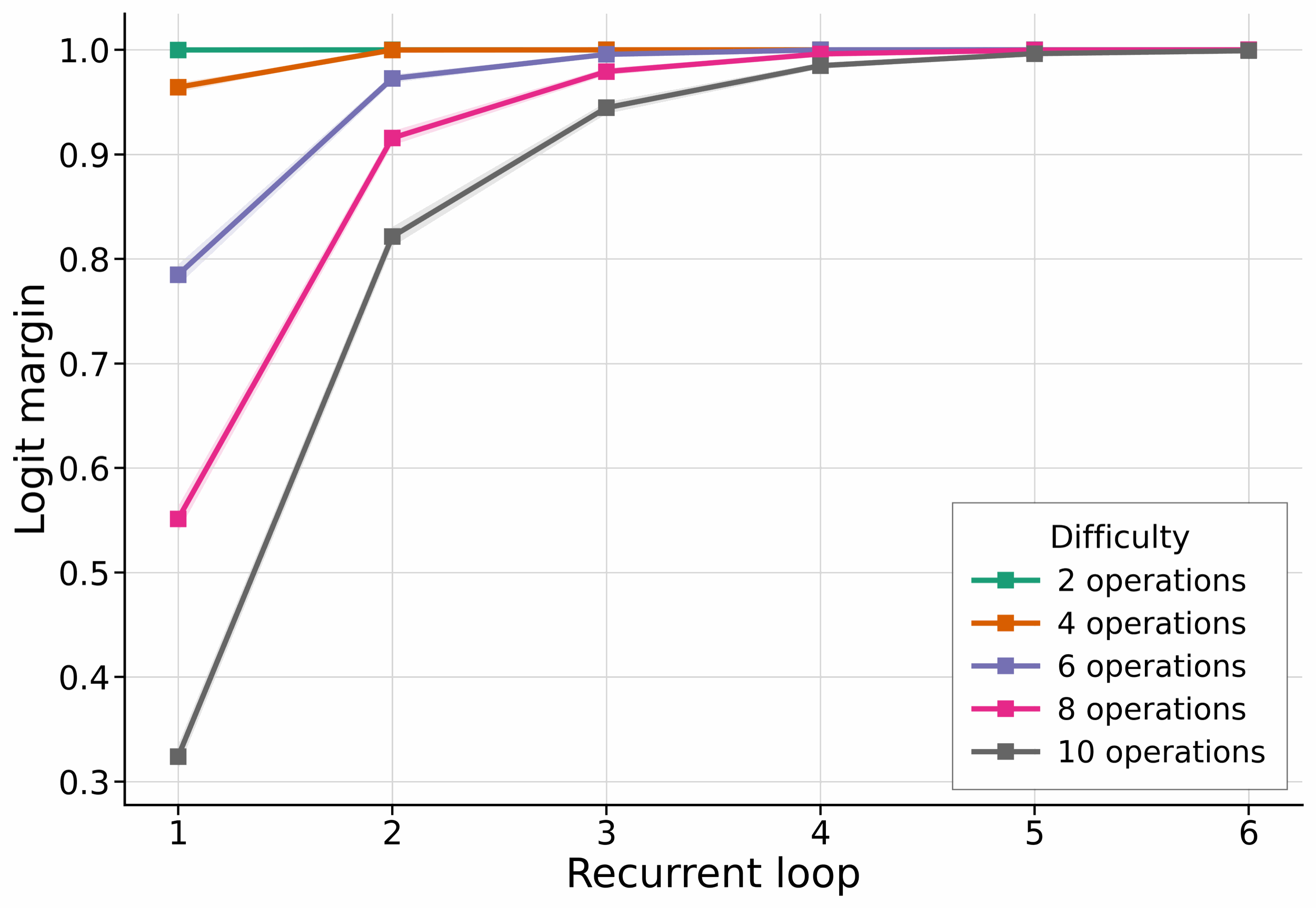}
        \caption{Logit margin}
        \label{fig:app_geom03_logit_margin_by_ops}
    \end{subfigure}

    \vspace{0.5em}

    \begin{subfigure}{0.48\linewidth}
        \centering
        \includegraphics[width=\linewidth]{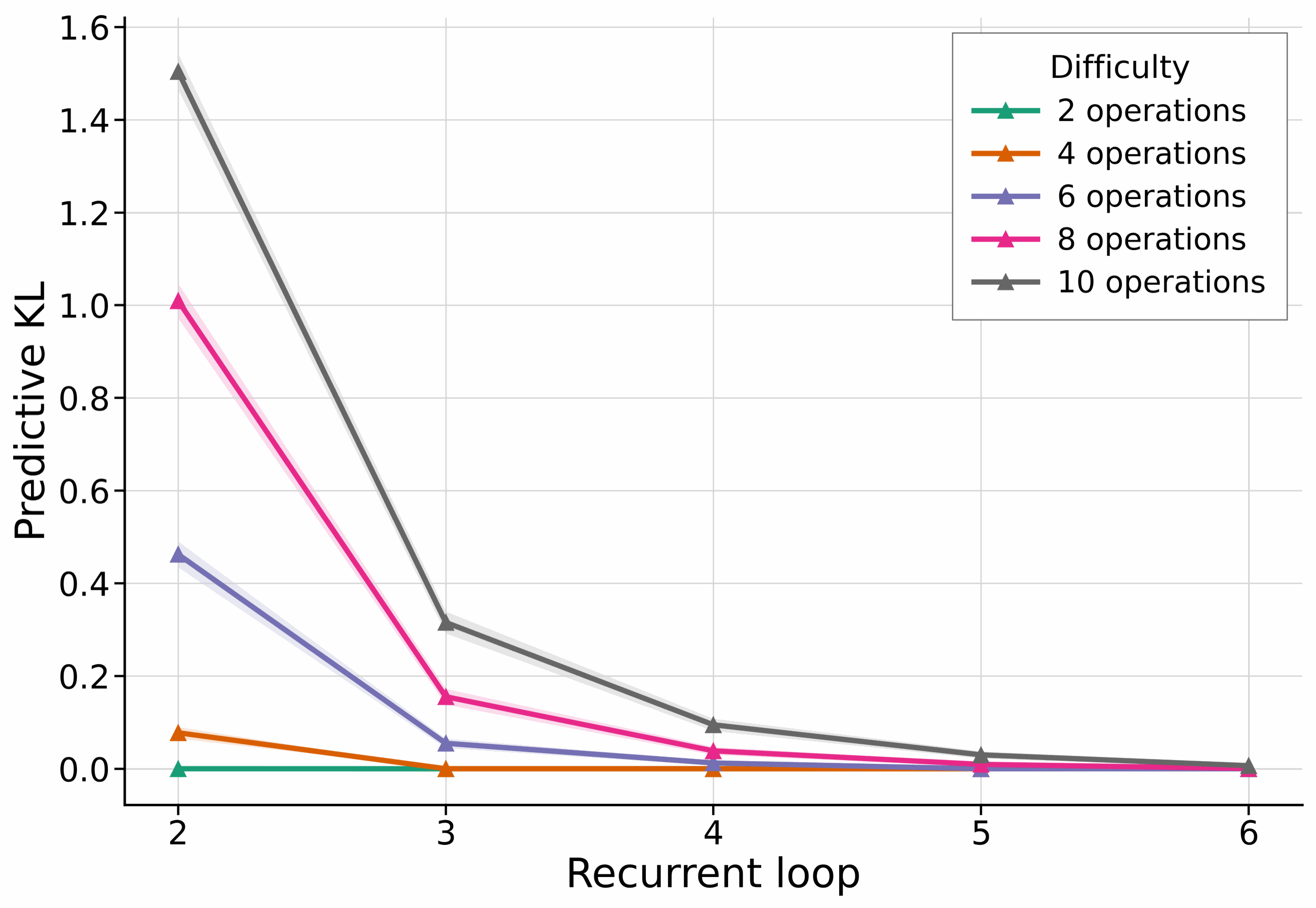}
        \caption{Predictive KL}
        \label{fig:app_geom03_predictive_kl_by_ops}
    \end{subfigure}
    \hfill
    \begin{subfigure}{0.48\linewidth}
        \centering
        \includegraphics[width=\linewidth]{images/fixed_prior/convergence/geom_fixed_lambd03/hidden_delta_by_num_ops.png}
        \caption{Hidden-state displacement}
        \label{fig:app_geom03_hidden_delta_by_ops}
    \end{subfigure}

    \caption{
    Difficulty-stratified trajectory diagnostics for the geometric fixed-prior model with $\lambda=0.3$.
    Compared with the uniform prior, the geometric prior shifts useful prediction-space signals earlier in the recurrent trajectory.
    Easier examples reach confident and stable states earlier, while harder examples continue changing for more loops.
    }
    \label{fig:app_geom03_convergence_by_ops}
\end{figure}

Figure~\ref{fig:app_geom03_convergence_by_ops} repeats the diagnostic for the geometric fixed-prior model with $\lambda = 0.3$. Relative to the uniform prior, the geometric prior concentrates supervision on earlier loops and shifts the useful prediction-space signals earlier in the trajectory. Easier examples reach confident, low-entropy and high-margin states within the first one or two loops, while harder examples continue to refine for several more loops. The difficulty ordering is preserved across all four signals, and hidden-state displacement is visibly less noisy than under the uniform prior. Table~\ref{tab:geom_difficulty_ordering} confirms this numerically. Entropy and logit margin remain near-perfectly ordered ($\rho = 0.98$), predictive KL improves to $\rho = 0.87$, and hidden-state displacement rises to $\rho = 0.69 \pm 0.34$, roughly double the uniform-prior value.

\begin{table}[htbp]
\centering
\footnotesize
\caption{
Difficulty ordering of the geometric fixed-prior MANO trajectory ($\lambda=0.3$) across three seeds.
Conventions follow Table~\ref{tab:uniform_difficulty_ordering}: positive signed Spearman correlation means harder examples appear harder.
}
\label{tab:geom_difficulty_ordering}
\begin{tabular}{lc}
\toprule
Signal & Signed Spearman $\rho$ \\
\midrule
Entropy & $0.98 \pm 0.03$ \\
Logit margin & $0.98 \pm 0.03$ \\
Predictive KL & $0.87 \pm 0.20$ \\
Hidden-state displacement & $0.69 \pm 0.34$ \\
\bottomrule
\end{tabular}
\end{table}

\subsection{Difficulty Extrapolation on MANO}
\label{app:mano_difficulty_extrapolation}

\begin{figure}[htbp]
\centering

\begin{subfigure}[t]{0.32\linewidth}
    \centering
    \includegraphics[
        width=\linewidth,
        trim={0.10in 0.05in 0.10in 0.08in},
        clip
    ]{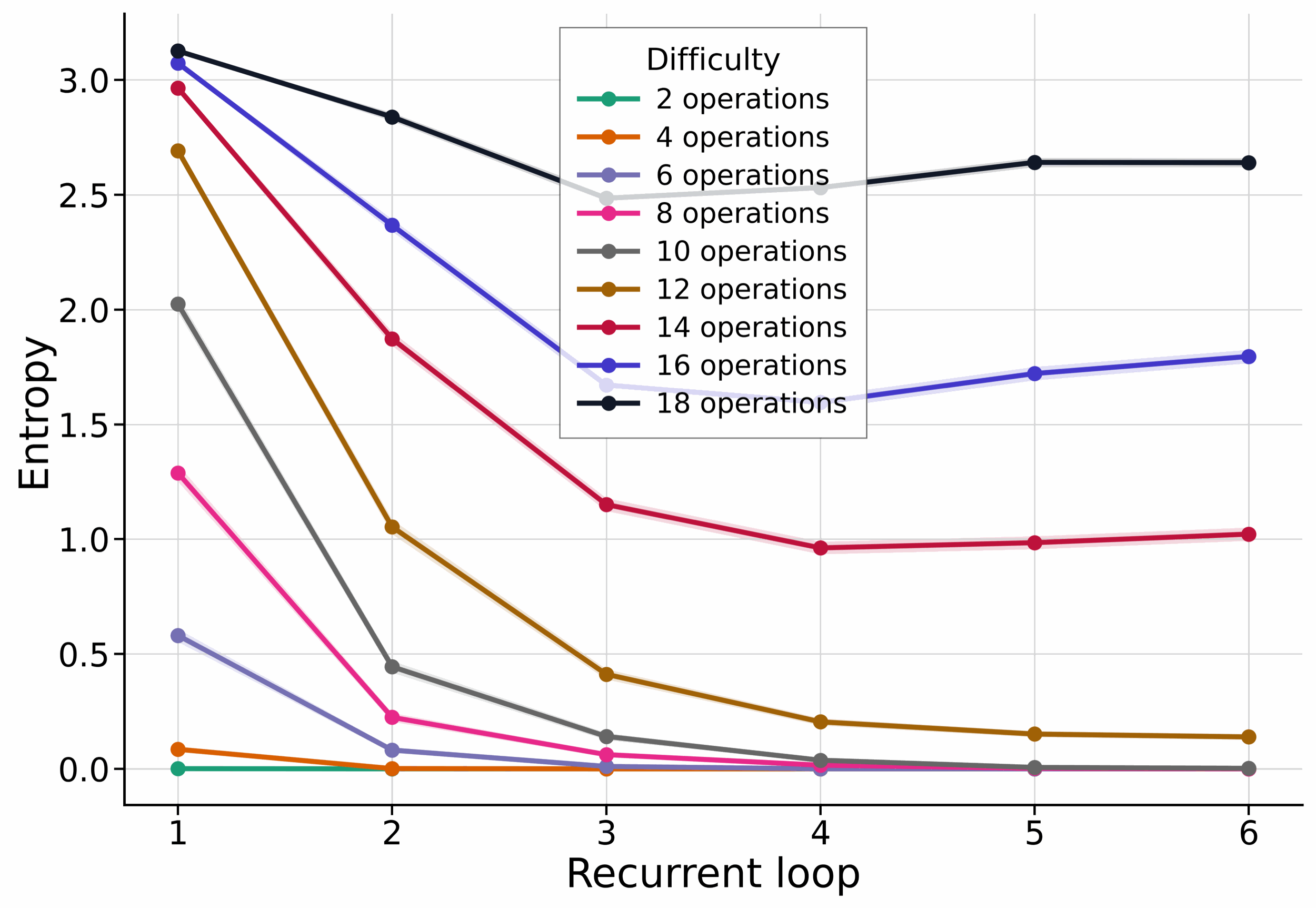}
    \caption{Entropy.}
\end{subfigure}
\hfill
\begin{subfigure}[t]{0.32\linewidth}
    \centering
    \includegraphics[
        width=\linewidth,
        trim={0.10in 0.05in 0.10in 0.08in},
        clip
    ]{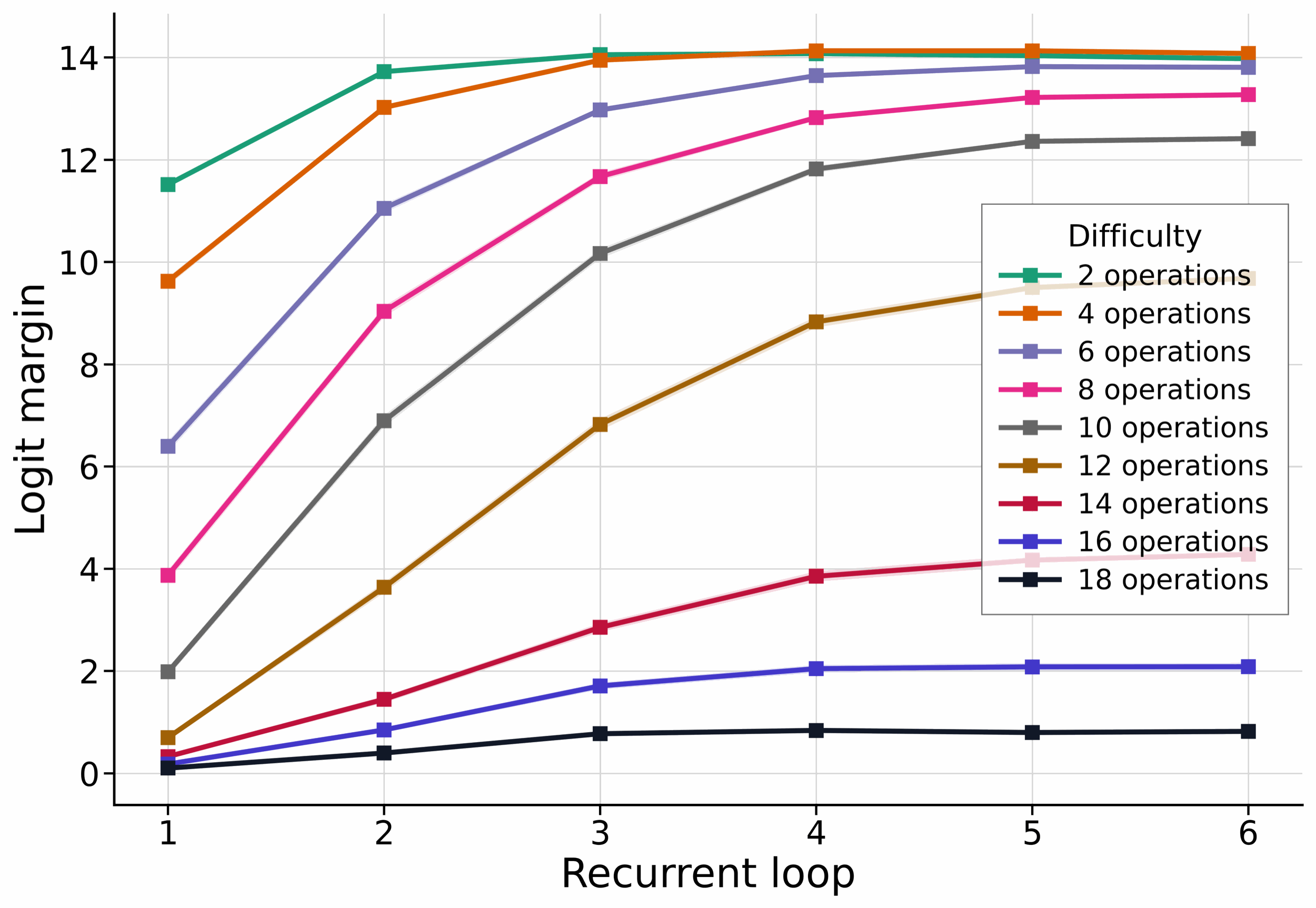}
    \caption{Logit margin.}
\end{subfigure}
\hfill
\begin{subfigure}[t]{0.32\linewidth}
    \centering
    \includegraphics[
        width=\linewidth,
        trim={0.10in 0.05in 0.10in 0.08in},
        clip
    ]{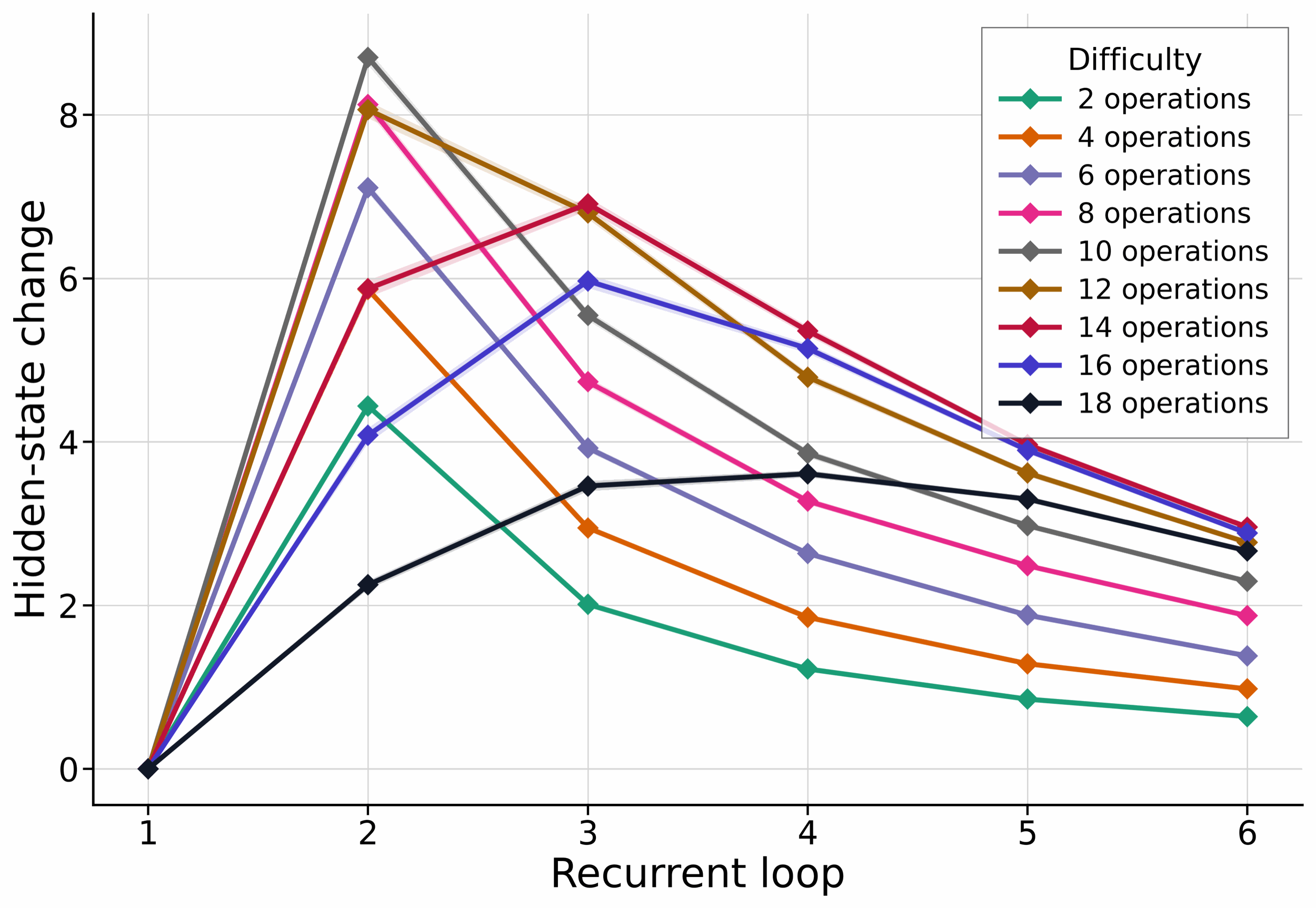}
    \caption{Hidden-state displacement.}
\end{subfigure}

\caption{
Difficulty extrapolation diagnostics for the geometric fixed-prior MANO model with \(\lambda=0.3\).
The model is trained on expressions with at most 10 operations and evaluated on harder expressions beyond that range.
Curves are grouped by operation count.
Harder expressions remain uncertain for more recurrent loops, develop logit margin more slowly, and show larger hidden-state changes across depth.
}
\label{fig:mano_difficulty_extrapolation_geom}
\end{figure}

\begin{figure}[htbp]
\centering

\begin{subfigure}[t]{0.32\linewidth}
    \centering
    \includegraphics[
        width=\linewidth,
        trim={0.10in 0.05in 0.10in 0.08in},
        clip
    ]{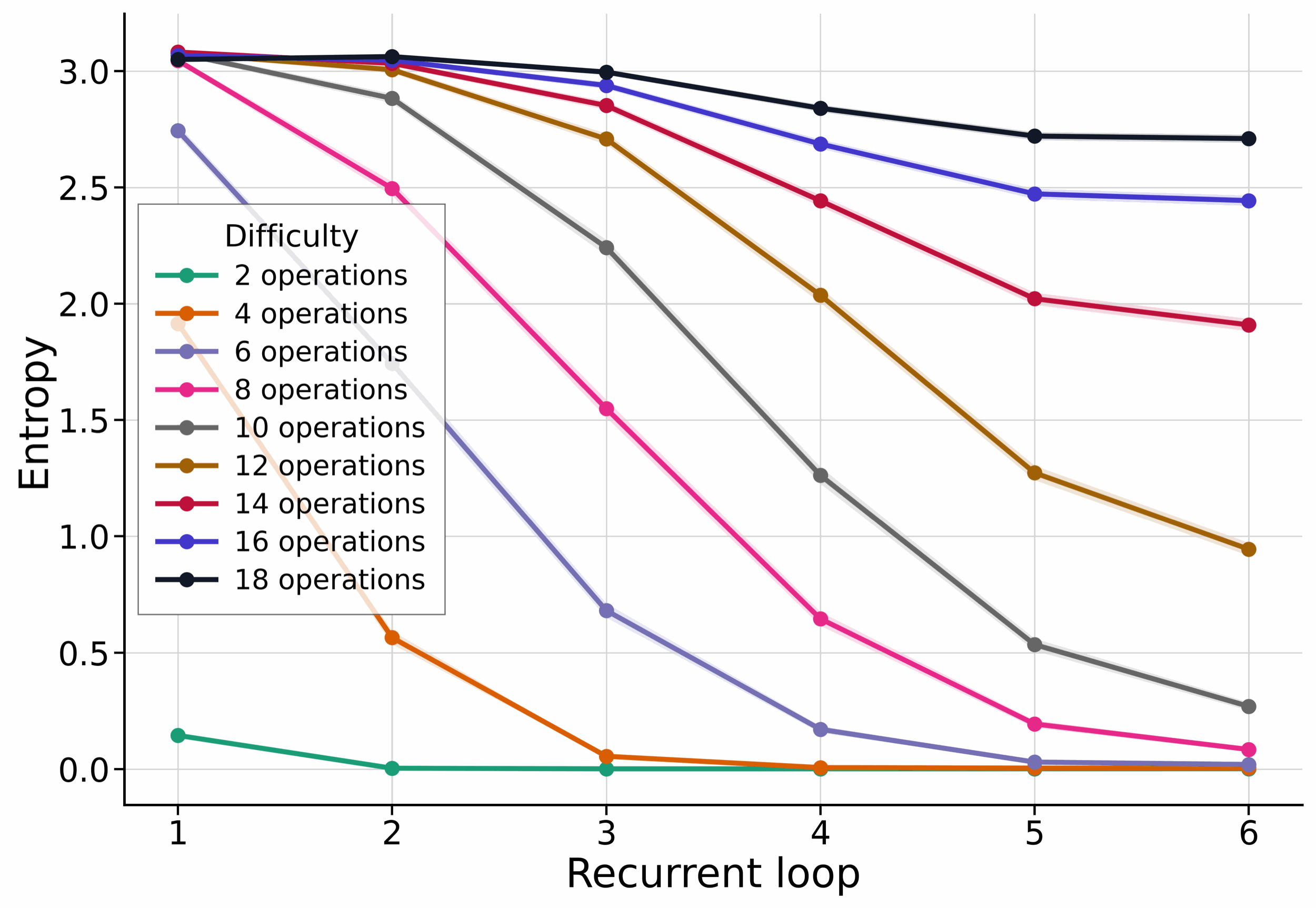}
    \caption{Entropy.}
\end{subfigure}
\hfill
\begin{subfigure}[t]{0.32\linewidth}
    \centering
    \includegraphics[
        width=\linewidth,
        trim={0.10in 0.05in 0.10in 0.08in},
        clip
    ]{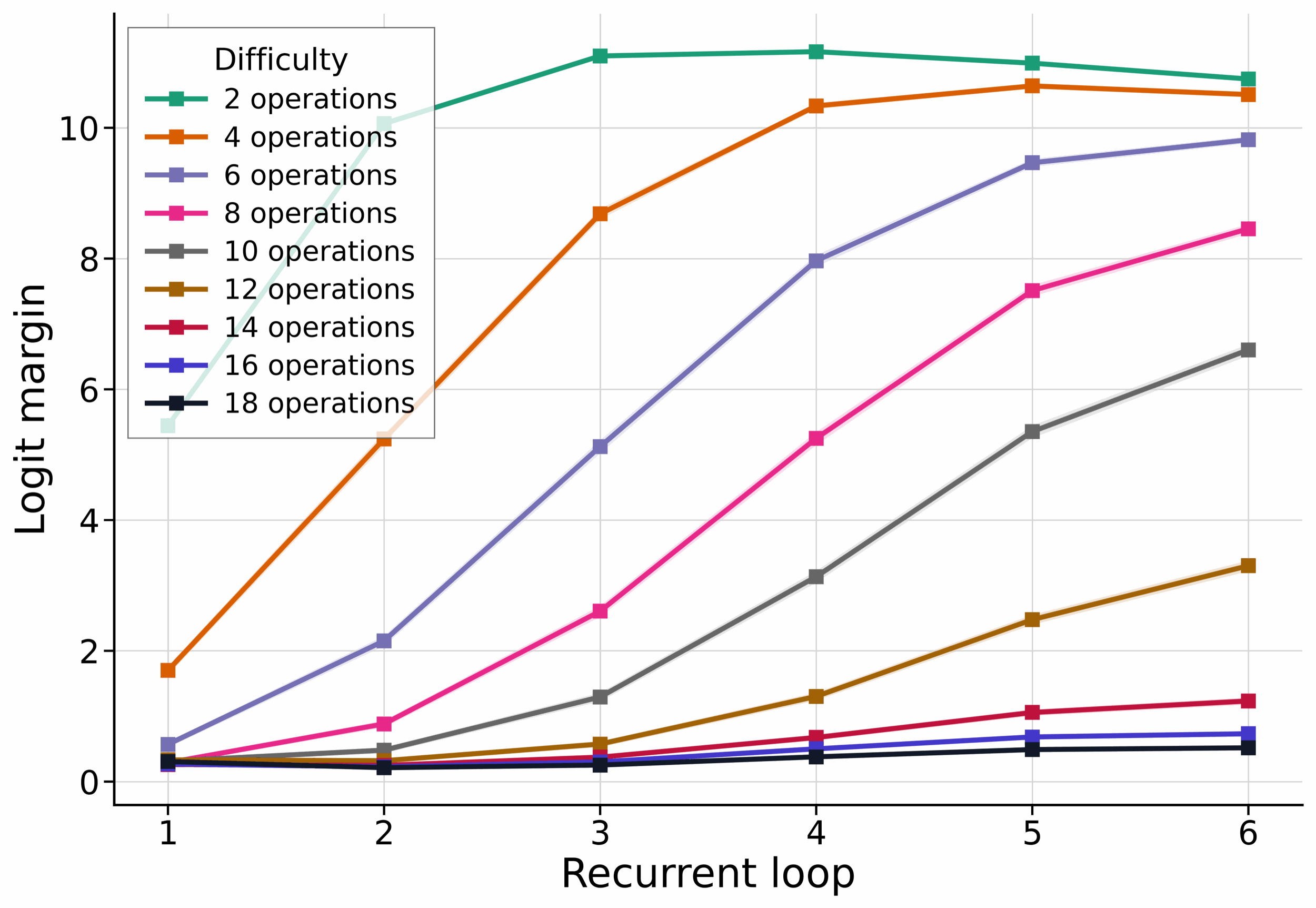}
    \caption{Logit margin.}
\end{subfigure}
\hfill
\begin{subfigure}[t]{0.32\linewidth}
    \centering
    \includegraphics[
        width=\linewidth,
        trim={0.10in 0.05in 0.10in 0.08in},
        clip
    ]{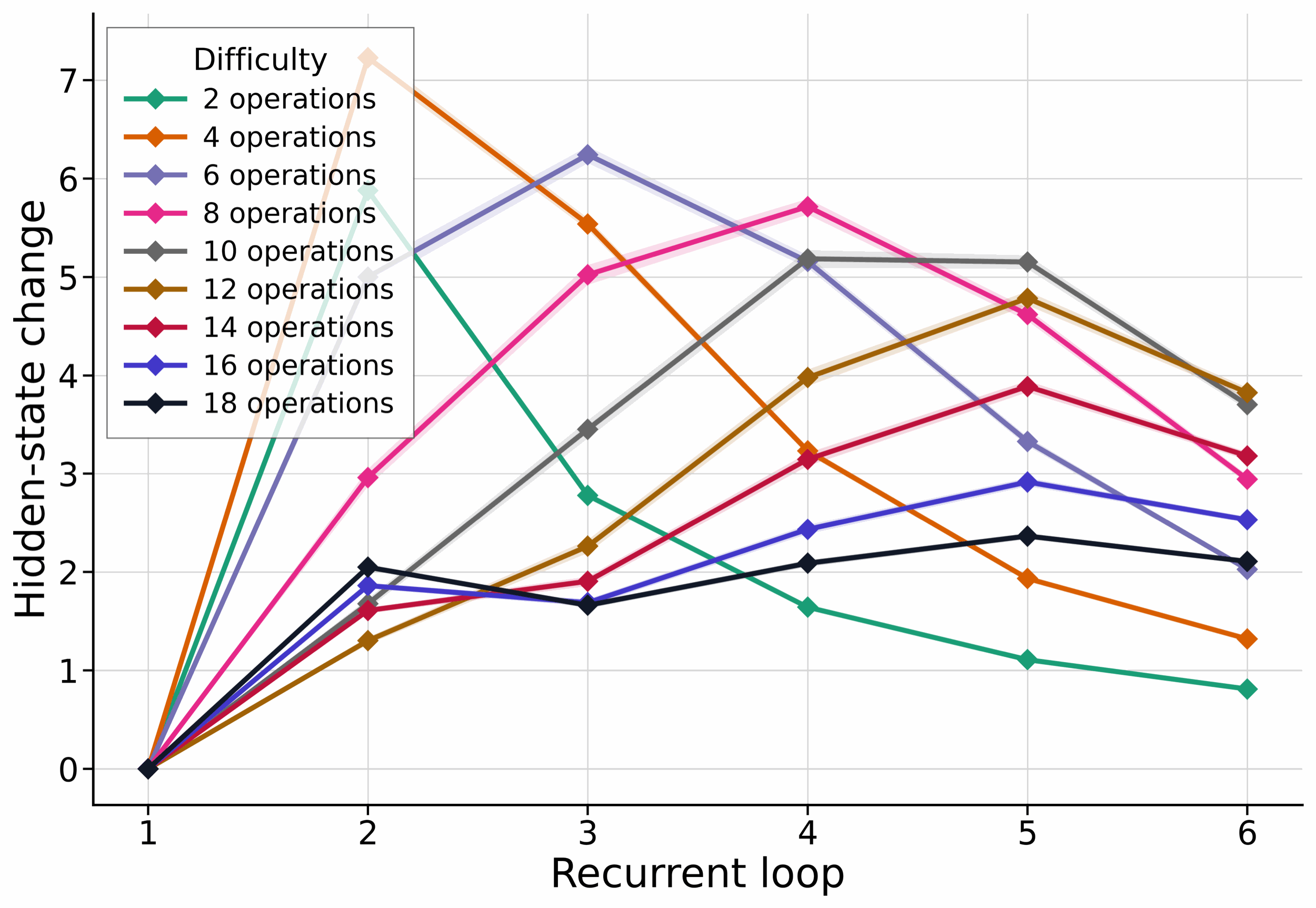}
    \caption{Hidden-state displacement.}
\end{subfigure}

\caption{
Difficulty extrapolation diagnostics for the uniform fixed-prior MANO model.
The model is trained on expressions with at most 10 operations and evaluated on harder expressions beyond that range.
The separation by operation count is less pronounced than under the geometric prior, but harder examples still tend to remain less confident and less converged for longer.
}
\label{fig:mano_difficulty_extrapolation_uniform}
\end{figure}

MANO provides a direct measure of input difficulty through the number of operations in each expression. We use this structure to test whether the difficulty-aware trajectory signals observed in Section~\ref{sec:results_fixed_prior} persist beyond the training difficulty range. We take our fixed-prior models and evaluate them on expressions with more operations. Thus, the harder examples are out of distribution with respect to training difficulty.

This experiment is not a length-generalization benchmark. Accuracy may degrade once expressions exceed the training range. Instead, we use it as a trajectory diagnostic. If the recurrent states encode a meaningful notion of difficulty, then harder out-of-distribution expressions should remain uncertain for more recurrent loops, develop confidence more slowly, and show slower convergence across hidden states.

Figures~\ref{fig:mano_difficulty_extrapolation_geom} and~\ref{fig:mano_difficulty_extrapolation_uniform} show this behavior for geometric and uniform fixed-prior trajectories. The geometric-prior trajectory gives a clear separation by operation count. The uniform trajectory is slightly noisier, but shows the same pattern. Across both priors, expressions with more operations remain higher-entropy for longer, develop logit margin more slowly, and tend to exhibit larger hidden-state changes across recurrent depth. This supports the interpretation that the difficulty-aware structure in the trajectory is not only a calibration artifact of the training distribution.

\section{Learned Gate Comparison}
\label{app:gate_vs_fixed}

This appendix reports the full gate and fixed-prior comparison across every prior setting, expanding the representative panels in Section~\ref{sec:results_fixed_vs_gate}. For each prior, we sweep the learned gate's regularization strength \(\beta\) and compare its native cumulative-threshold readout against post-hoc readouts applied to the fixed-prior trajectory. Appendix~\ref{app:linear-gate} covers the linear gate and Appendix~\ref{app:mlp-gate} covers the MLP gate. Across both gate classes and all priors, the same pattern holds: fixed-prior trajectories paired with simple confidence readouts provide a strong low-depth frontier, and increasing gate capacity narrows but does not close the gap.

\subsection{Linear Gate}
\label{app:linear-gate}

\begin{figure}[htbp]
    \centering

    \begin{subfigure}{0.31\linewidth}
        \centering
        \includegraphics[width=\linewidth]{images/linear_gate/pareto/uniform_linear_gate_vs_fixed_heuristics.png}
        \caption{Uniform prior}
        \label{fig:app_linear_gate_vs_fixed_uniform}
    \end{subfigure}
    \hfill
    \begin{subfigure}{0.31\linewidth}
        \centering
        \includegraphics[width=\linewidth]{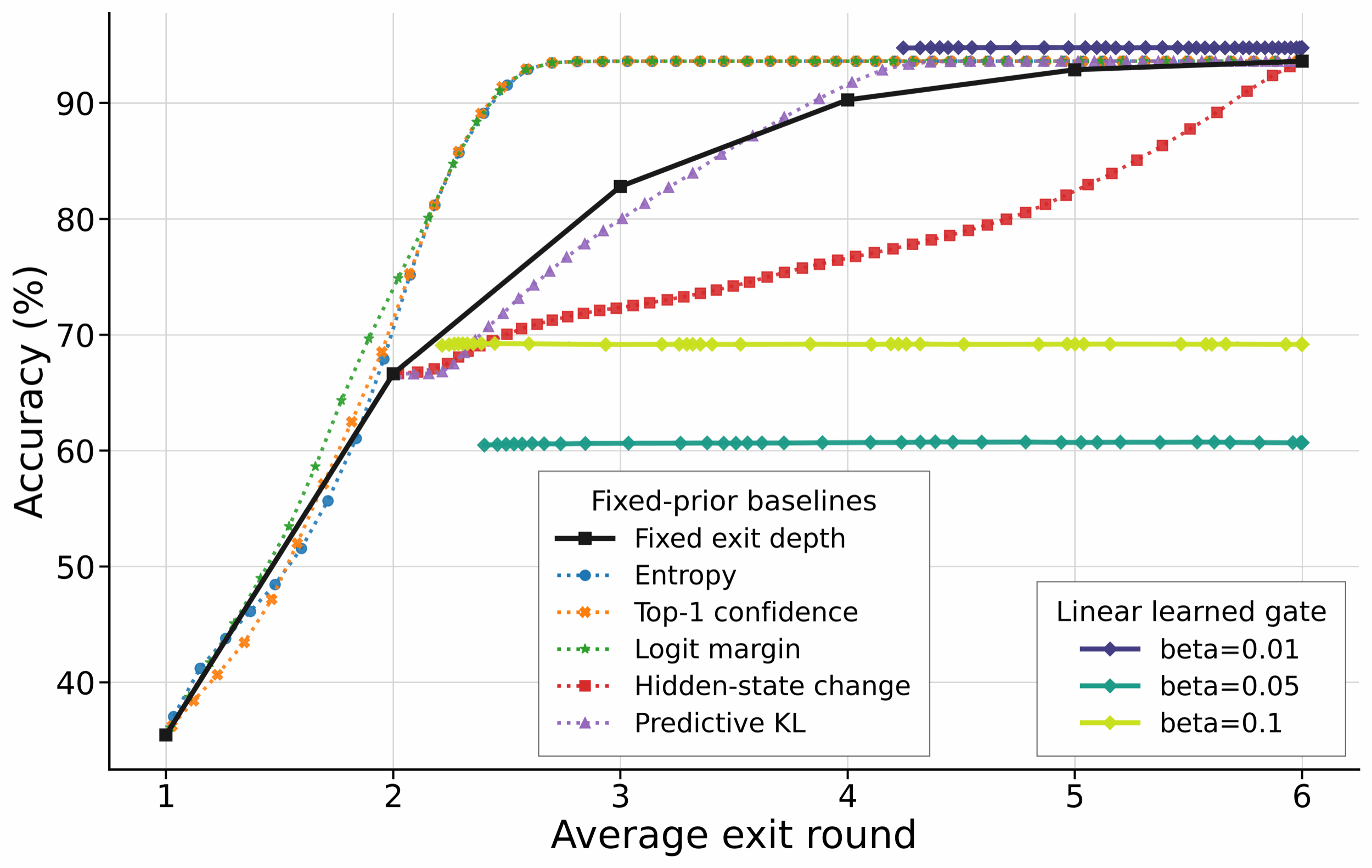}
        \caption{Geometric prior, $\lambda=0.2$}
        \label{fig:app_linear_gate_vs_fixed_geom02}
    \end{subfigure}
    \hfill
    \begin{subfigure}{0.31\linewidth}
        \centering
        \includegraphics[width=\linewidth]{images/linear_gate/pareto/geom_lambd03_linear_gate_vs_fixed_heuristics.png}
        \caption{Geometric prior, $\lambda=0.3$}
        \label{fig:app_linear_gate_vs_fixed_geom03}
    \end{subfigure}

    \vspace{0.6em}

    \begin{subfigure}{0.42\linewidth}
        \centering
        \includegraphics[width=\linewidth]{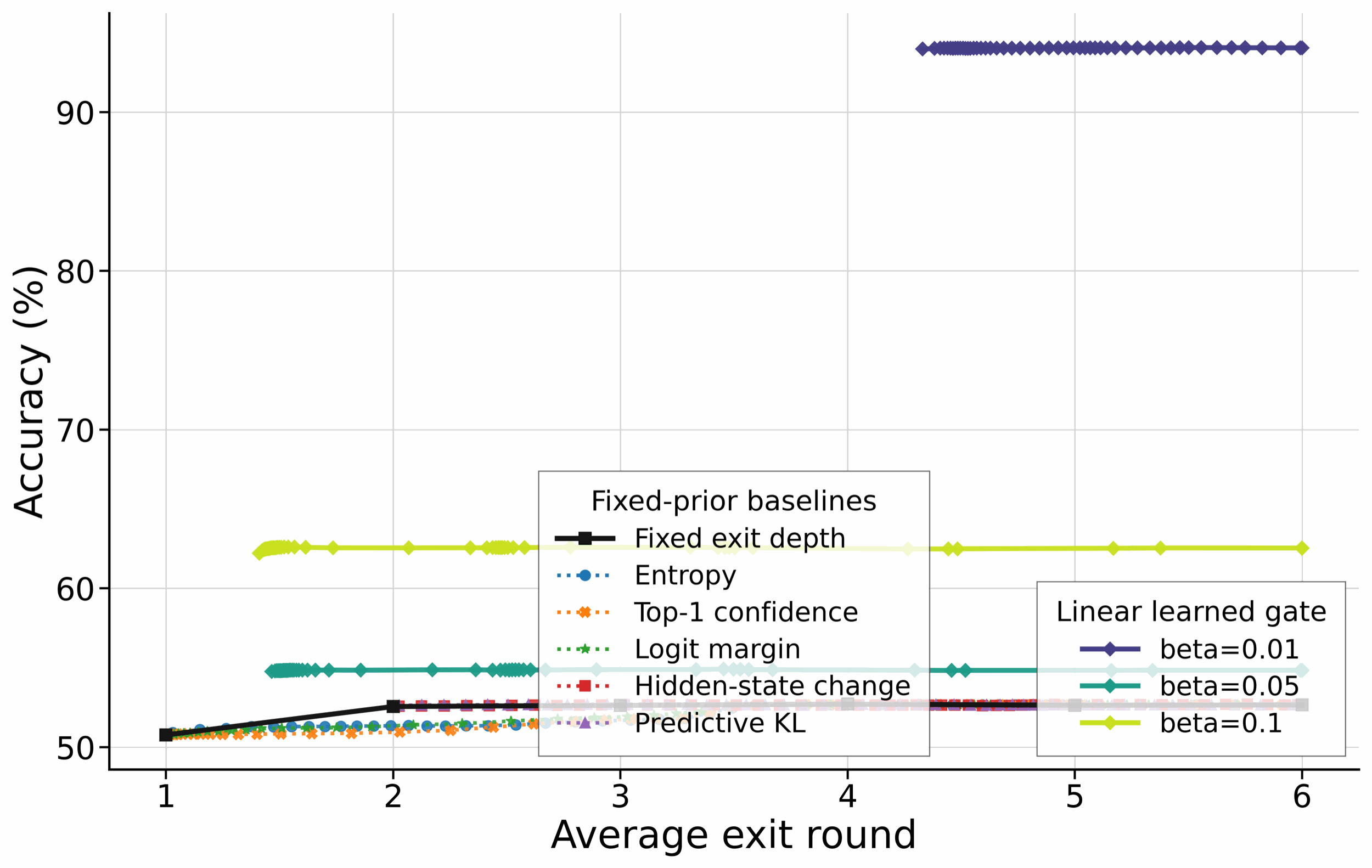}
        \caption{Geometric prior, $\lambda=0.5$}
        \label{fig:app_linear_gate_vs_fixed_geom05}
    \end{subfigure}
    \hfill
    \begin{subfigure}{0.42\linewidth}
        \centering
        \includegraphics[width=\linewidth]{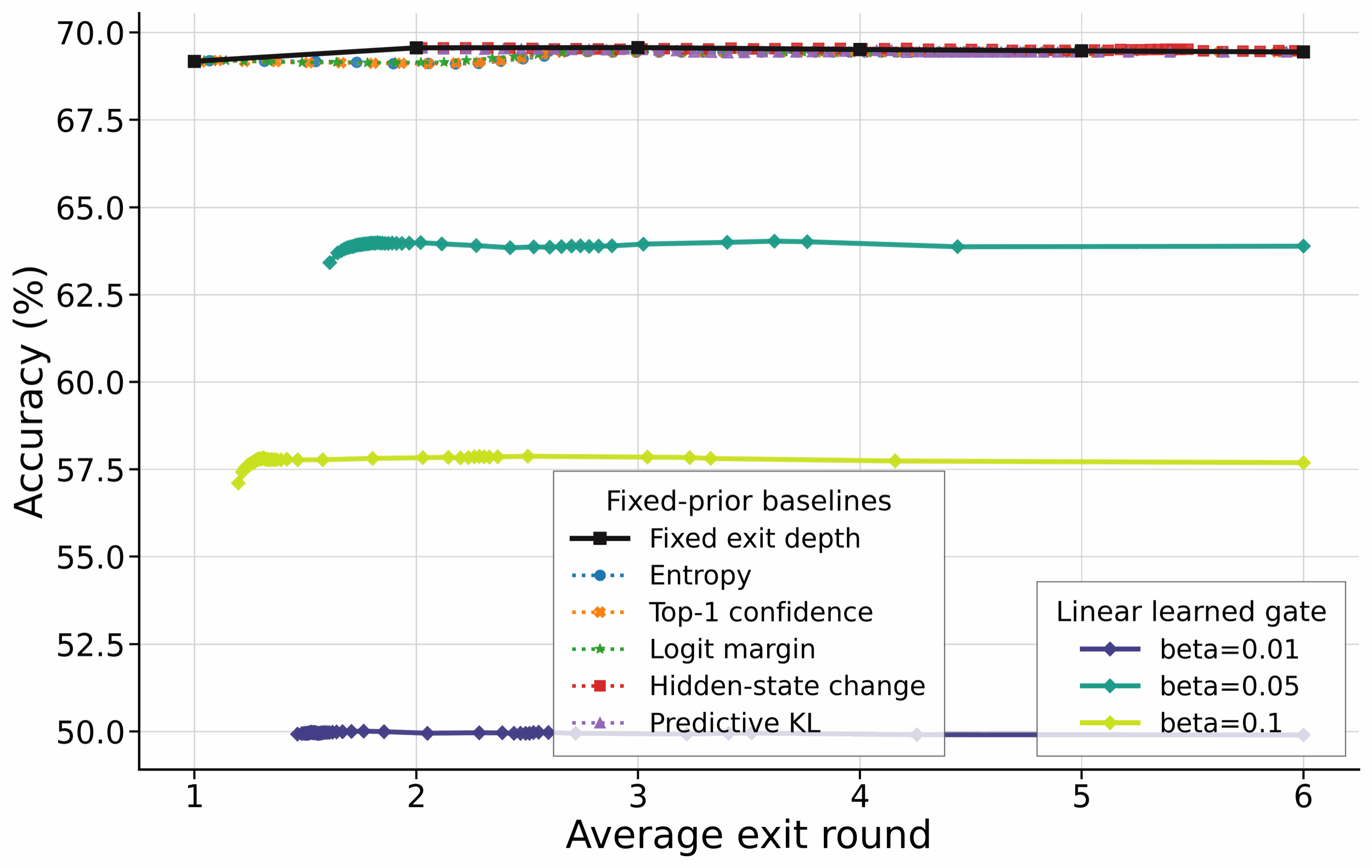}
        \caption{Geometric prior, $\lambda=0.7$}
        \label{fig:app_linear_gate_vs_fixed_geom07}
    \end{subfigure}

    \caption{
    Fixed-prior readouts compared with learned linear-gate models across all fixed-prior settings.
    Each panel plots test accuracy against average exit depth.
    Fixed-prior curves are obtained by sweeping post-hoc readout thresholds, while learned-gate curves are obtained by sweeping the cumulative gate threshold.
    }
    \label{fig:app_linear_gate_vs_fixed_all_priors}
\end{figure}

Figure~\ref{fig:app_linear_gate_vs_fixed_all_priors} compares fixed-prior readouts against learned linear-gate models across all fixed-prior settings. In every panel, the fixed-prior trajectory with simple confidence readouts such as top-1 confidence and logit margin traces a good compute--quality frontier, reaching high accuracy at low average exit depth. The learned linear gate behaves differently across the regularization sweep. Small $\beta$ allows the gate to spread exit mass across depths but does not produce a frontier that dominates the post-hoc readouts, while large $\beta$ pulls the exit distribution toward the prior and collapses the gate onto a near-fixed-depth operating point, visible as the gate curve contracting to a small range of average depths. Under the more aggressive geometric priors ($\lambda = 0.5, 0.7$), both the gate and the trajectory saturate at a lower accuracy ceiling, consistent with the trajectory degradation noted in Figure~\ref{fig:appendix_fixed_prior_pareto_sweep}. The overall message matches the main text. Learning an input-dependent halting distribution with a linear gate is not sufficient to recover the frontier that a well-shaped fixed-prior trajectory already exposes.

\subsection{MLP Gate}
\label{app:mlp-gate}

\begin{figure}[htbp]
    \centering

    \begin{subfigure}{0.31\linewidth}
        \centering
        \includegraphics[width=\linewidth]{images/mlp_gate/pareto/uniform_mlp_gate_vs_fixed_heuristics.png}
        \caption{Uniform prior}
        \label{fig:app_mlp_gate_vs_fixed_uniform}
    \end{subfigure}
    \hfill
    \begin{subfigure}{0.31\linewidth}
        \centering
        \includegraphics[width=\linewidth]{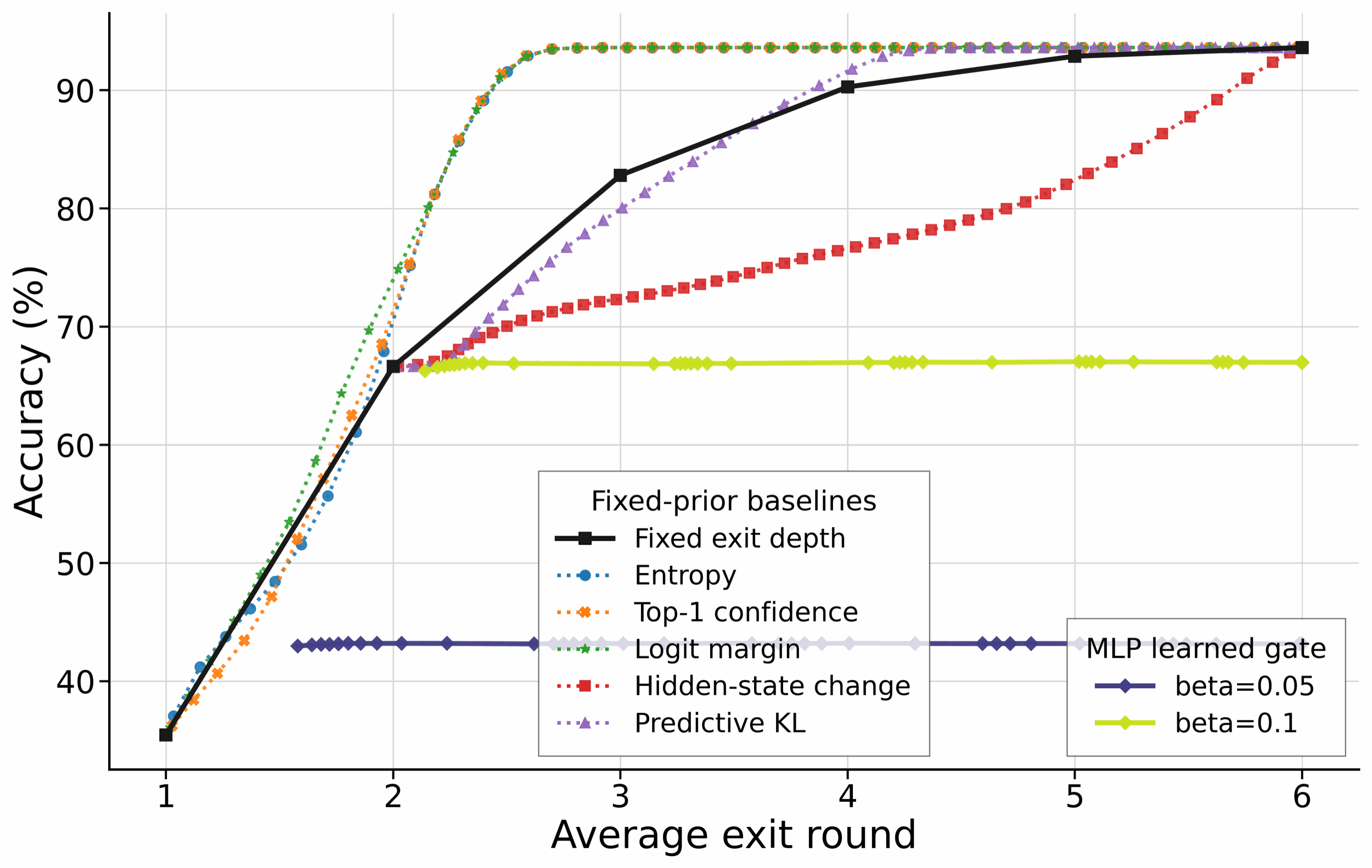}
        \caption{Geometric prior, $\lambda=0.2$}
        \label{fig:app_mlp_gate_vs_fixed_geom02}
    \end{subfigure}
    \hfill
    \begin{subfigure}{0.31\linewidth}
        \centering
        \includegraphics[width=\linewidth]{images/mlp_gate/pareto/geom_lambd03_mlp_gate_vs_fixed_heuristics.png}
        \caption{Geometric prior, $\lambda=0.3$}
        \label{fig:app_mlp_gate_vs_fixed_geom03}
    \end{subfigure}

    \vspace{0.6em}

    \begin{subfigure}{0.44\linewidth}
        \centering
        \includegraphics[width=\linewidth]{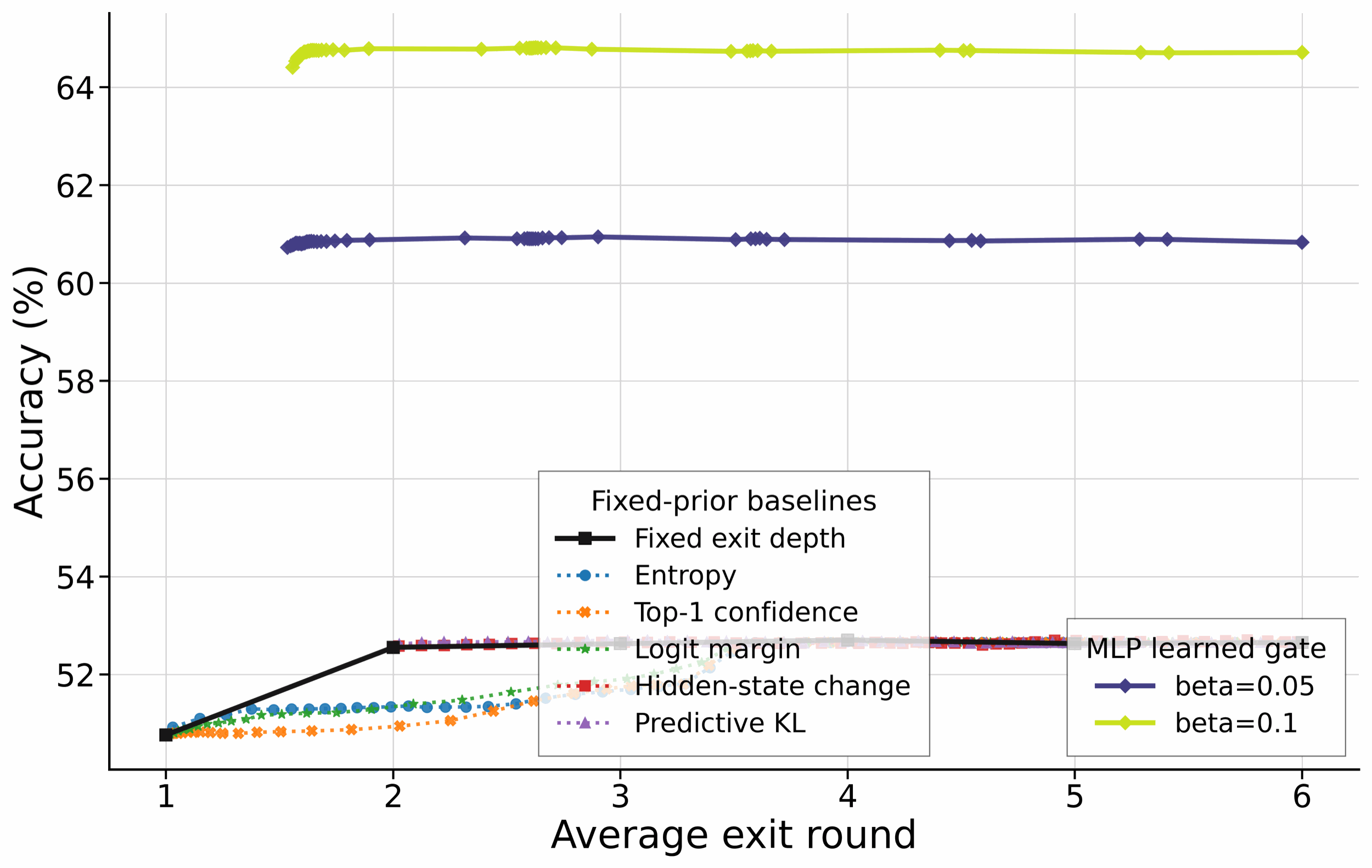}
        \caption{Geometric prior, $\lambda=0.5$}
        \label{fig:app_mlp_gate_vs_fixed_geom05}
    \end{subfigure}
    \hfill
    \begin{subfigure}{0.44\linewidth}
        \centering
        \includegraphics[width=\linewidth]{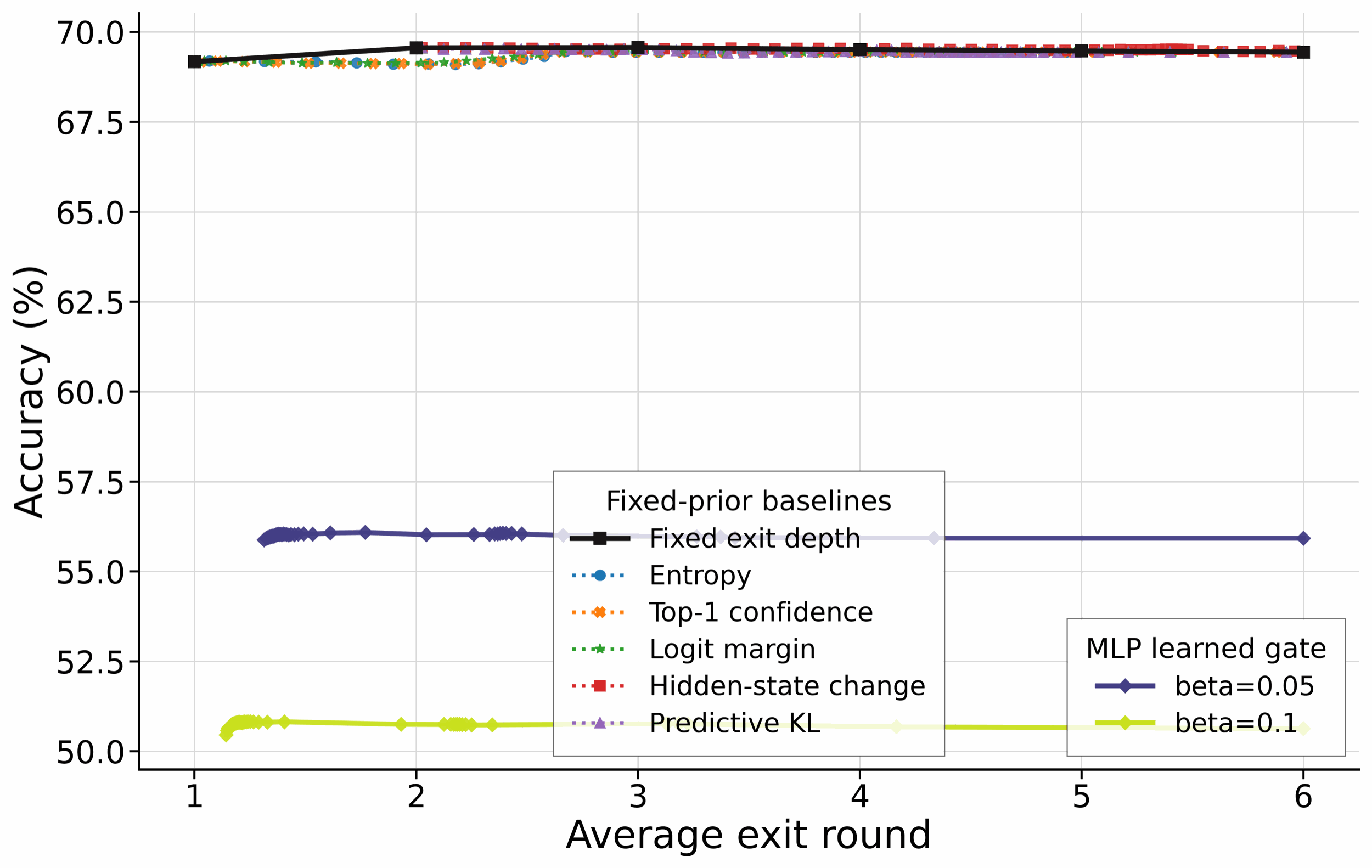}
        \caption{Geometric prior, $\lambda=0.7$}
        \label{fig:app_mlp_gate_vs_fixed_geom07}
    \end{subfigure}

    \caption{
    Fixed-prior readouts compared with learned MLP-gate models across all fixed-prior settings.
    Each panel plots test accuracy against average exit depth.
    Fixed-prior curves are obtained by sweeping post-hoc readout thresholds, while learned-gate curves are obtained by sweeping the cumulative gate threshold.
    }
    \label{fig:app_mlp_gate_vs_fixed_all_priors}
\end{figure}

Figure~\ref{fig:app_mlp_gate_vs_fixed_all_priors} repeats the comparison with the learned halt head replaced by a two-layer MLP gate. Increasing gate expressivity does improve the learned-gate baseline relative to the linear gate, producing somewhat smoother frontiers and reaching higher accuracy at intermediate depths in several panels. However, the additional capacity does not change the qualitative outcome. The fixed-prior trajectory with simple post-hoc readouts still provides a solid low-depth frontier across priors, and the MLP gate remains limited by the trajectory it induces during joint training rather than by the form of the halt head. This is consistent with the frozen-trajectory experiments in Appendix~\ref{app:posthoc-gates}, which show that fitting a more expressive gate on a jointly trained trajectory does not recover the frontier available from a fixed-prior trajectory. Taken together, the linear and MLP comparisons indicate that gate expressivity is not the binding constraint, but the trajectory formation is.

\section{Trajectory Diagnostics}
\label{app:forced_exit_all}

\begin{figure}[htbp]
    \centering
    \begin{subfigure}{0.48\linewidth}
        \centering
        \includegraphics[width=\linewidth]{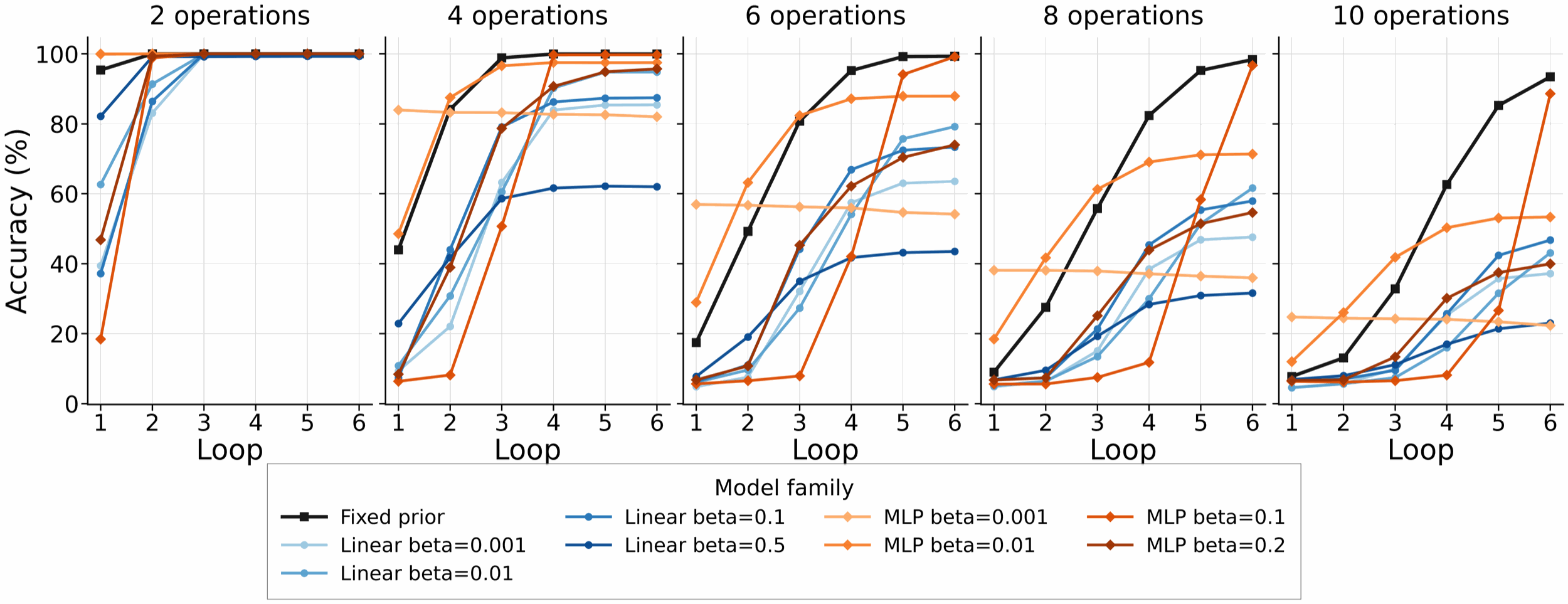}
        \caption{Uniform prior}
        \label{fig:app_uniform_forced_exit_performance}
    \end{subfigure}
    \hfill
    \begin{subfigure}{0.48\linewidth}
        \centering
        \includegraphics[width=\linewidth]{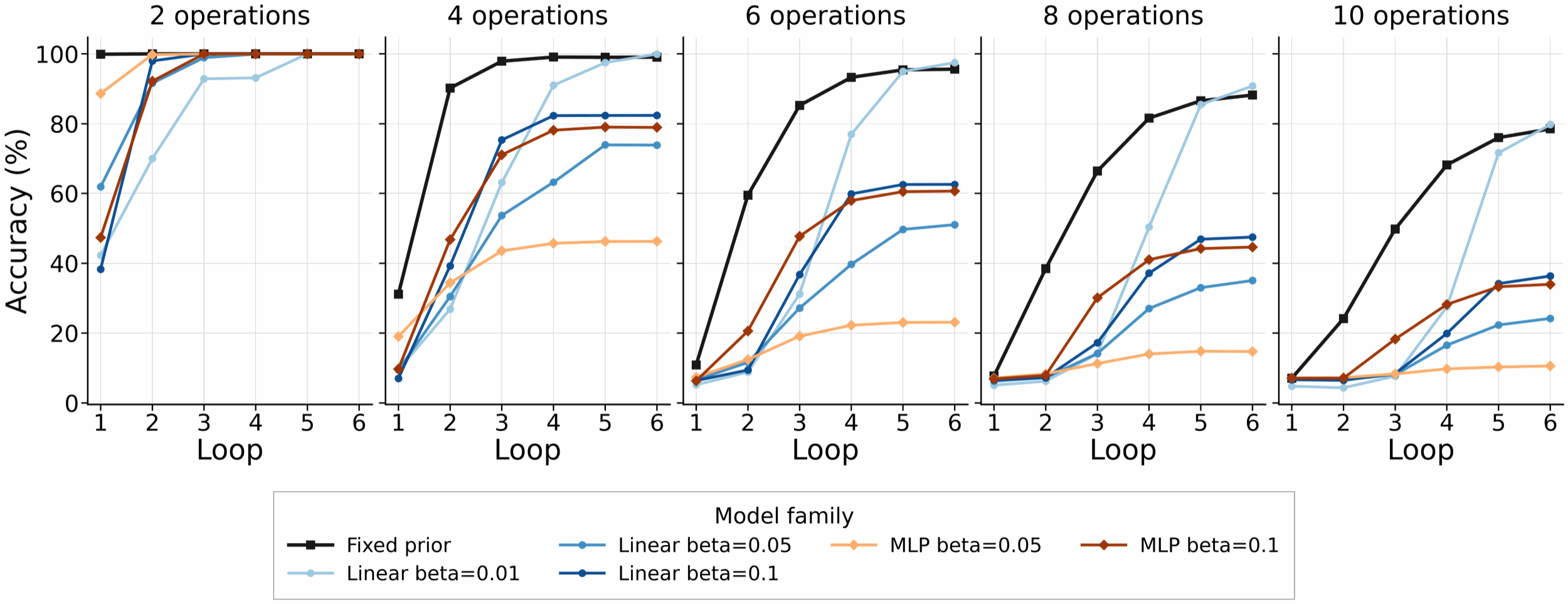}
        \caption{Geometric prior, $\lambda=0.2$}
        \label{fig:app_geom02_forced_exit_performance}
    \end{subfigure}

    \vspace{0.5em}

    \begin{subfigure}{0.48\linewidth}
        \centering
        \includegraphics[width=\linewidth]{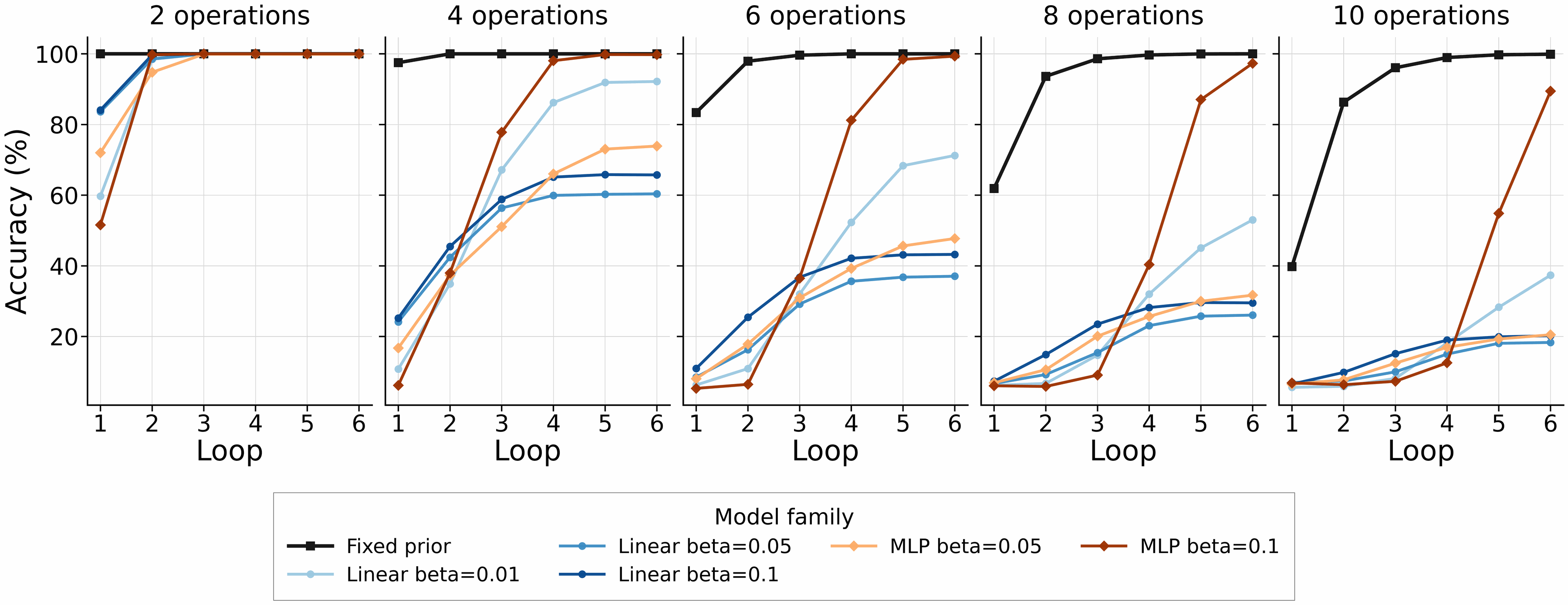}
        \caption{Geometric prior, $\lambda=0.3$}
        \label{fig:app_geom03_forced_exit_performance}
    \end{subfigure}
    \hfill
    \begin{subfigure}{0.48\linewidth}
        \centering
        \includegraphics[width=\linewidth]{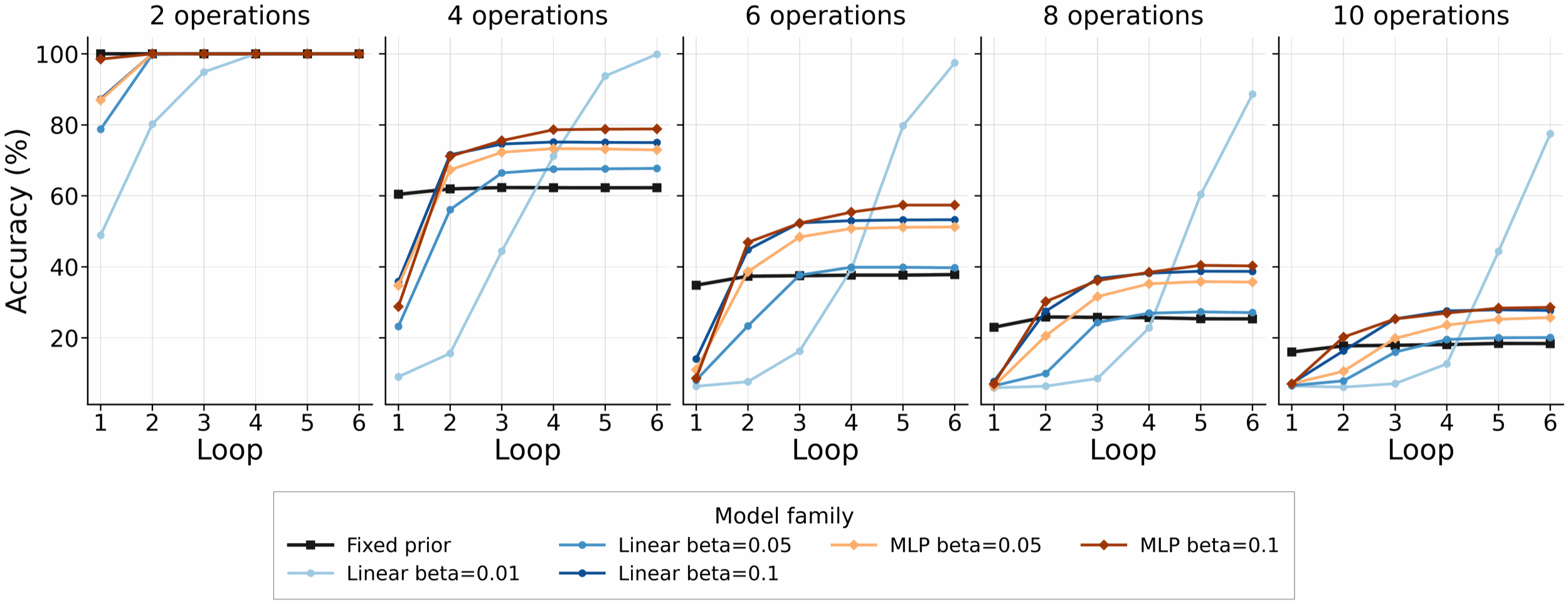}
        \caption{Geometric prior, $\lambda=0.5$}
        \label{fig:app_geom05_forced_exit_performance}
    \end{subfigure}

    \vspace{0.5em}

    \begin{subfigure}{0.48\linewidth}
        \centering
        \includegraphics[width=\linewidth]{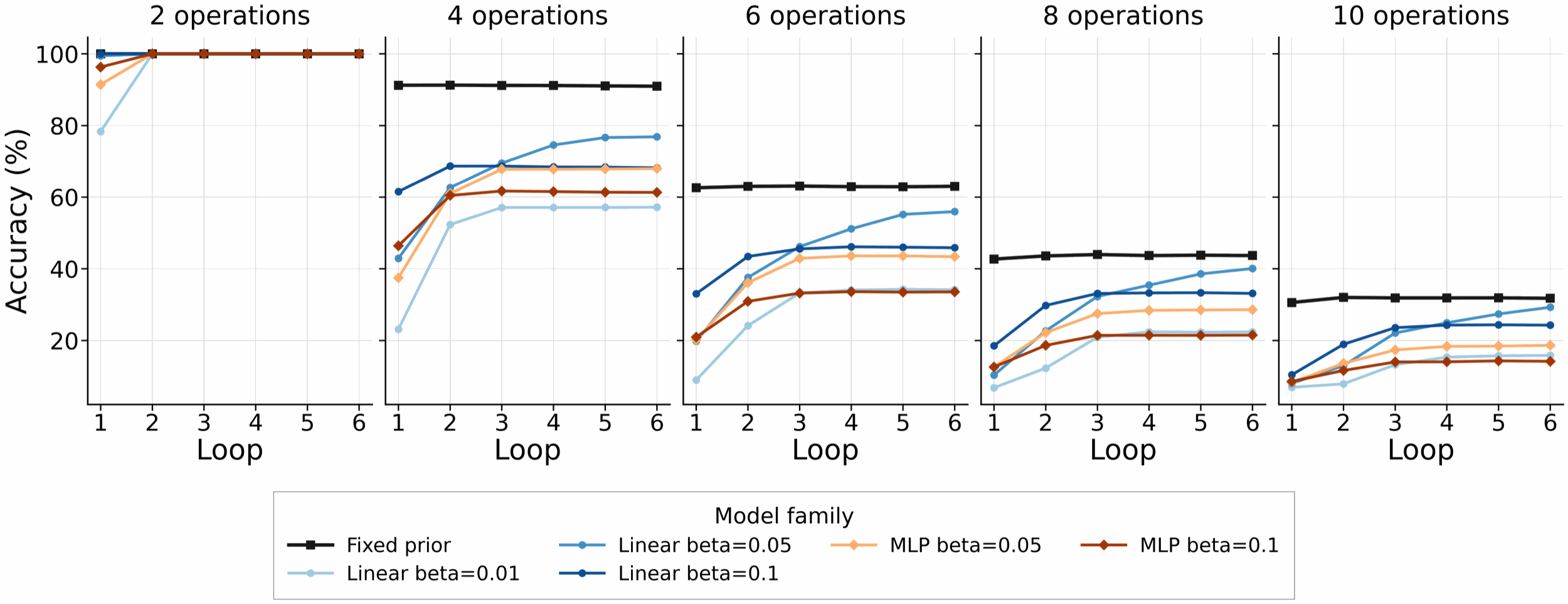}
        \caption{Geometric prior, $\lambda=0.7$}
        \label{fig:app_geom07_forced_exit_performance}
    \end{subfigure}

    \caption{
    Forced-exit accuracy by task difficulty across fixed-prior settings.
    Each panel evaluates the prediction at every recurrent loop without using an adaptive readout.
    }
    \label{fig:app_forced_exit_performance_by_difficulty}
\end{figure}

\begin{figure}[htbp]
    \centering
    \begin{subfigure}{0.48\linewidth}
        \centering
        \includegraphics[width=\linewidth]{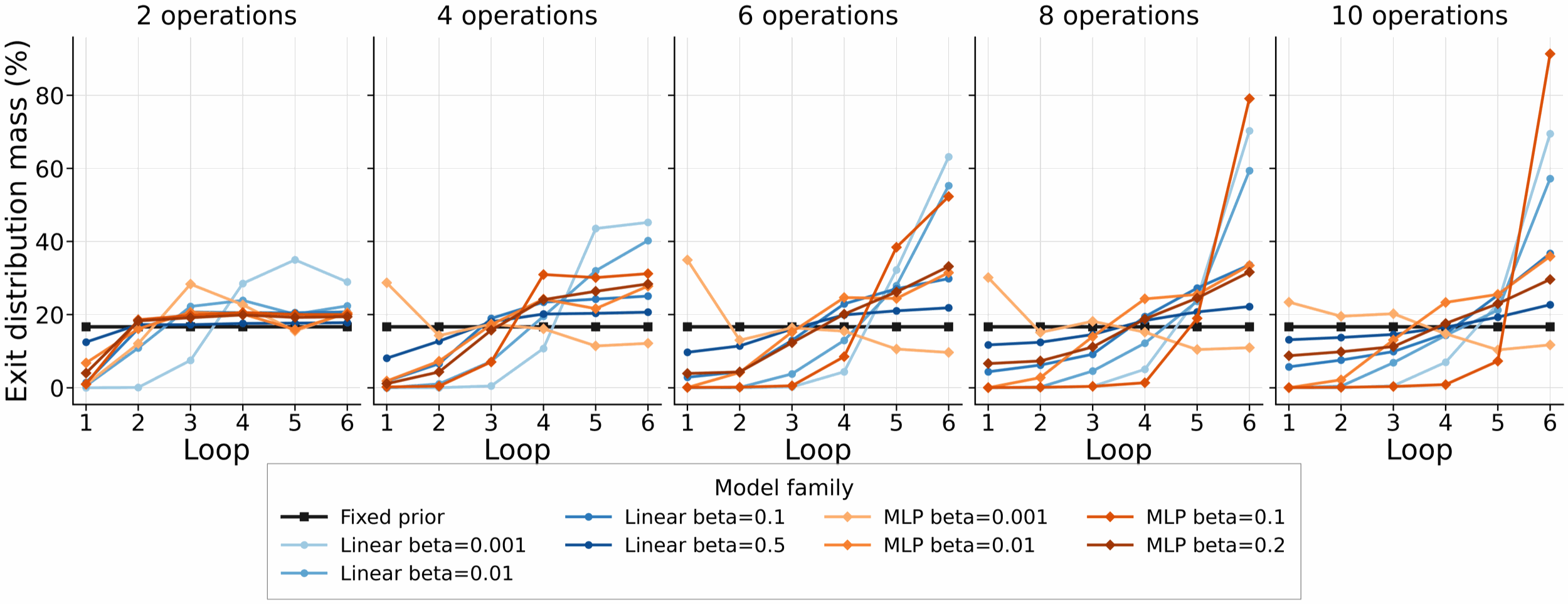}
        \caption{Uniform prior}
        \label{fig:app_uniform_exit_distribution}
    \end{subfigure}
    \hfill
    \begin{subfigure}{0.48\linewidth}
        \centering
        \includegraphics[width=\linewidth]{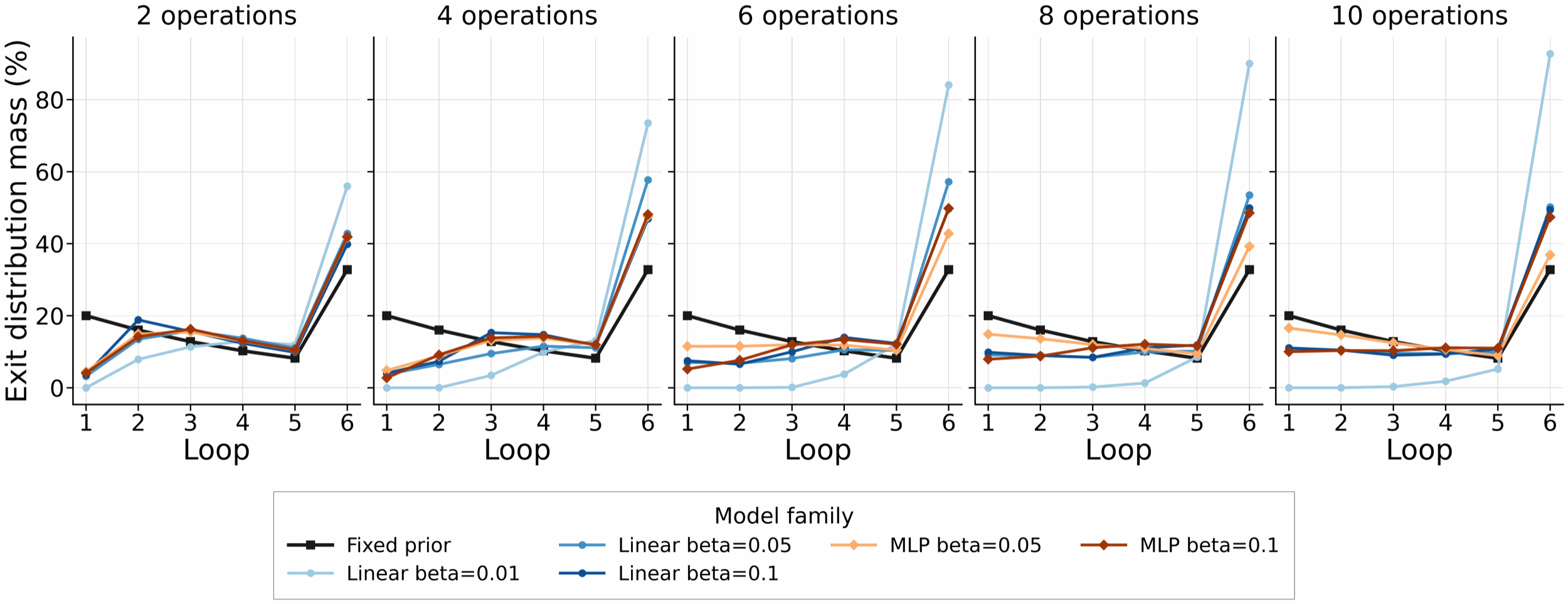}
        \caption{Geometric prior, $\lambda=0.2$}
        \label{fig:app_geom02_exit_distribution}
    \end{subfigure}

    \vspace{0.5em}

    \begin{subfigure}{0.48\linewidth}
        \centering
        \includegraphics[width=\linewidth]{images/forced_exit_trajectory_comparison/geom_lambd03_exit_distribution_by_difficulty.png}
        \caption{Geometric prior, $\lambda=0.3$}
        \label{fig:app_geom03_exit_distribution}
    \end{subfigure}
    \hfill
    \begin{subfigure}{0.48\linewidth}
        \centering
        \includegraphics[width=\linewidth]{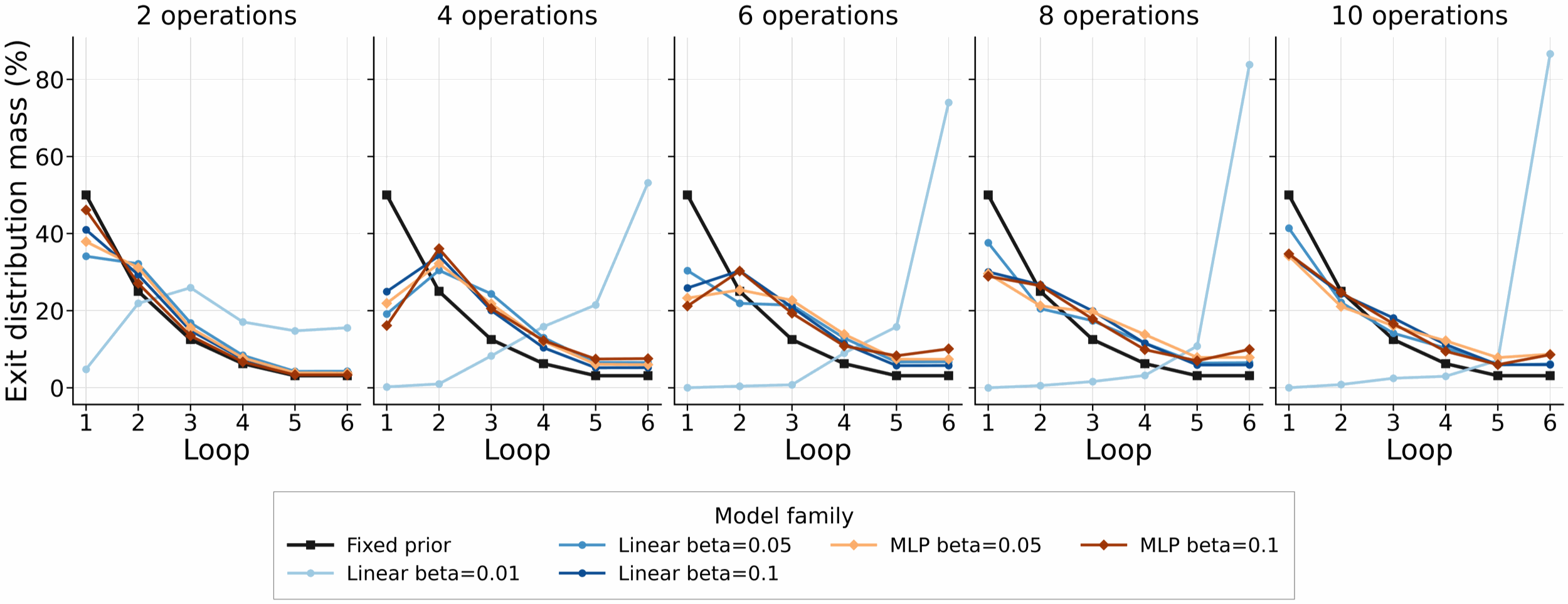}
        \caption{Geometric prior, $\lambda=0.5$}
        \label{fig:app_geom05_exit_distribution}
    \end{subfigure}

    \vspace{0.5em}

    \begin{subfigure}{0.48\linewidth}
        \centering
        \includegraphics[width=\linewidth]{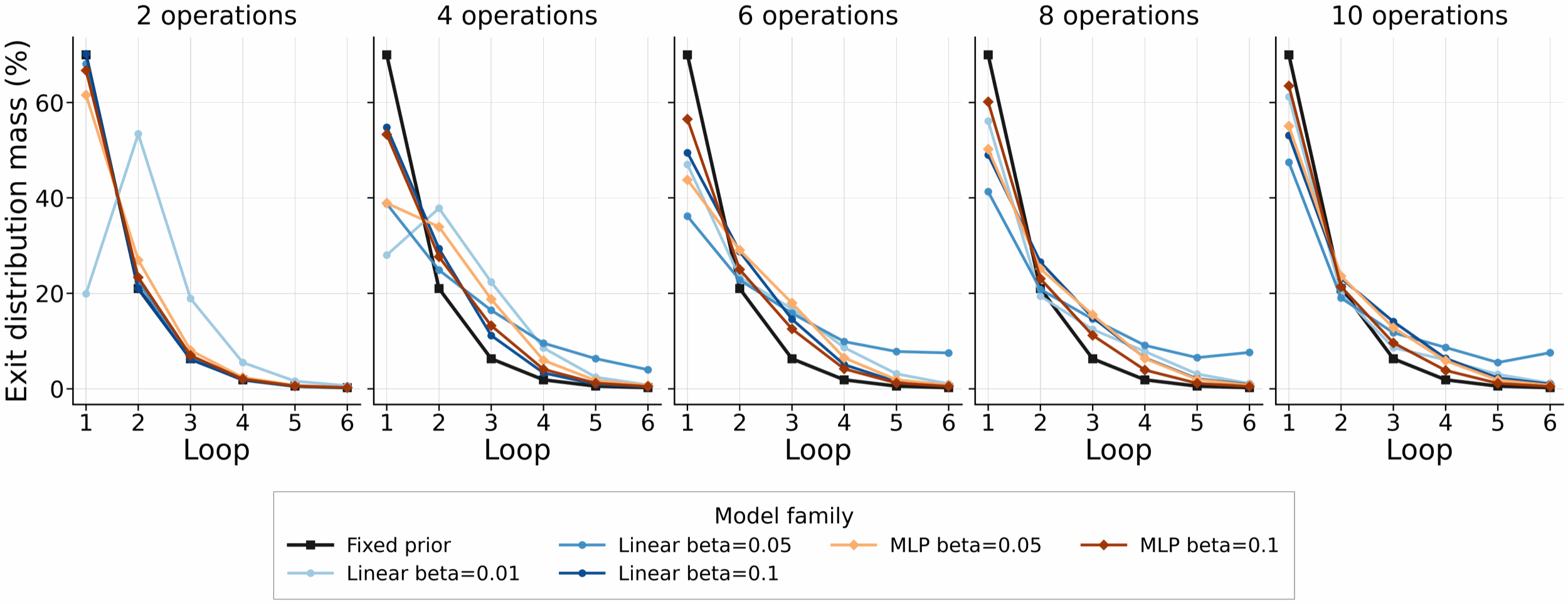}
        \caption{Geometric prior, $\lambda=0.7$}
        \label{fig:app_geom07_exit_distribution}
    \end{subfigure}

    \caption{
    Learned exit distributions by task difficulty across prior settings.
    Each panel shows the average learned exit probability assigned to each recurrent loop for examples grouped by task difficulty.
    }
    \label{fig:app_exit_distribution_by_difficulty}
\end{figure}

This appendix expands the forced-exit and exit-distribution analysis of Section~\ref{sec:training-objective-shapes-trajectories} to all prior and gate settings. Figure~\ref{fig:app_forced_exit_performance_by_difficulty} reports forced-exit accuracy by difficulty, and Figure~\ref{fig:app_exit_distribution_by_difficulty} reports the corresponding learned exit distributions. Forced exits evaluate the prediction at every recurrent loop without applying any adaptive stopping rule, so they measure the per-depth quality of the trajectory itself. The learned exit distributions show, for the gate models, how exit mass is allocated across depths as a function of task difficulty.

Figure~\ref{fig:app_forced_exit_performance_by_difficulty} reports forced-exit accuracy grouped by operation count across all fixed-prior and learned-gate settings. The fixed-prior trajectory produces useful intermediate predictions early in the loop, including on harder examples, and its per-depth accuracy rises smoothly with both depth and decreasing difficulty. The learned-gate trajectories show a different per-depth profile. Depending on $\beta$ and gate class, accuracy at early loops is often lower, and the gain from additional loops is concentrated later in the trajectory. This confirms that the gate objective changes which recurrent states become accurate, not only the rule used to select among them. The contrast is most pronounced on the harder difficulty bins, where the fixed-prior trajectory already supports strong early exits but the jointly trained gate trajectory does not.

Figure~\ref{fig:app_exit_distribution_by_difficulty} shows the average learned exit distribution over recurrent loops for the gate models, grouped by difficulty. The gates do learn a difficulty-aware allocation: easier examples receive more exit mass at early loops, while harder examples shift mass toward later depths. However, read together with the forced-exit curves, this allocation is not sufficient on its own. Assigning later exits to hard examples only helps if the corresponding late recurrent states are accurate, and the forced-exit panels show that the jointly trained trajectories do not always produce such states. This is the central diagnostic behind the main-text claim that adaptive-compute performance depends jointly on trajectory formation and exit selection, rather than on the exit distribution alone.

\section{Parity Experiments}
\label{app:parity_diagnostic}

To check that the fixed-prior readout behavior is not specific to MANO, we run an additional small parity experiment. Each input is a binary sequence and the target is its parity. Unlike MANO, this task does not involve arithmetic expression evaluation. We use this task only as a lightweight robustness check, and do not repeat the full MANO benchmark ablation.

\begin{figure}[htbp]
\centering

\begin{subfigure}[t]{0.48\linewidth}
    \centering
    \includegraphics[width=\linewidth]{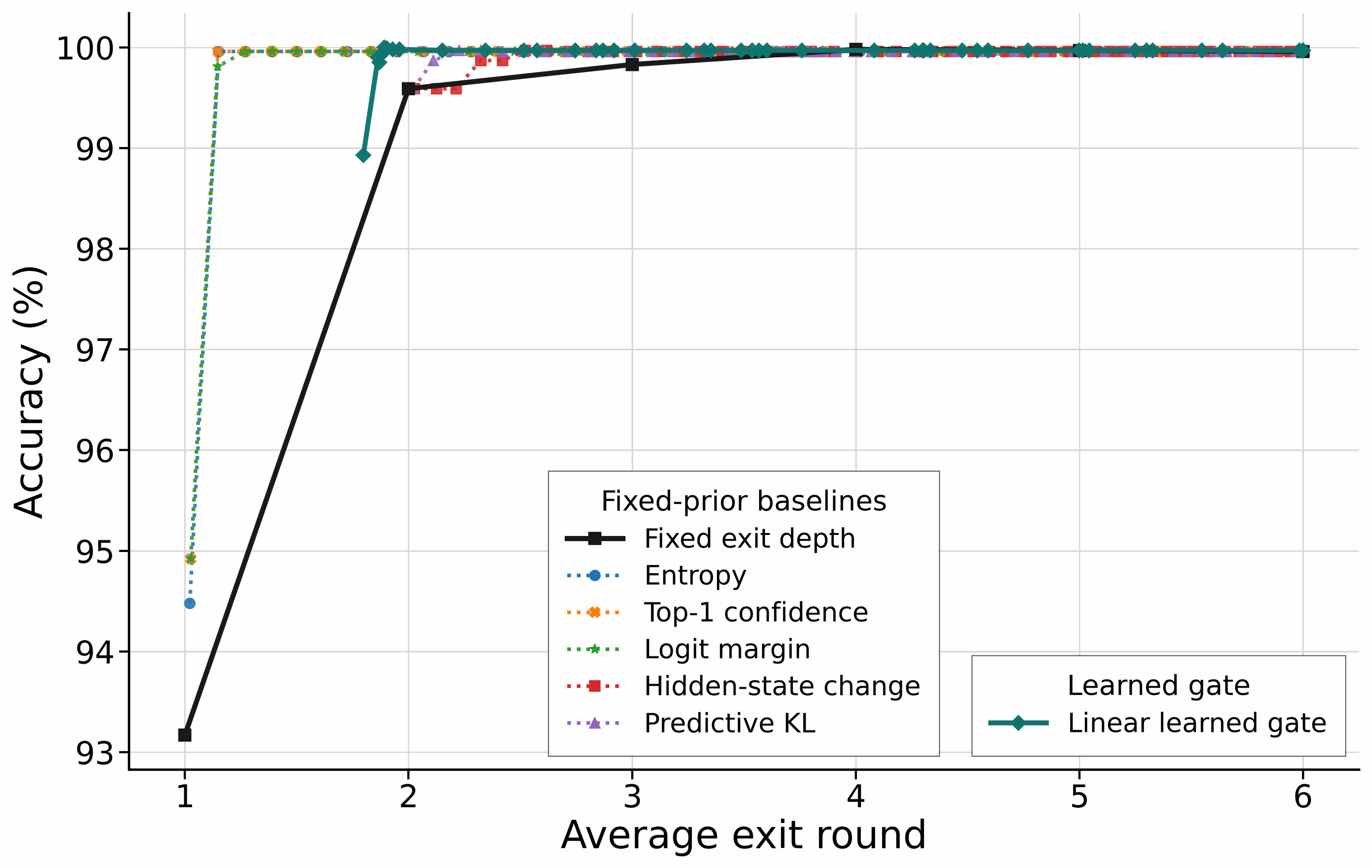}
    \caption{Uniform prior, linear learned gate.}
\end{subfigure}
\hfill
\begin{subfigure}[t]{0.48\linewidth}
    \centering
    \includegraphics[width=\linewidth]{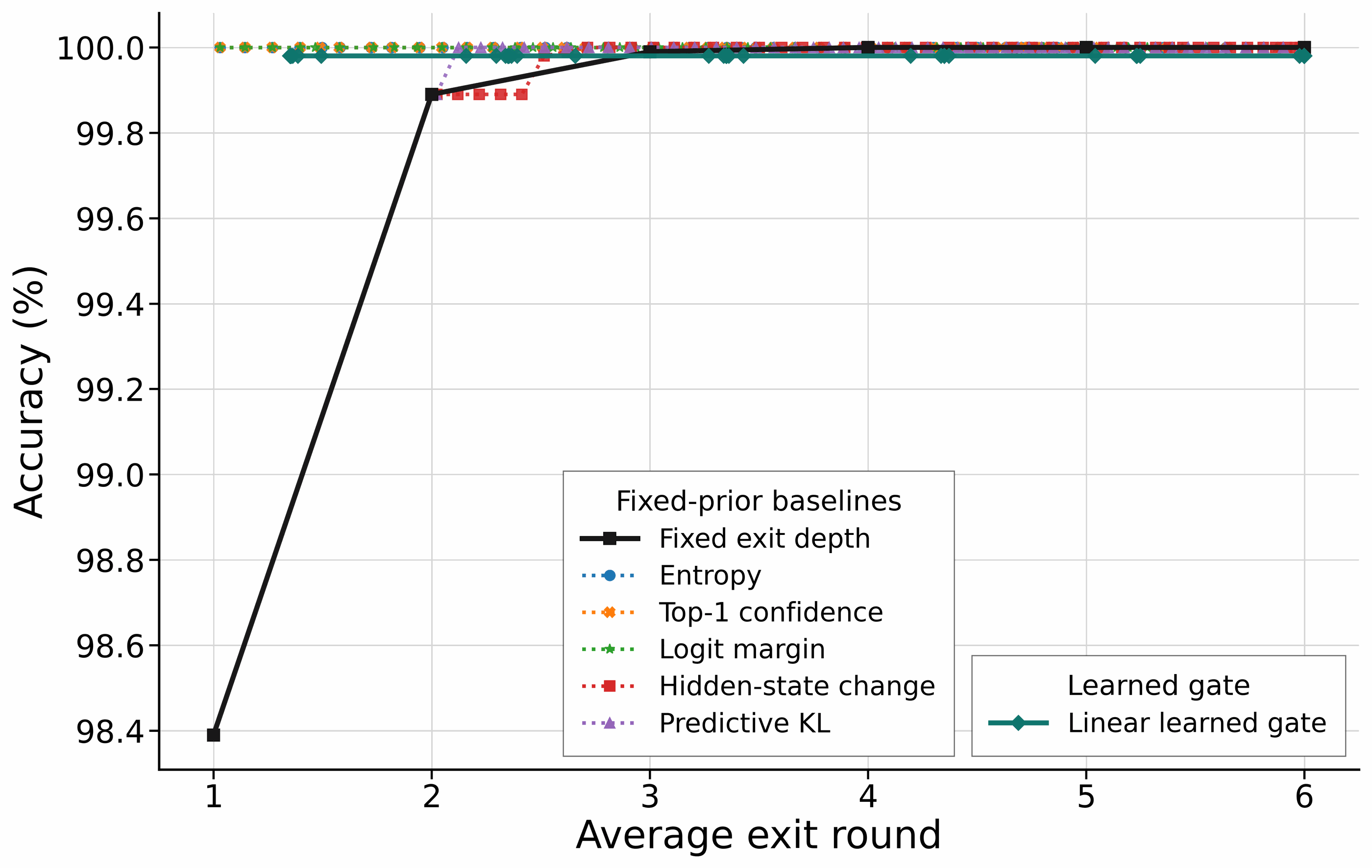}
    \caption{Geometric prior, linear learned gate.}
\end{subfigure}

\vspace{0.8em}

\begin{subfigure}[t]{0.48\linewidth}
    \centering
    \includegraphics[width=\linewidth]{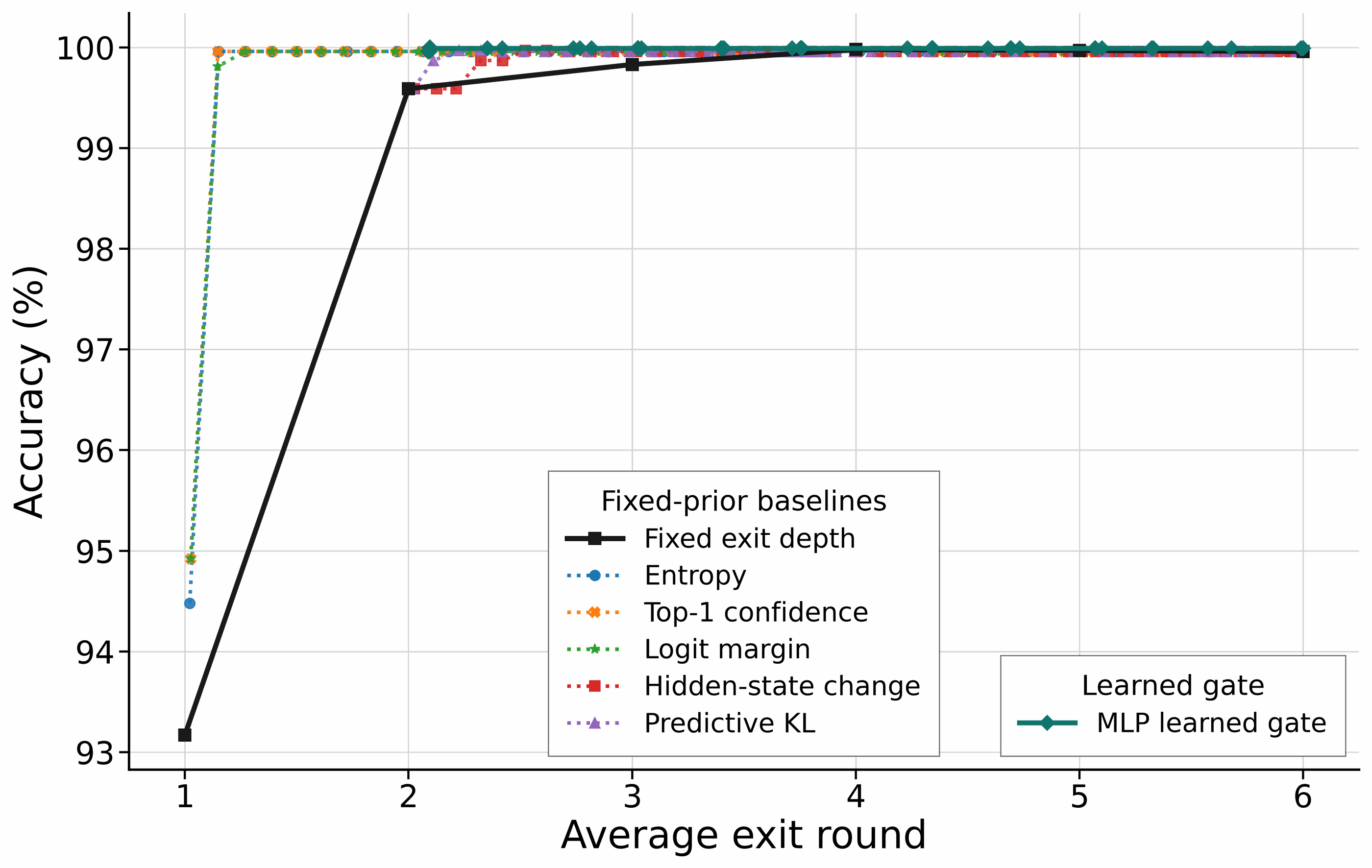}
    \caption{Uniform prior, MLP learned gate.}
\end{subfigure}
\hfill
\begin{subfigure}[t]{0.48\linewidth}
    \centering
    \includegraphics[width=\linewidth]{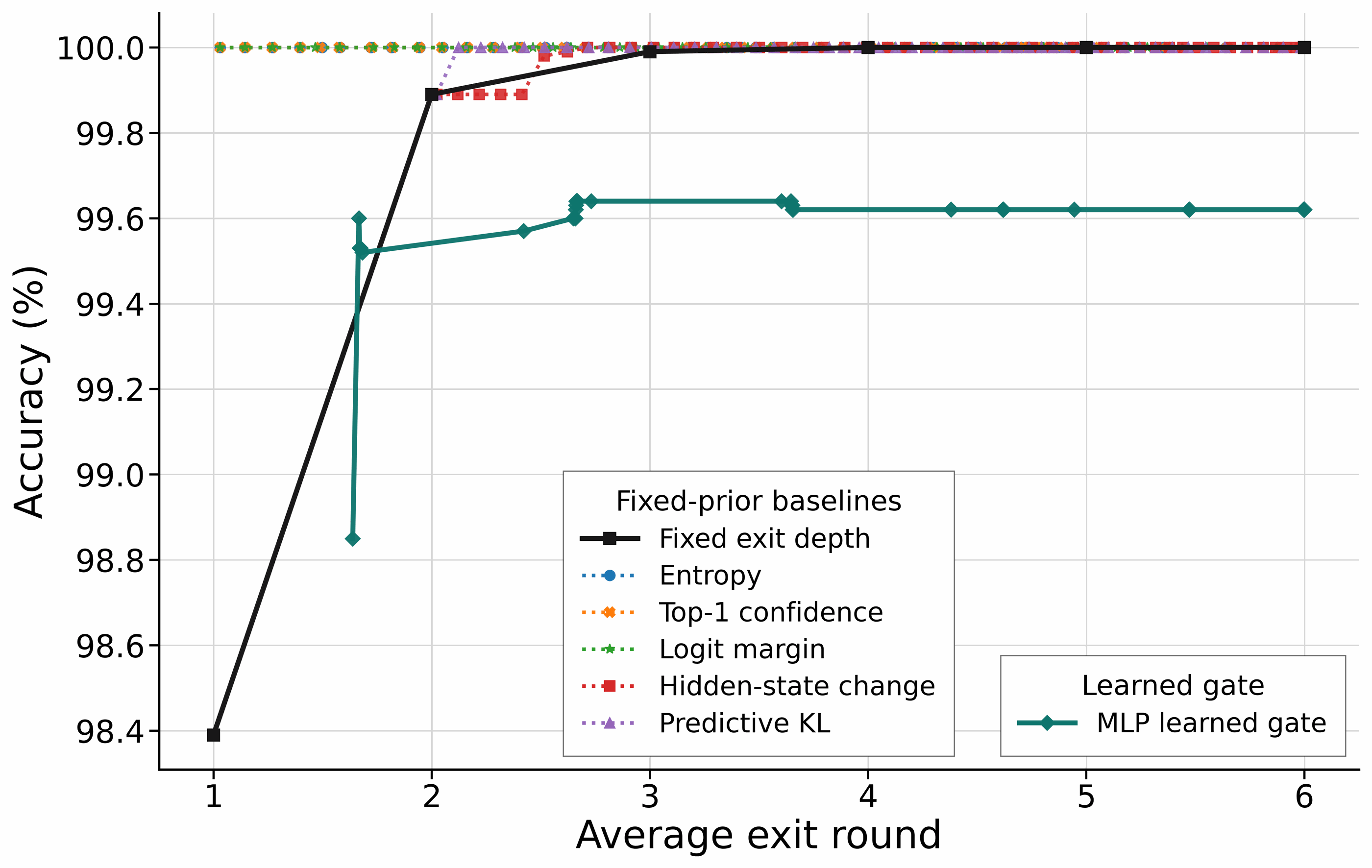}
    \caption{Geometric prior, MLP learned gate.}
\end{subfigure}

\caption{
Parity diagnostic comparing fixed-prior readouts with learned gates.
}
\label{fig:parity_fixed_vs_gate}
\end{figure}

\begin{table}[htbp]
\centering
\caption{
Parity diagnostic over three seeds.
We report mean $\pm$ standard deviation for validation-selected operating points on the held-out test split.
}
\label{tab:parity_seed_summary}
\begin{tabular}{llcc}
\toprule
Training setting & Best Readout & Accuracy (\%) & Avg. loops \\
\midrule
Fixed uniform & Top-1 confidence & $99.6 \pm 0.7$ & $1.30 \pm 0.17$ \\
Fixed geometric, $\lambda=0.3$ & Top-1 confidence & $99.0 \pm 0.8$ & $1.24 \pm 0.28$ \\
\midrule
Linear gate, uniform & Gate & $99.6 \pm 0.3$ & $1.77 \pm 0.08$ \\
MLP gate, uniform & Gate & $99.6 \pm 0.1$ & $2.37 \pm 0.46$ \\
Linear gate, geometric $\lambda=0.3$ & Gate & $98.8 \pm 0.1$ & $1.42 \pm 0.14$ \\
MLP gate, geometric $\lambda=0.3$ & Gate & $98.7 \pm 0.2$ & $2.28 \pm 0.65$ \\
\bottomrule
\end{tabular}
\end{table}

We train looped Transformer models on parity sequences with bit lengths from 1 to 40. For each seed, training examples are generated online and balanced across bit lengths. We evaluate six settings: fixed-prior uniform, fixed-prior geometric with \(\lambda=0.3\), learned linear gates under both priors, and learned MLP gates under both priors. Learned gates use \(\beta=0.1\). All models use \(T=6\) recurrent loops.

Figure~\ref{fig:parity_fixed_vs_gate} shows representative compute--accuracy curves. Because parity is a simple binary classification task, most settings approach saturation, and the accuracy differences are less striking than on MANO. The relevant comparison is therefore the compute required to reach the saturated regime. Under this lens, the fixed-prior models provide useful evidence. Simple confidence readouts reach near-ceiling accuracy at average depths comparable to, and often lower than, the learned gates. Thus, even in a setting where the task is nearly solved by all methods, the recurrent trajectory contains enough stopping information for inexpensive post-hoc readouts to recover competitive adaptive computation.

Table~\ref{tab:parity_seed_summary} summarizes the same parity diagnostic across three seeds, showing that fixed-prior confidence readouts reach near-ceiling accuracy at low average depth.

\begin{figure}[htbp]
\centering

\begin{subfigure}[t]{0.32\linewidth}
    \centering
    \includegraphics[
        width=\linewidth,
        trim={0.10in 0.05in 0.10in 0.08in},
        clip
    ]{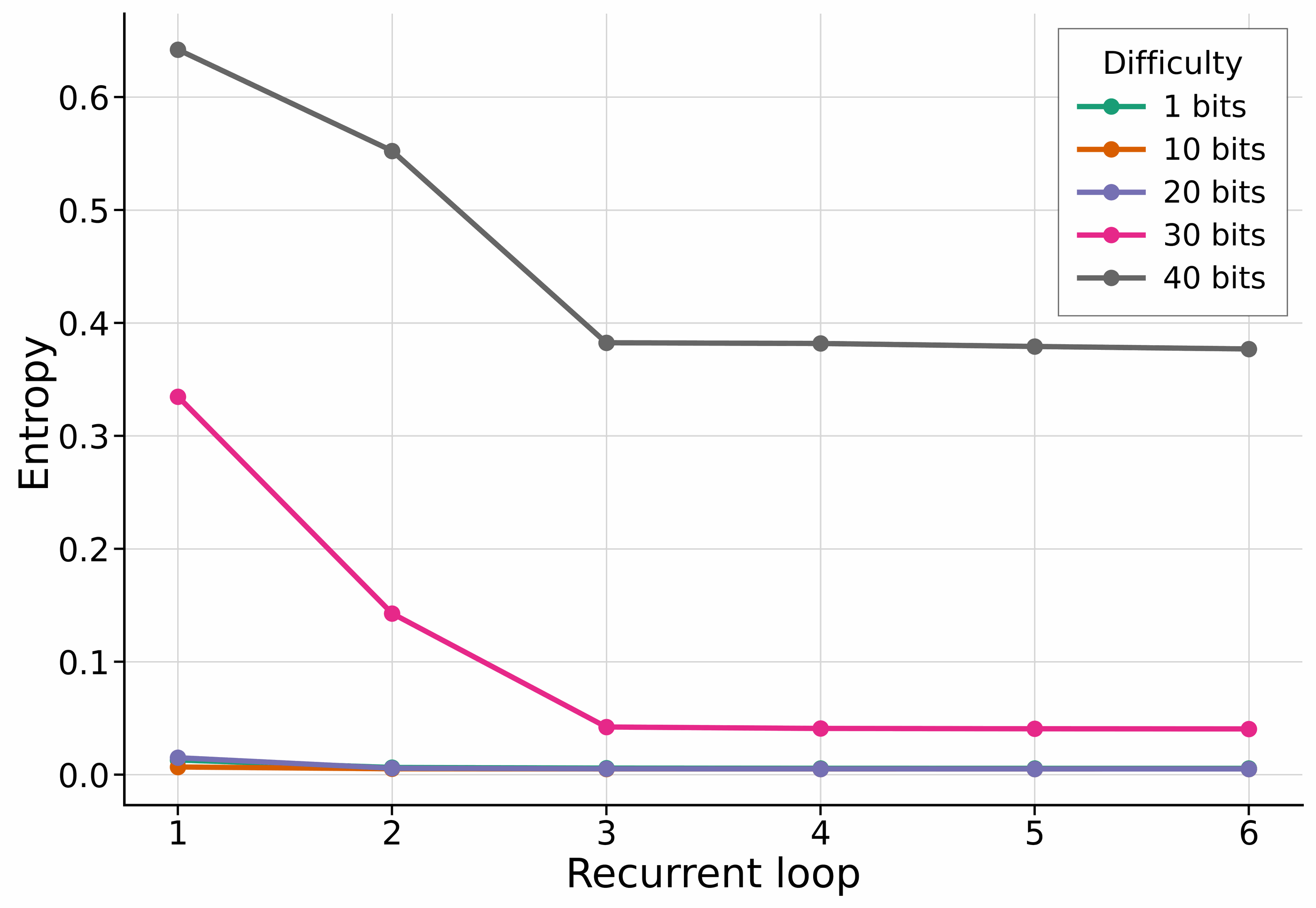}
    \caption{Entropy.}
\end{subfigure}
\hfill
\begin{subfigure}[t]{0.32\linewidth}
    \centering
    \includegraphics[
        width=\linewidth,
        trim={0.10in 0.05in 0.10in 0.08in},
        clip
    ]{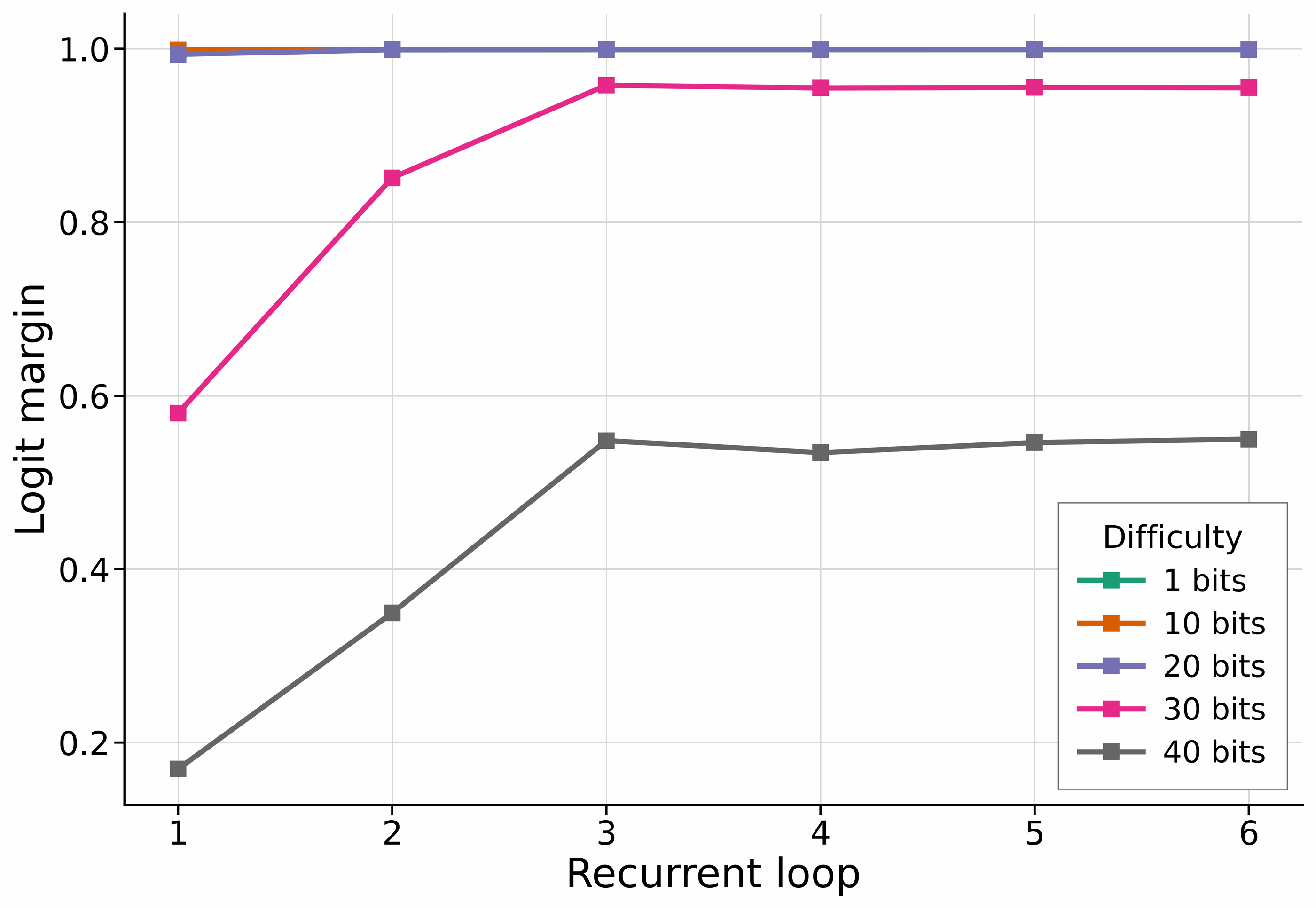}
    \caption{Logit margin.}
\end{subfigure}
\hfill
\begin{subfigure}[t]{0.32\linewidth}
    \centering
    \includegraphics[
        width=\linewidth,
        trim={0.10in 0.05in 0.10in 0.08in},
        clip
    ]{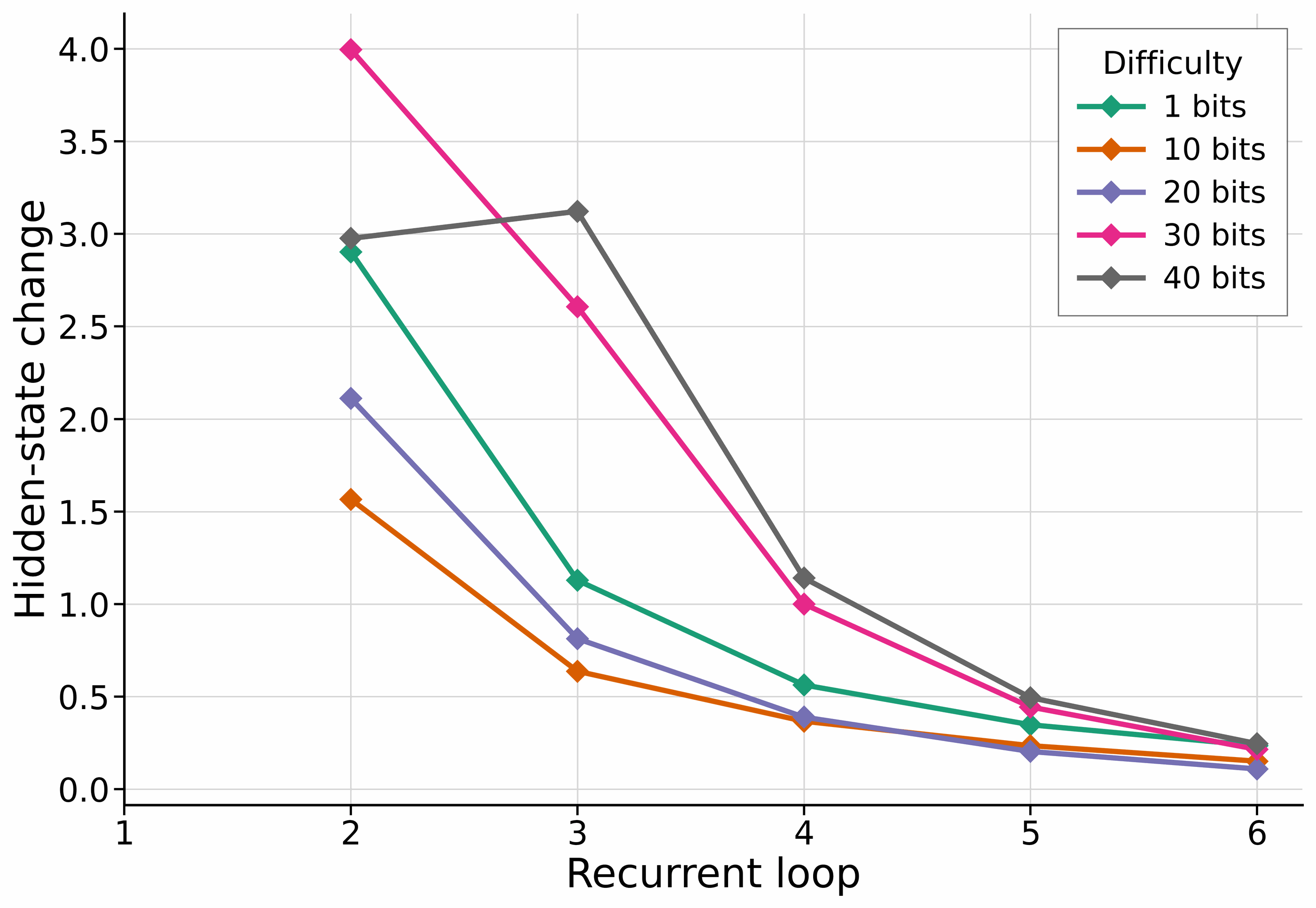}
    \caption{Hidden-state displacement.}
\end{subfigure}

\caption{
Difficulty-aware trajectory diagnostics on parity.
The geometric fixed-prior model is trained on bit strings with lengths from 1 to 40.
Curves are grouped by bit length.
Longer sequences remain uncertain for more recurrent loops, develop logit margin more slowly, and show larger hidden-state changes across depth.
}
\label{fig:parity_difficulty_diagnostics}
\end{figure}

Figure~\ref{fig:parity_difficulty_diagnostics} shows the corresponding difficulty-aware trajectory diagnostics. We use bit length as a simple difficulty variable. Although parity is not an arithmetic expression task, longer inputs require aggregating information across more positions. We therefore group validation and test examples by bit length and inspect trajectory statistics across recurrent depth. The fixed-prior trajectories show the same qualitative structure as in MANO. Longer sequences remain higher-entropy for more recurrent loops and develop logit margin more slowly. This suggests that the recurrent trajectory reflects input difficulty even in a non-MANO synthetic task.

\section{Additional Ouro Results}
\label{app:ouro_pareto}

This appendix reports the full per-benchmark readout comparison for the Ouro checkpoints, expanding the two representative curves in Section~\ref{sec:large-scale}. For each model and benchmark we plot benchmark accuracy against the average loops, comparing the pretrained ponder gate with confidence and convergence readouts applied to the same recurrent trajectories. We do not train or modify the models, so the trajectory is fixed by pretraining and only the readout varies.

\begin{figure}[htbp]
    \centering
    \begin{subfigure}{0.32\linewidth}
        \centering
        \includegraphics[width=\linewidth]{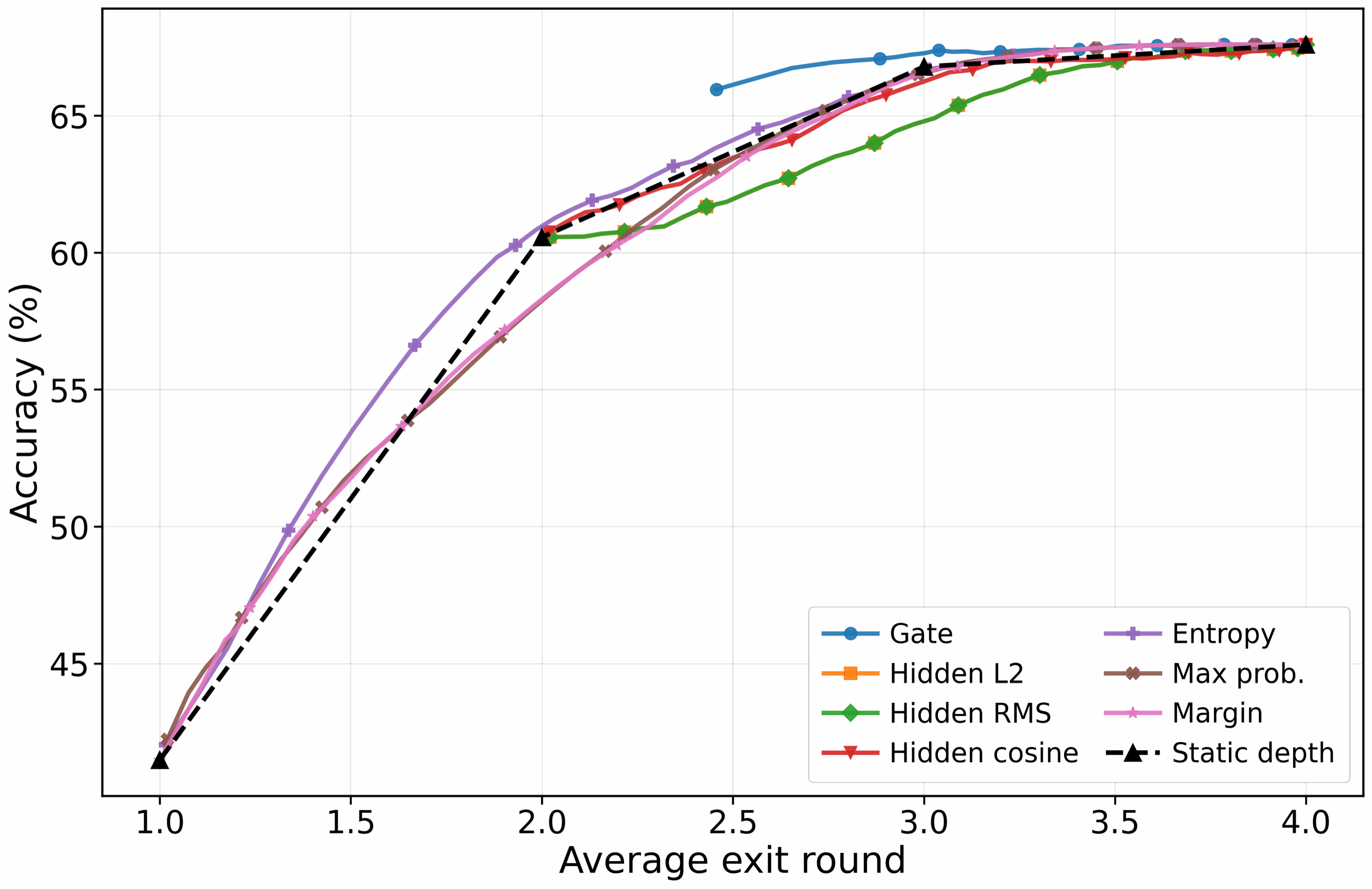}
        \caption{MMLU}
        \label{fig:app_ouro_14b_mmlu}
    \end{subfigure}
    \hfill
    \begin{subfigure}{0.32\linewidth}
        \centering
        \includegraphics[width=\linewidth]{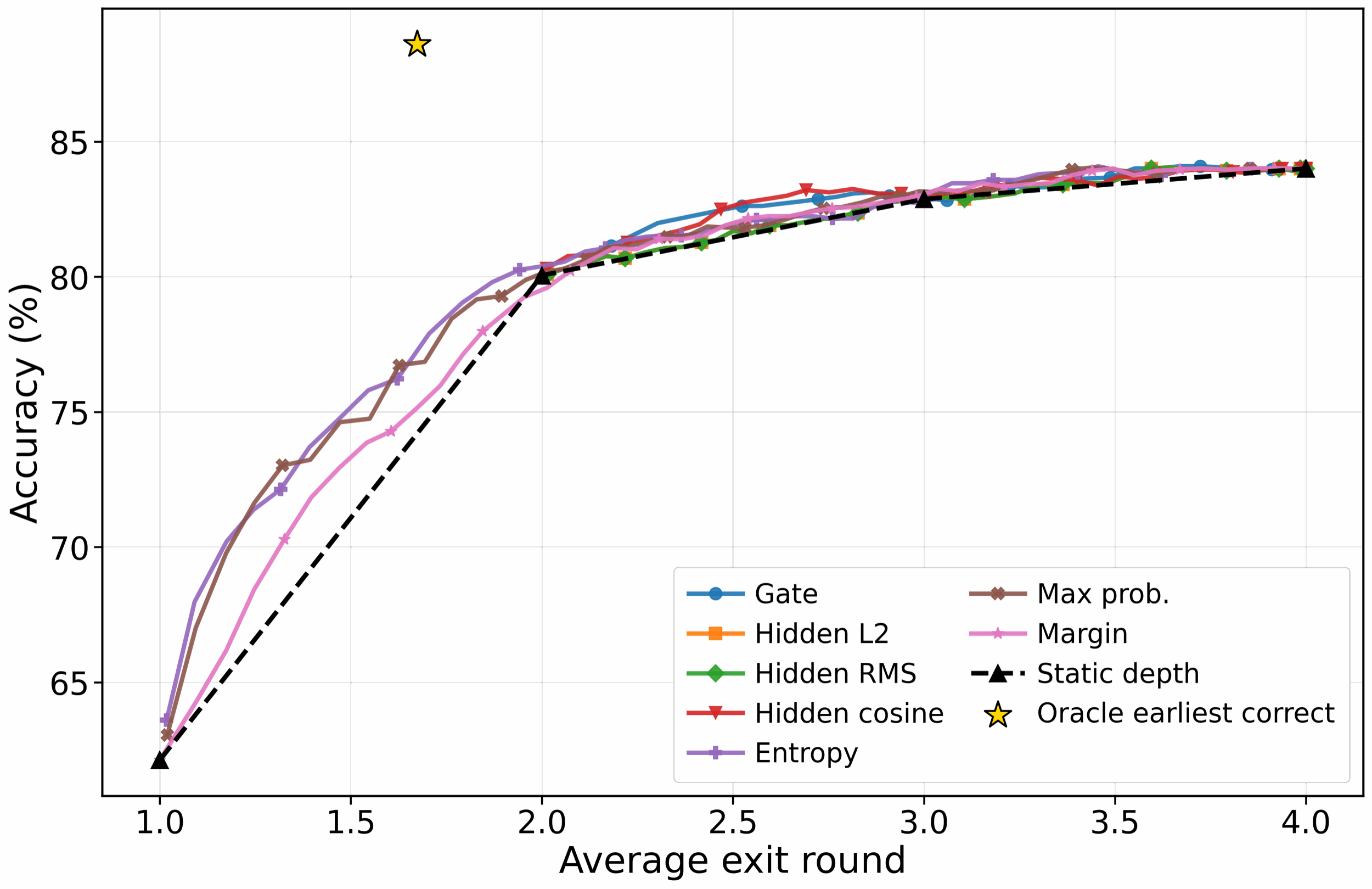}
        \caption{ARC-Easy}
        \label{fig:app_ouro_14b_arc_easy}
    \end{subfigure}
    \hfill
    \begin{subfigure}{0.32\linewidth}
        \centering
        \includegraphics[width=\linewidth]{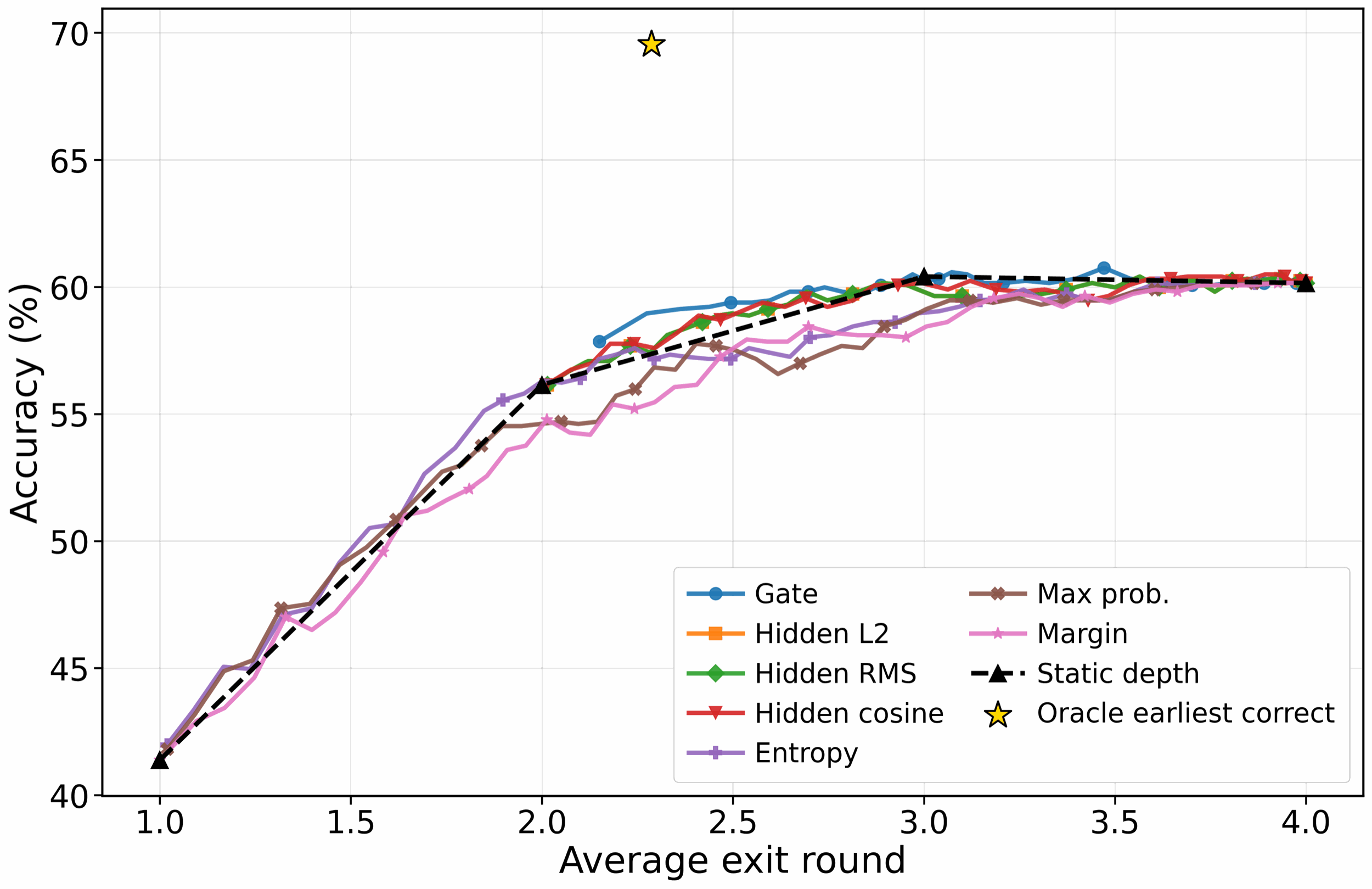}
        \caption{ARC-Challenge}
        \label{fig:app_ouro_14b_arc_challenge}
    \end{subfigure}

    \vspace{0.5em}

    \begin{subfigure}{0.32\linewidth}
        \centering
        \includegraphics[width=\linewidth]{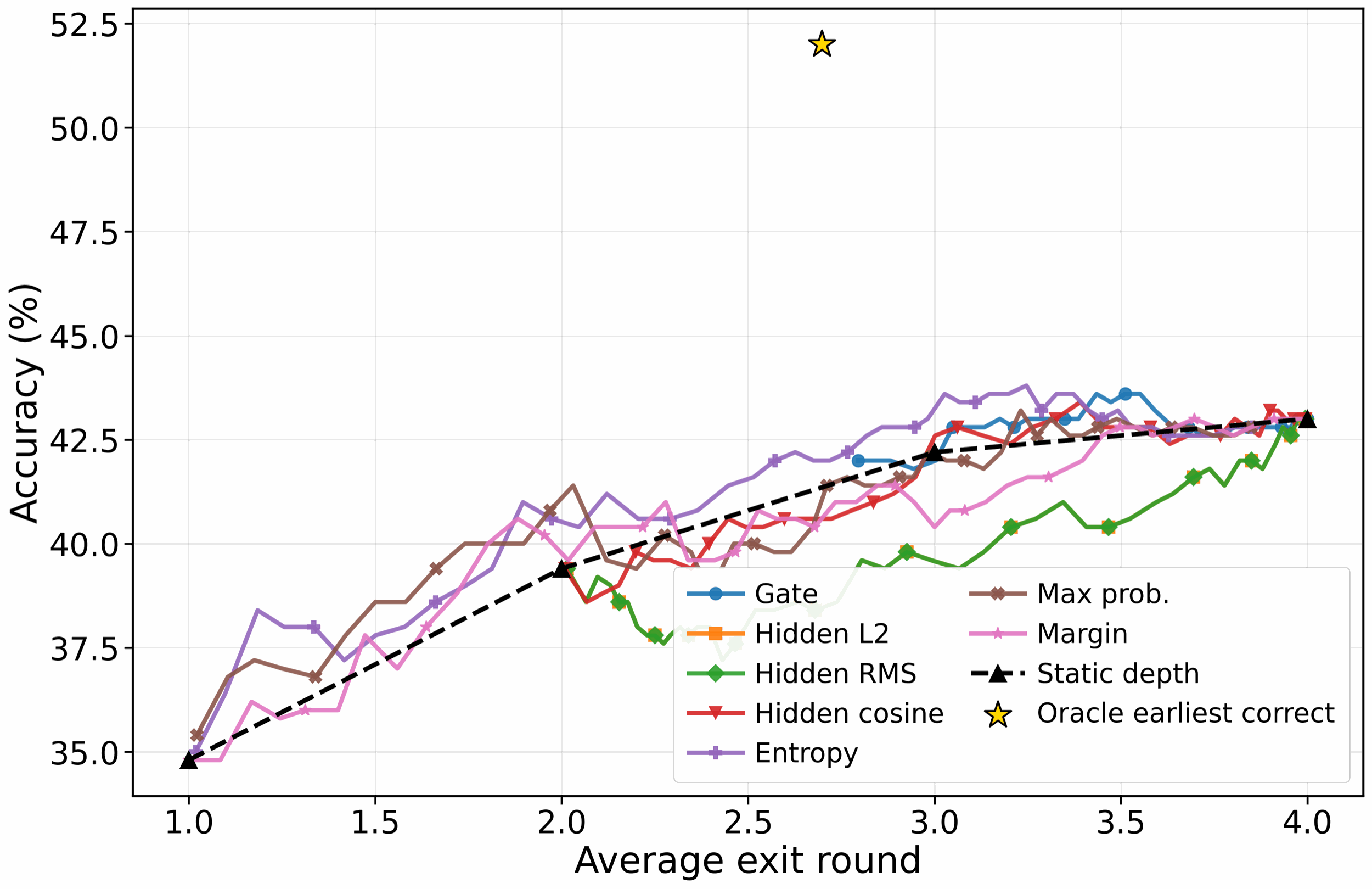}
        \caption{OpenBookQA}
        \label{fig:app_ouro_14b_openbookqa}
    \end{subfigure}
    \hfill
    \begin{subfigure}{0.32\linewidth}
        \centering
        \includegraphics[width=\linewidth]{images/ouro_pareto/1.4b/figure5_hellaswag_10shot_acc_norm.png}
        \caption{HellaSwag}
        \label{fig:app_ouro_14b_hellaswag}
    \end{subfigure}
    \hfill
    \begin{subfigure}{0.32\linewidth}
        \centering
        \includegraphics[width=\linewidth]{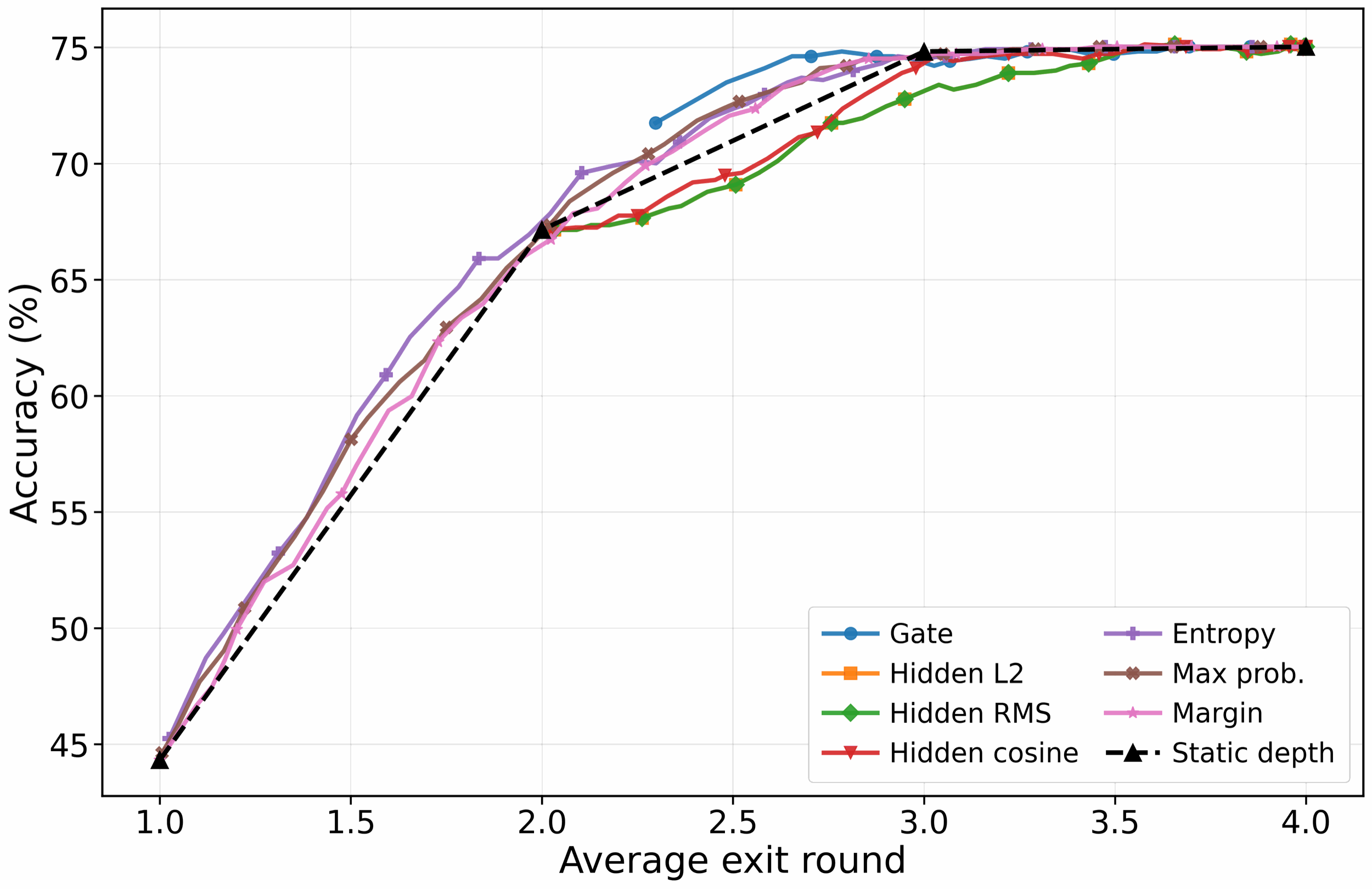}
        \caption{CommonsenseQA}
        \label{fig:app_ouro_14b_commonsenseqa}
    \end{subfigure}

    \caption{
    Full readout comparison for Ouro-1.4B.
    Each panel plots benchmark accuracy against average loops.
    The pretrained ponder gate is compared with confidence and convergence readouts applied to the same recurrent trajectories.
    }
    \label{fig:app_ouro_14b_all}
\end{figure}

\begin{figure}[htbp]
    \centering
    \begin{subfigure}{0.32\linewidth}
        \centering
        \includegraphics[width=\linewidth]{images/ouro_pareto/2.6b/figure5_ByteDance_Ouro-2.6B_mmlu_5shot_acc.png}
        \caption{MMLU}
        \label{fig:app_ouro_26b_mmlu}
    \end{subfigure}
    \hfill
    \begin{subfigure}{0.32\linewidth}
        \centering
        \includegraphics[width=\linewidth]{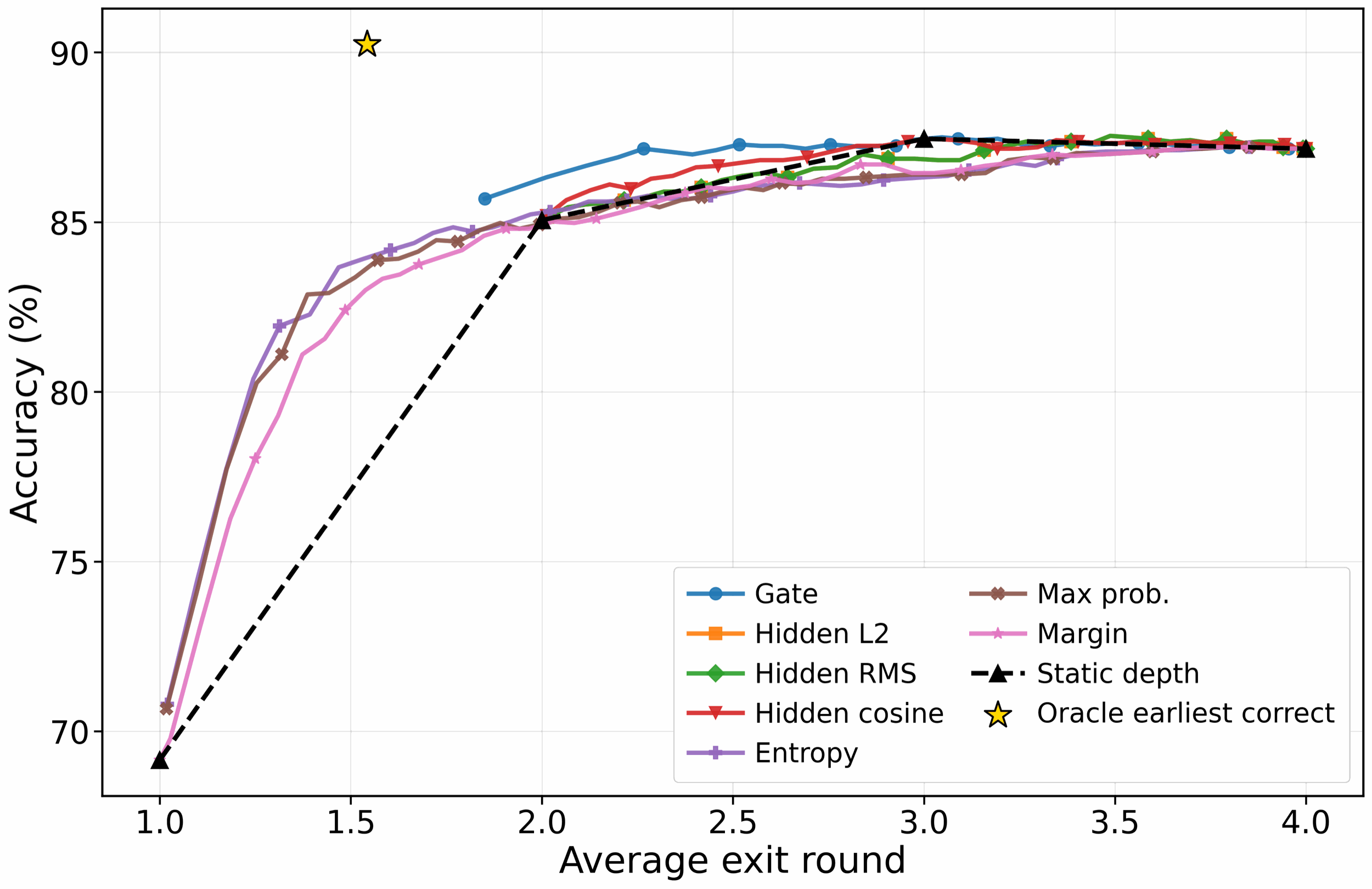}
        \caption{ARC-Easy}
        \label{fig:app_ouro_26b_arc_easy}
    \end{subfigure}
    \hfill
    \begin{subfigure}{0.32\linewidth}
        \centering
        \includegraphics[width=\linewidth]{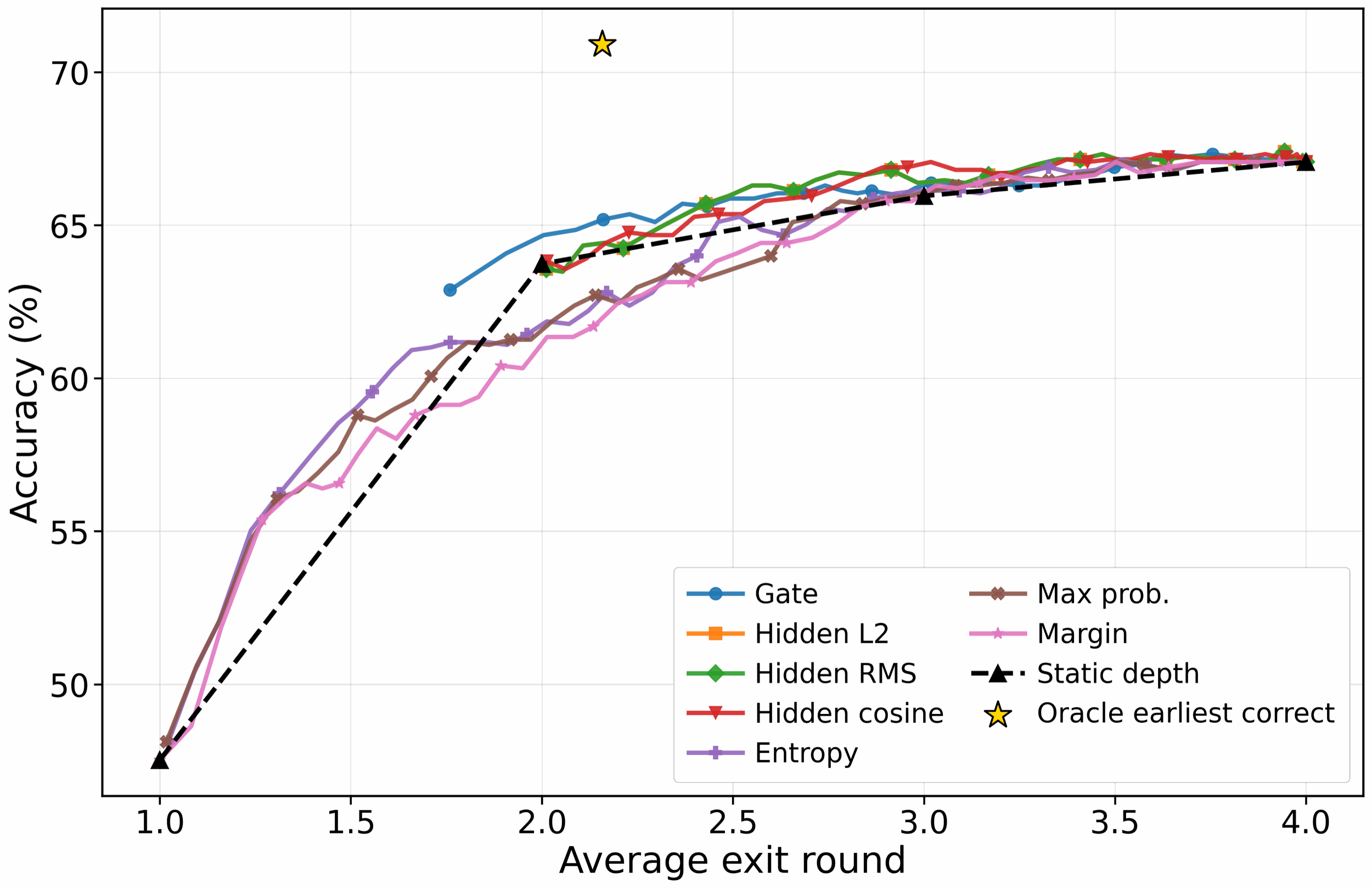}
        \caption{ARC-Challenge}
        \label{fig:app_ouro_26b_arc_challenge}
    \end{subfigure}

    \vspace{0.5em}

    \begin{subfigure}{0.32\linewidth}
        \centering
        \includegraphics[width=\linewidth]{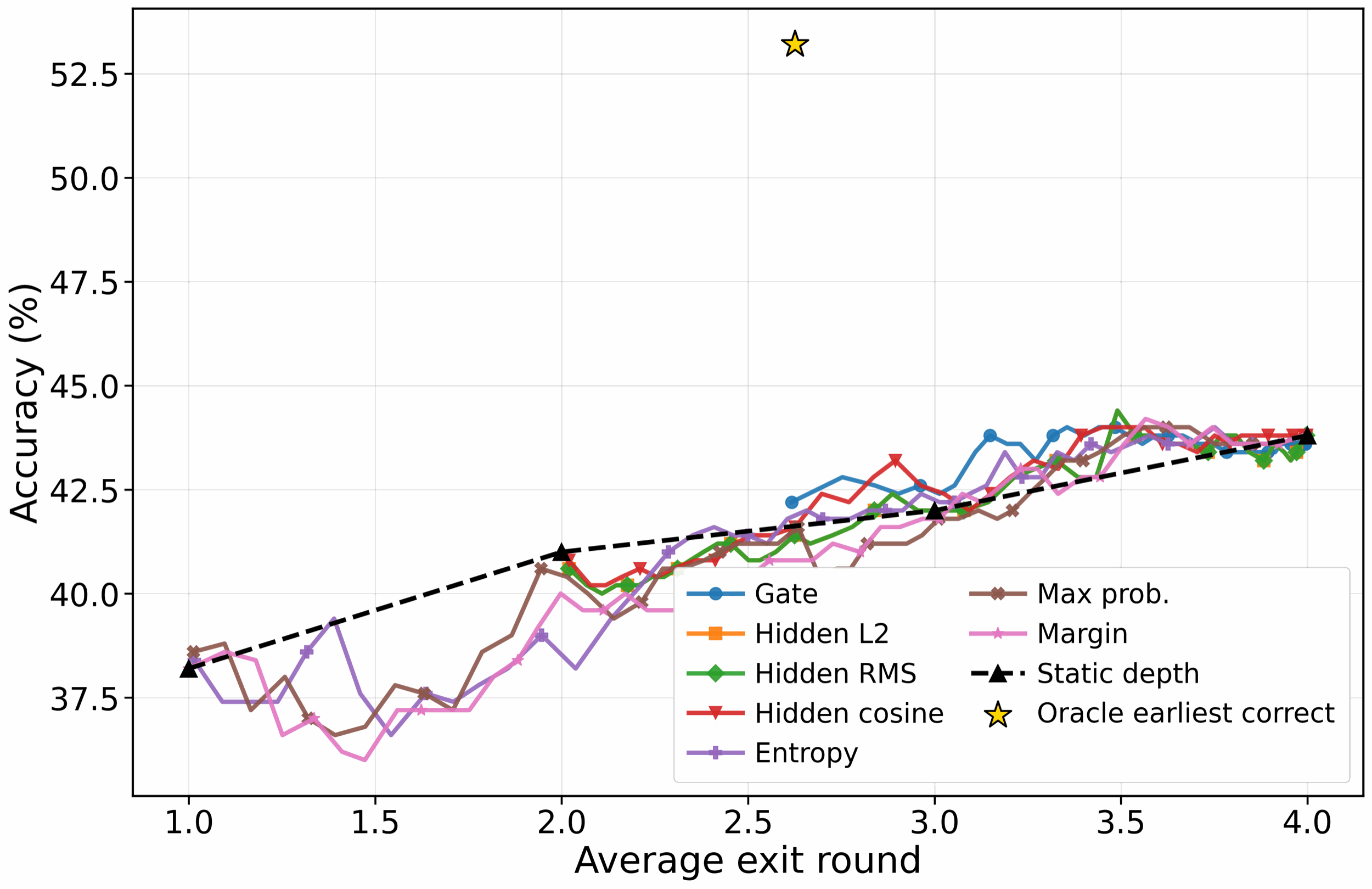}
        \caption{OpenBookQA}
        \label{fig:app_ouro_26b_openbookqa}
    \end{subfigure}
    \hfill
    \begin{subfigure}{0.32\linewidth}
        \centering
        \includegraphics[width=\linewidth]{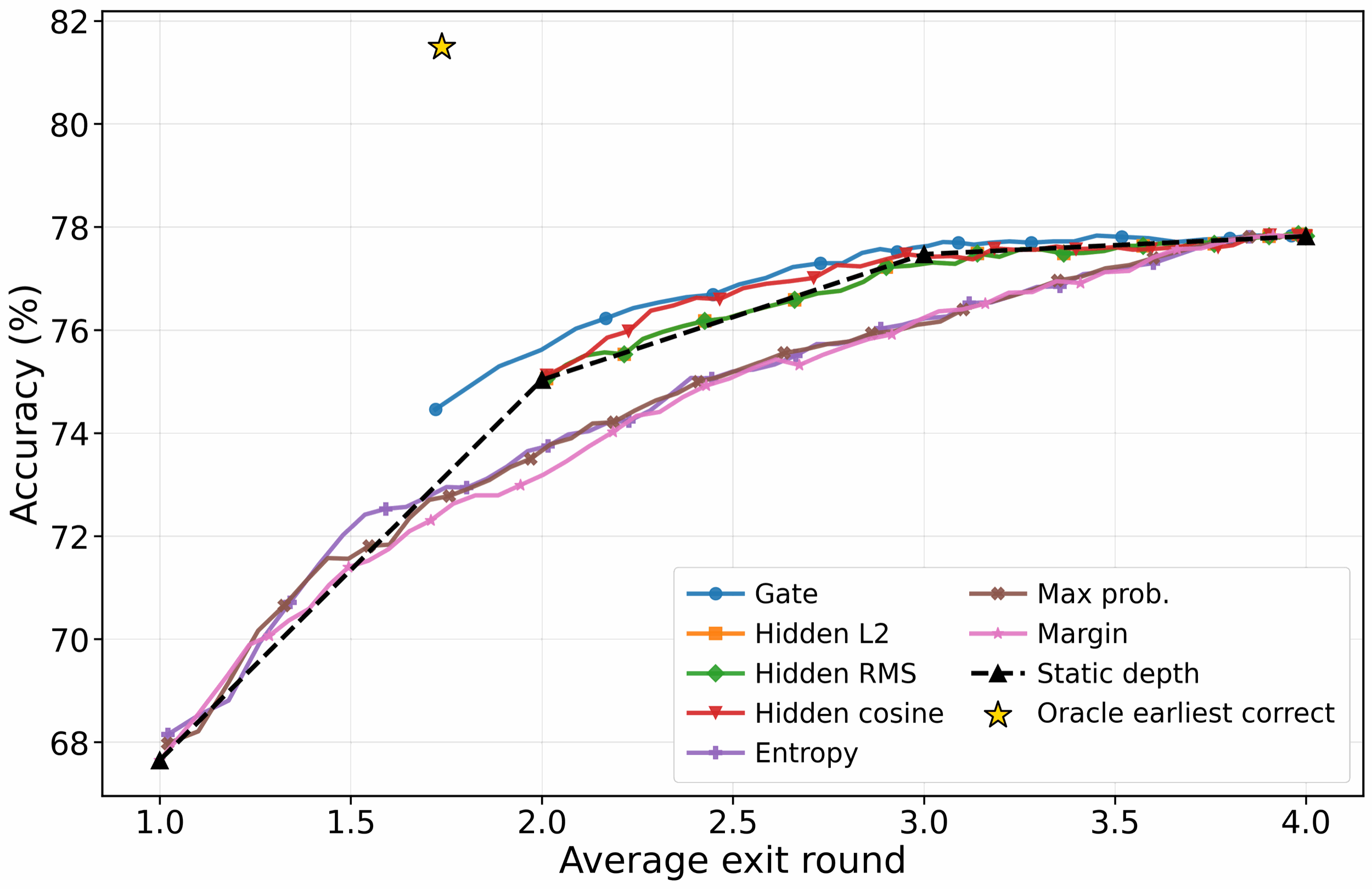}
        \caption{HellaSwag}
        \label{fig:app_ouro_26b_hellaswag}
    \end{subfigure}
    \hfill
    \begin{subfigure}{0.32\linewidth}
        \centering
        \includegraphics[width=\linewidth]{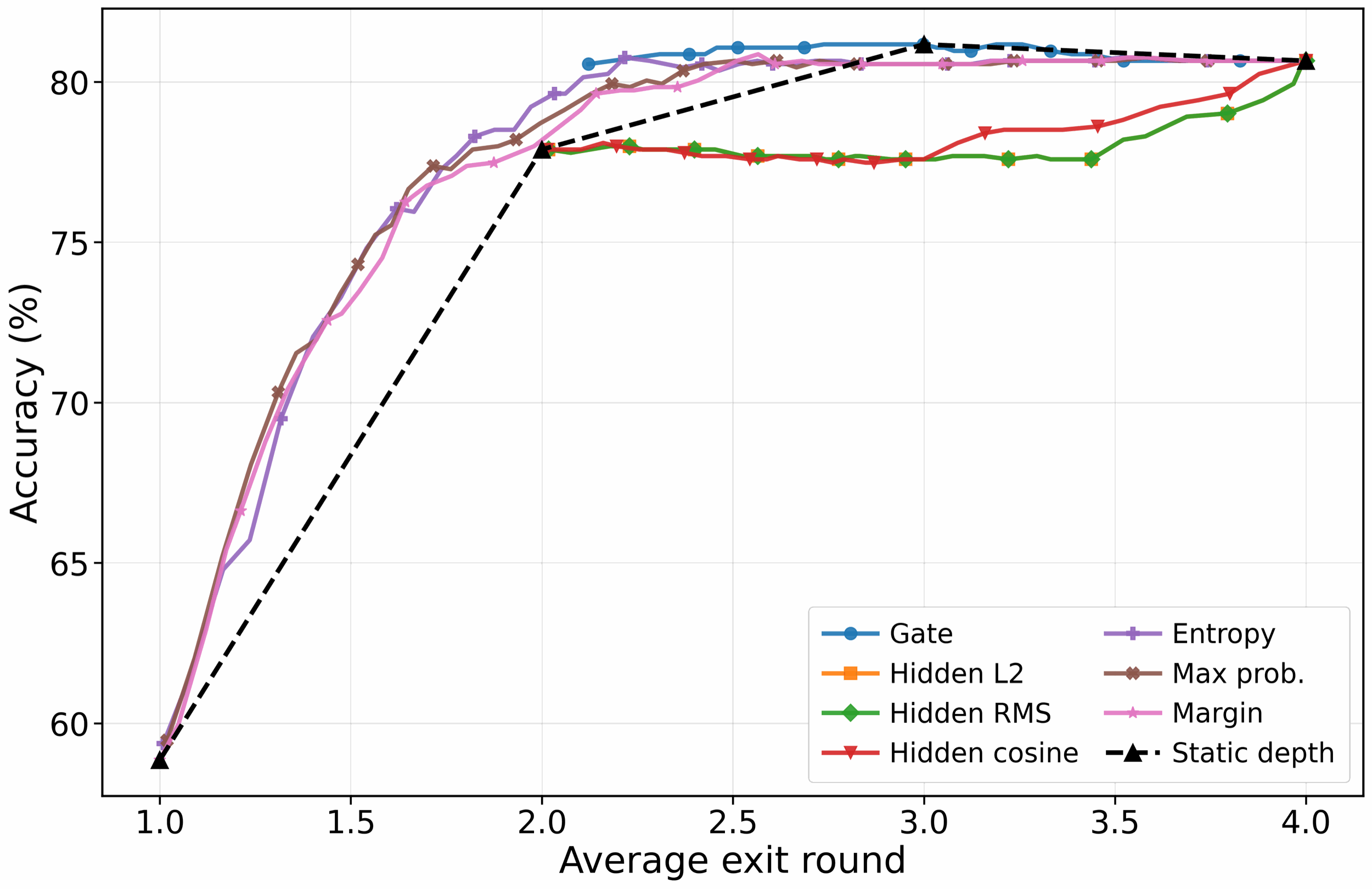}
        \caption{CommonsenseQA}
        \label{fig:app_ouro_26b_commonsenseqa}
    \end{subfigure}

    \caption{
    Full readout comparison for Ouro-2.6B.
    Each panel plots benchmark accuracy against average loops.
    The pretrained ponder gate is compared with confidence and convergence readouts applied to the same recurrent trajectories.
    }
    \label{fig:app_ouro_26b_all}
\end{figure}

Figure~\ref{fig:app_ouro_14b_all} shows all six benchmarks for Ouro-1.4B, and Figure~\ref{fig:app_ouro_26b_all} shows the corresponding curves for Ouro-2.6B. Across both model sizes, the post-hoc readouts track the pretrained gate closely and in several benchmarks match or improve on its compute--accuracy tradeoff. The pretrained gate remains a competitive operating point and is never far from the frontier, but it is not uniformly Pareto-optimal. The best readout varies across benchmark and model size, and simple confidence or convergence criteria such as maximum probability, entropy, logit margin, and hidden-state cosine similarity recover comparable or better tradeoffs in many cases. The static-depth baseline included in each panel exits at the same loop number for every example and is generally dominated by the adaptive readouts, confirming that the gains come from input-dependent stopping rather than from a uniform reduction in depth. These full grids support the same conclusion as the numerical summary in Table~\ref{tab:ouro_readout_summary}. The recurrent trajectory exposes stopping signals that are not fully captured by the pretrained gate.

The oracle earliest-correct marker shown in several panels indicates the accuracy ceiling achievable if one could always exit at the first correct loop, and the gap between it and the realized readouts indicates the remaining room for improvement in stopping rules.

\section{Frozen Trajectory Gate Fitting}
\label{app:posthoc-gates}

This appendix expands the frozen-trajectory experiments of Section~\ref{sec:posthoc-gate-fitting}. The goal is to separate two possible causes of weak learned-gate performance: an underexpressive or underfit gate readout, versus a trajectory shaped poorly by joint gate training. We freeze a pretrained recurrent trajectory, including the backbone and LM head, and train a new exit gate on top of the frozen hidden states. Because the trajectory is fixed, the new gate can only act as a readout. It cannot change the recurrent states or the per-depth predictions. We train both linear and MLP post-hoc gates, under two objectives: the PonderNet-style objective used for the native gates, and a marginal-utility objective inspired by Ouro Stage-II training that predicts, for each adjacent pair of depths, whether continuing one more loop is worthwhile.

\begingroup
\captionsetup[subfigure]{font=scriptsize,skip=1pt}

\begin{figure}[htbp]
\centering
\begin{subfigure}[t]{0.42\linewidth}
    \centering
    \includegraphics[width=\linewidth,height=1.5in,keepaspectratio]{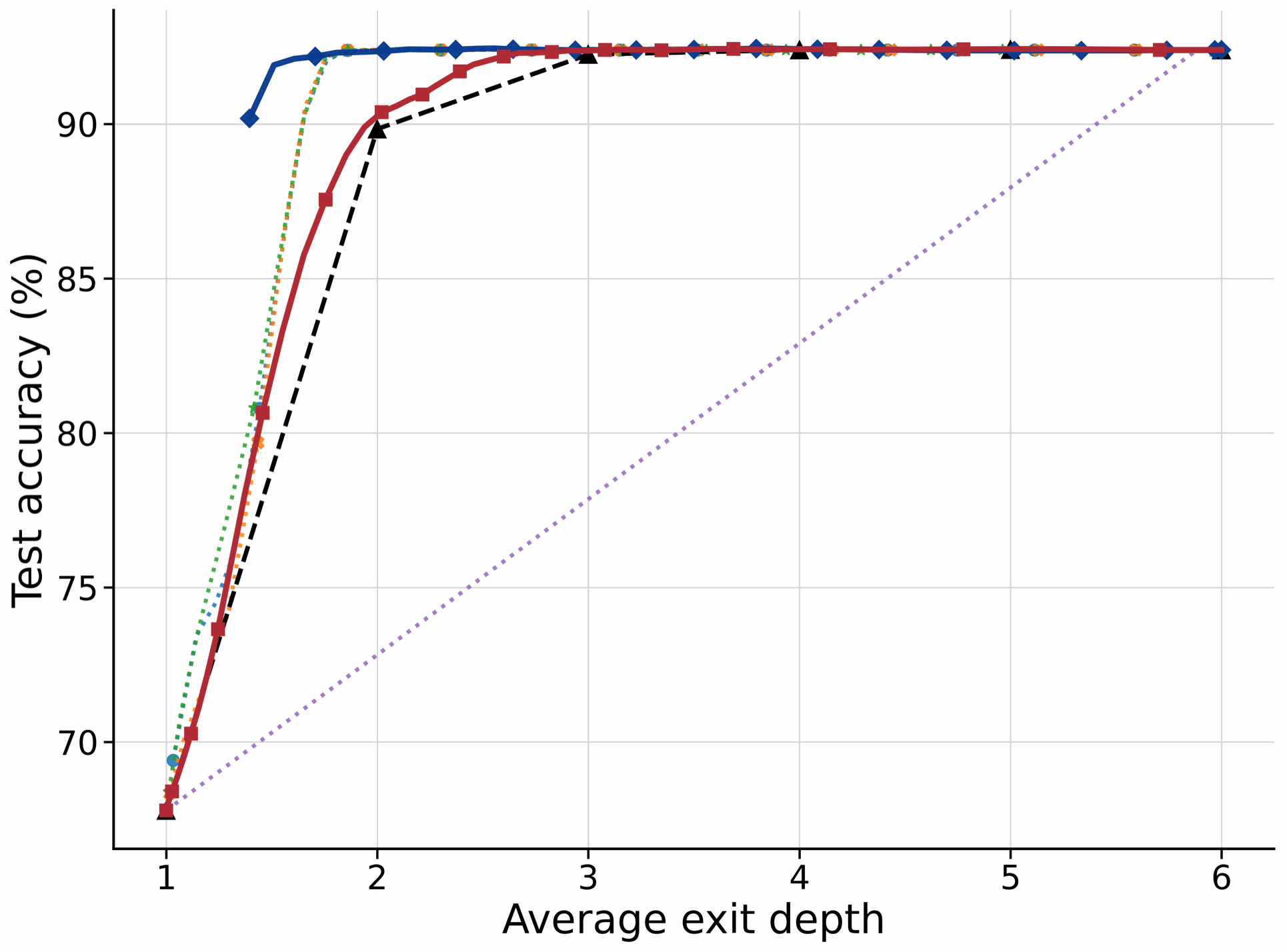}
    \caption{Uniform fixed prior, linear post-hoc gate.}
\end{subfigure}
\hfill
\begin{subfigure}[t]{0.544\linewidth}
    \centering
    \includegraphics[width=\linewidth,height=1.5in,keepaspectratio]{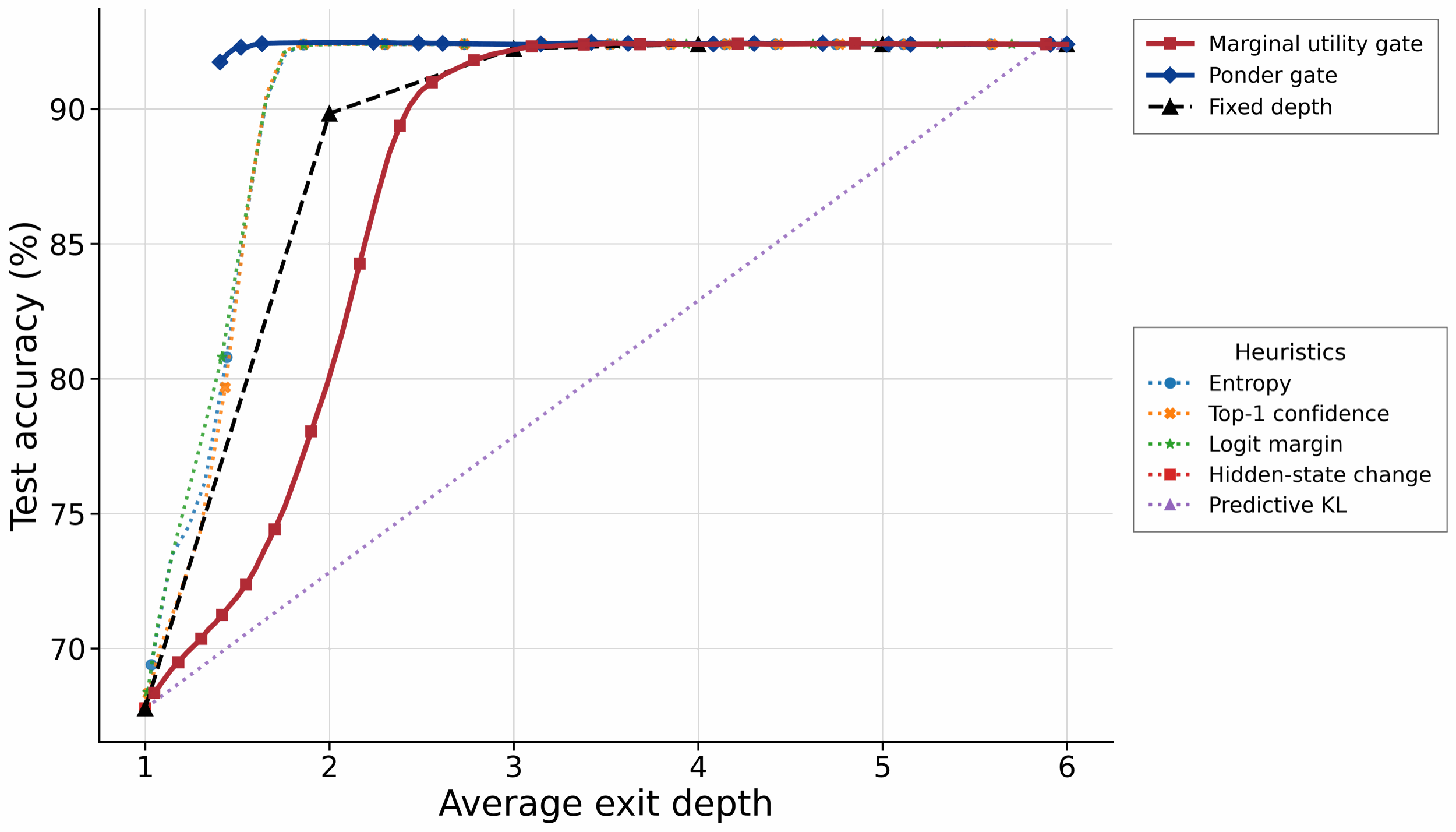}
    \caption{Uniform fixed prior, MLP post-hoc gate.}
\end{subfigure}

\vspace{0.3em}
\begin{subfigure}[t]{0.42\linewidth}
    \centering
    \includegraphics[width=\linewidth,height=1.5in,keepaspectratio]{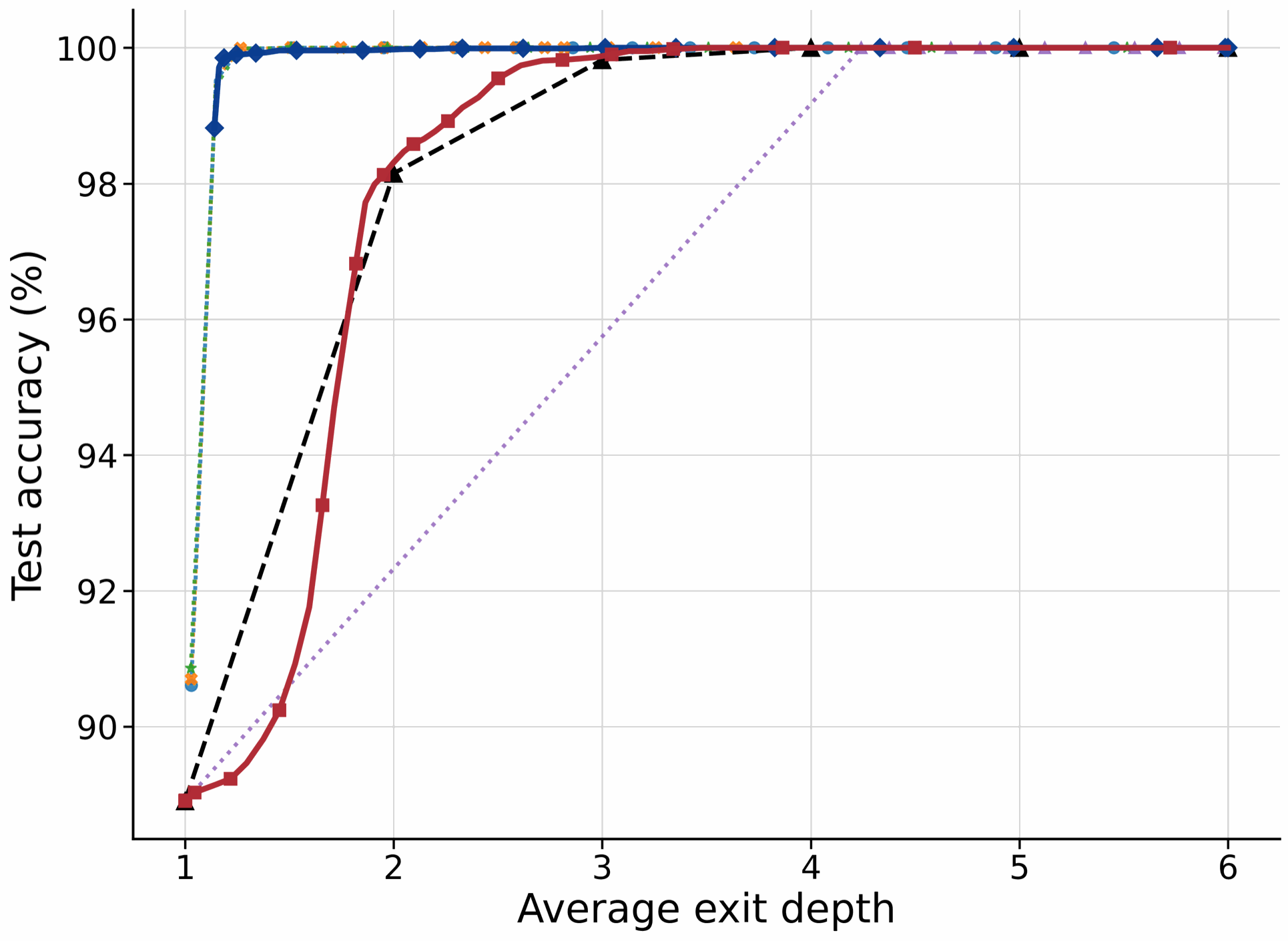}
    \caption{Geometric fixed prior, $\lambda=0.3$, linear post-hoc gate.}
\end{subfigure}
\hfill
\begin{subfigure}[t]{0.544\linewidth}
    \centering
    \includegraphics[width=\linewidth,height=1.5in,keepaspectratio]{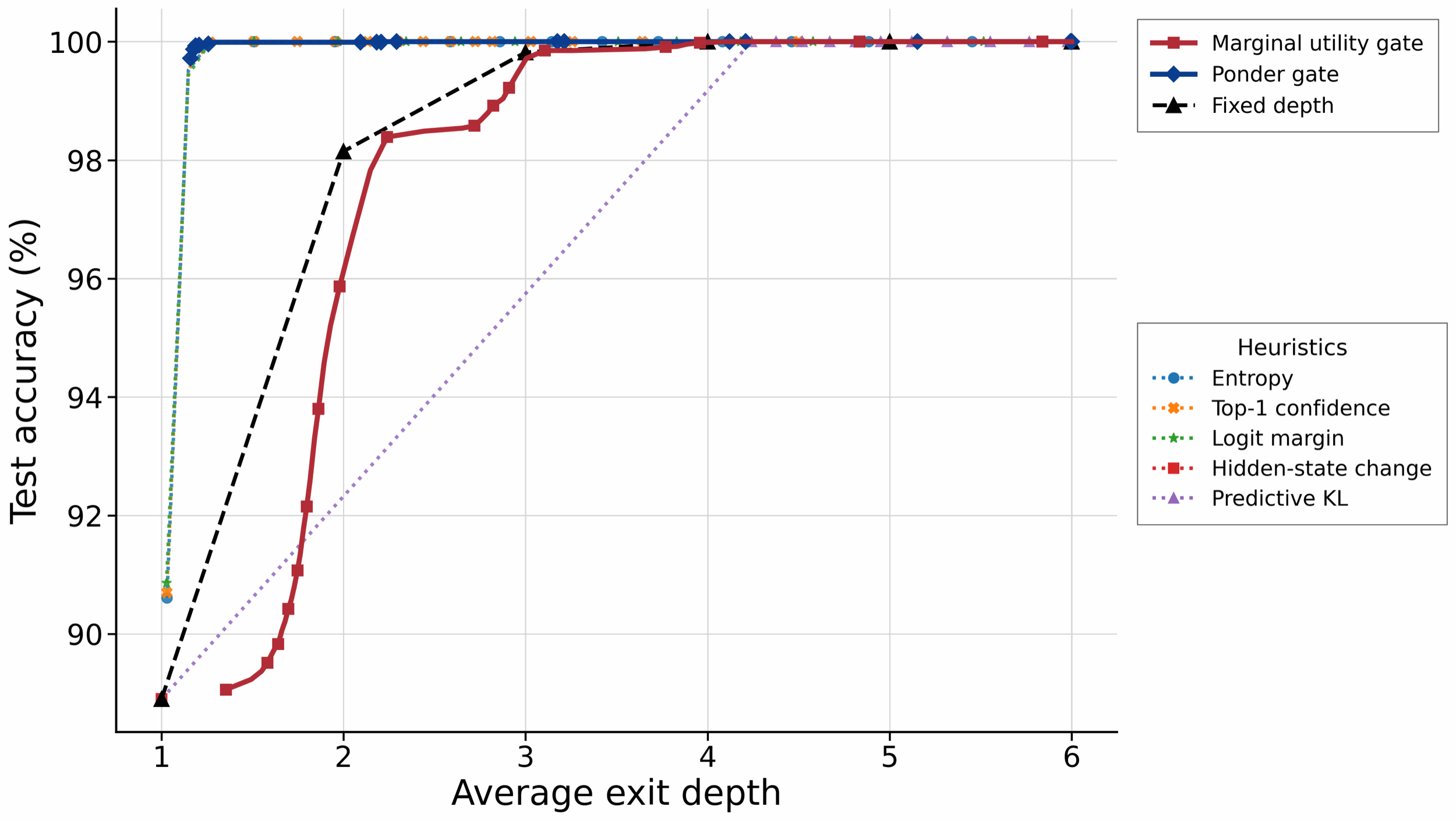}
    \caption{Geometric fixed prior, $\lambda=0.3$, MLP post-hoc gate.}
\end{subfigure}
\caption{
Post-hoc gate training on frozen fixed-prior MANO trajectories.
Both linear and MLP post-hoc gates are trained while keeping the recurrent backbone and LM head fixed.
}
\label{fig:posthoc_gate_fixed_appendix}
\end{figure}

\begin{figure}[htbp]
\centering
\begin{subfigure}[t]{0.42\linewidth}
    \centering
    \includegraphics[width=\linewidth,height=1.5in,keepaspectratio]{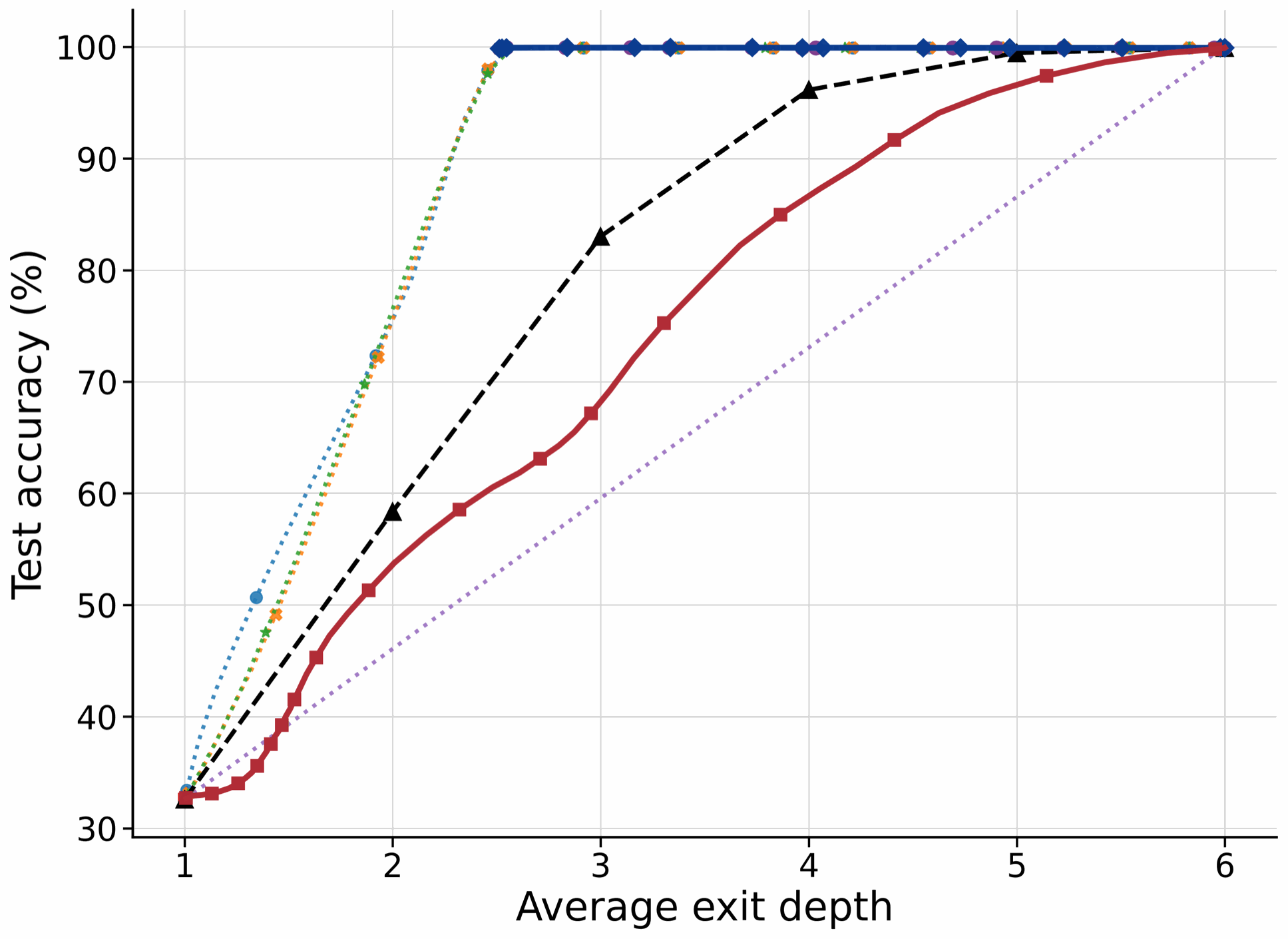}
    \caption{Uniform linear-gate trajectory, $\beta=0.1$, linear post-hoc gate.}
\end{subfigure}
\hfill
\begin{subfigure}[t]{0.545\linewidth}
    \centering
    \includegraphics[width=\linewidth,height=1.5in,keepaspectratio]{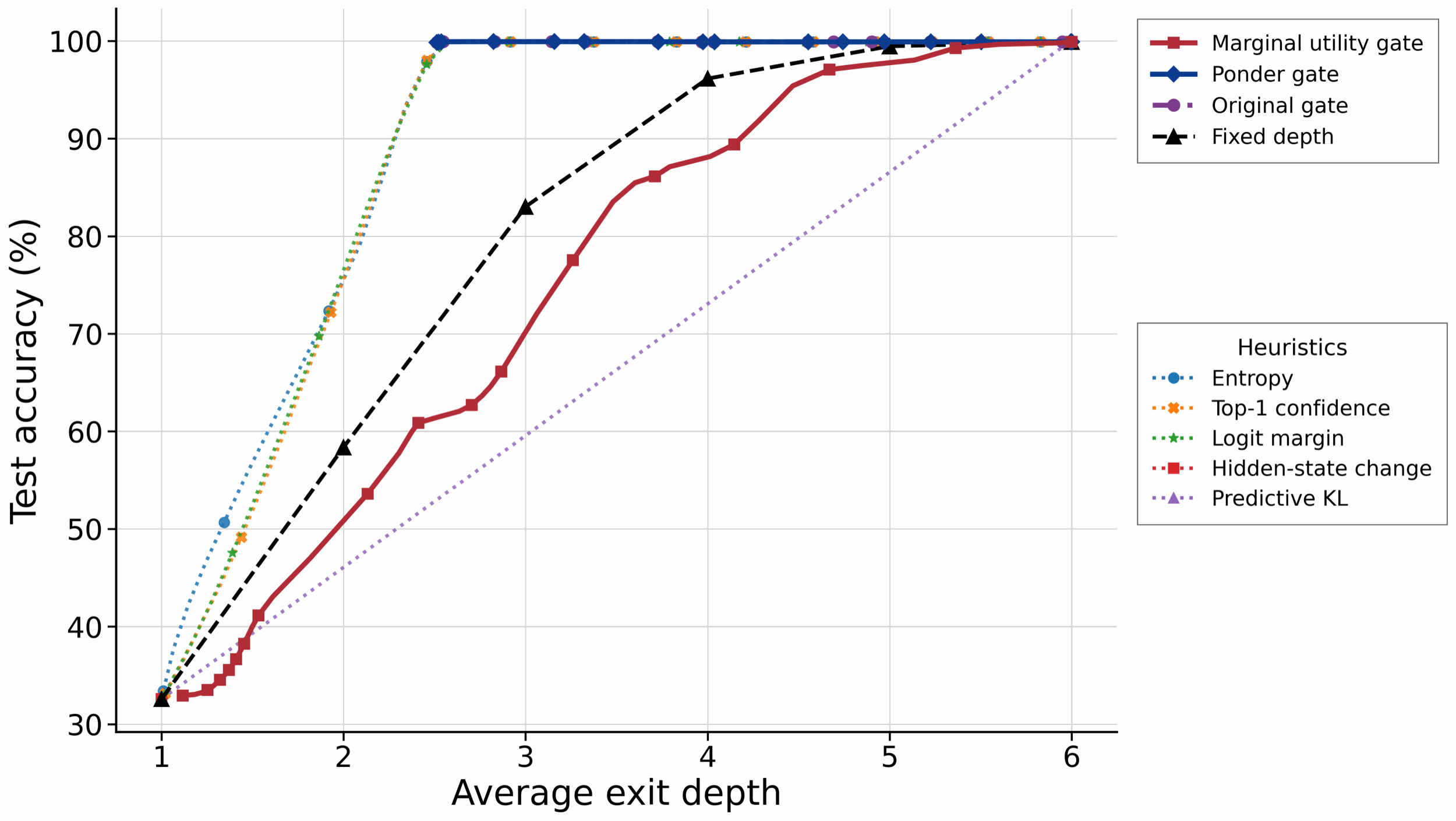}
    \caption{Uniform linear-gate trajectory, $\beta=0.1$, MLP post-hoc gate.}
\end{subfigure}

\vspace{0.3em}
\begin{subfigure}[t]{0.42\linewidth}
    \centering
    \includegraphics[width=\linewidth,height=1.5in,keepaspectratio]{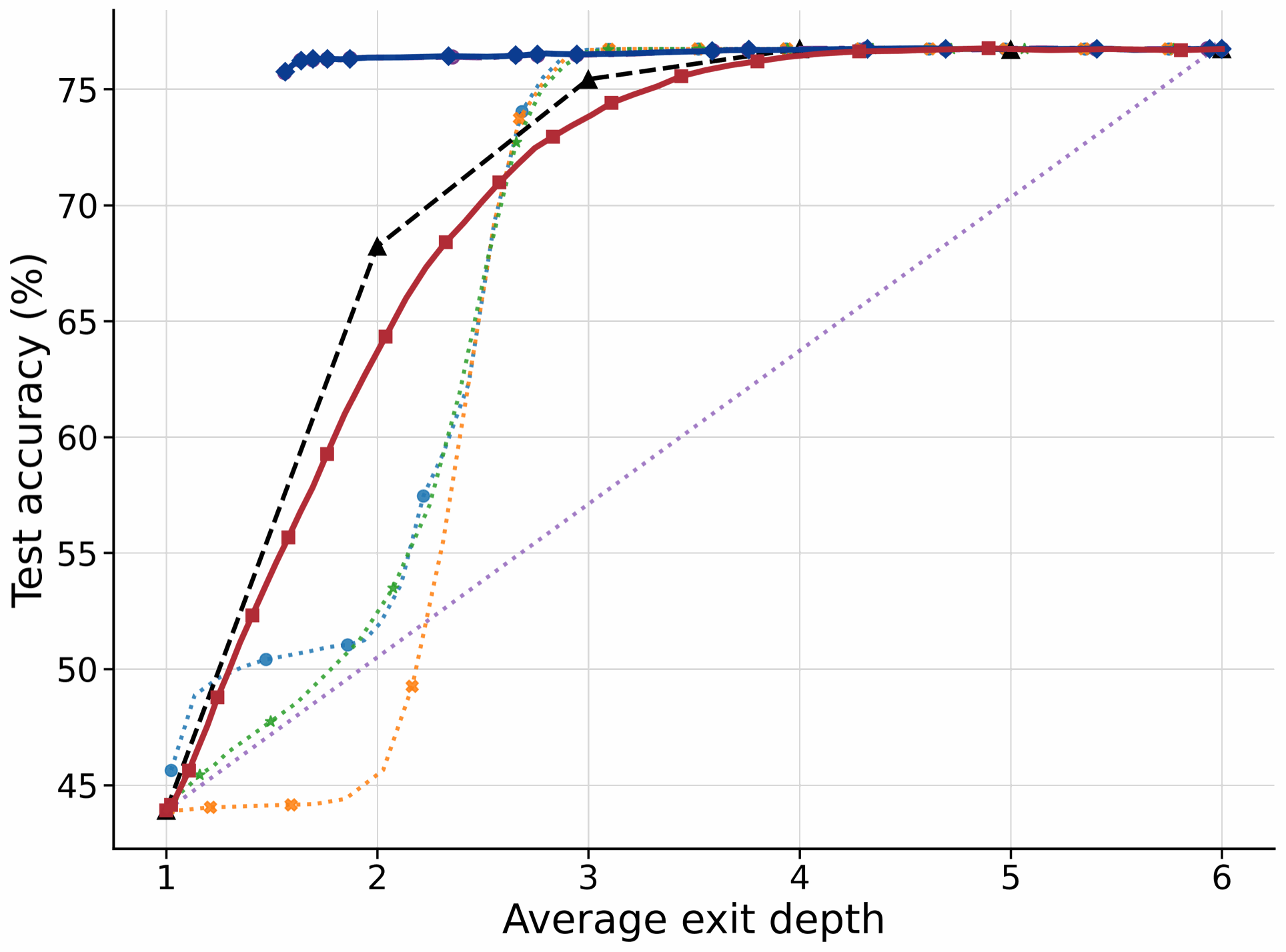}
    \caption{Geometric linear-gate trajectory, $\lambda=0.3$, $\beta=0.1$, linear post-hoc gate.}
\end{subfigure}
\hfill
\begin{subfigure}[t]{0.545\linewidth}
    \centering
    \includegraphics[width=\linewidth,height=1.5in,keepaspectratio]{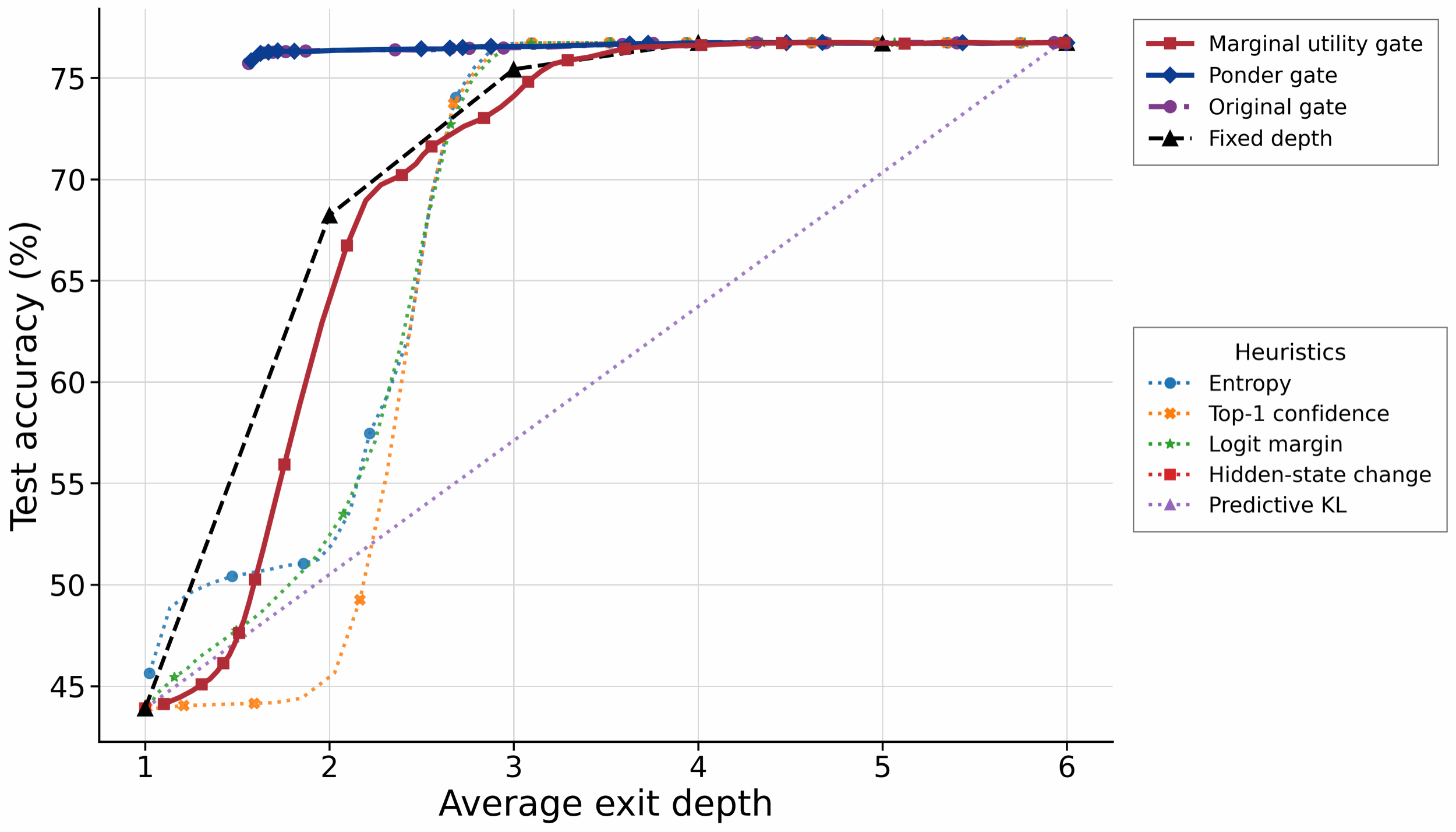}
    \caption{Geometric linear-gate trajectory, $\lambda=0.3$, $\beta=0.1$, MLP post-hoc gate.}
\end{subfigure}
\caption{
Post-hoc gate training on frozen MANO trajectories originally trained with a linear learned gate.
The native learned gate is evaluated when available, and newly trained post-hoc gates are compared against heuristic readouts and fixed-depth exits.
}
\label{fig:posthoc_gate_linear_learned_appendix}
\end{figure}

\begin{figure}[htbp]
\centering
\begin{subfigure}[t]{0.42\linewidth}
    \centering
    \includegraphics[width=\linewidth,height=1.5in,keepaspectratio]{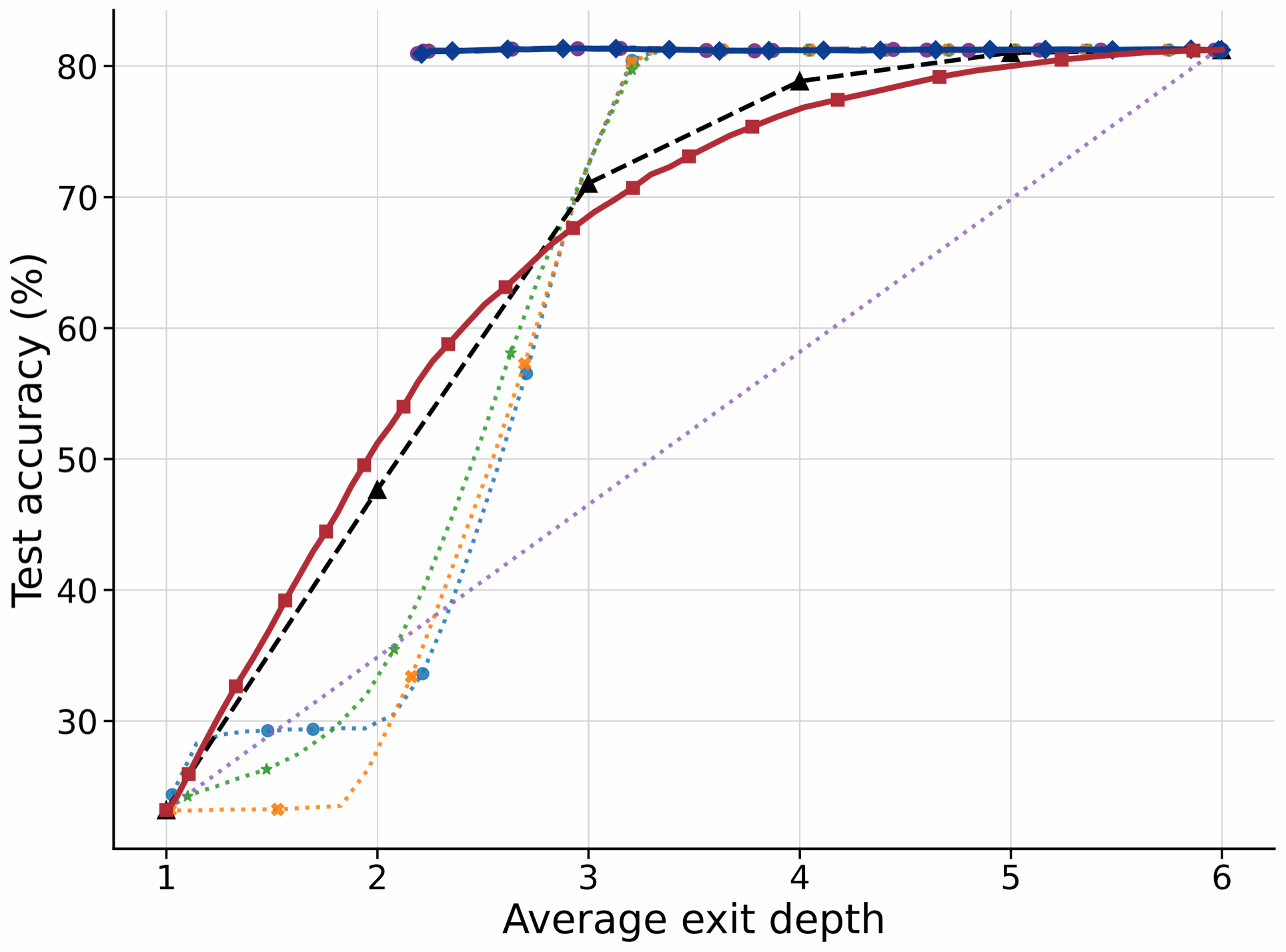}
    \caption{Uniform MLP-gate trajectory, $\beta=0.1$, linear post-hoc gate.}
\end{subfigure}
\hfill
\begin{subfigure}[t]{0.545\linewidth}
    \centering
    \includegraphics[width=\linewidth,height=1.5in,keepaspectratio]{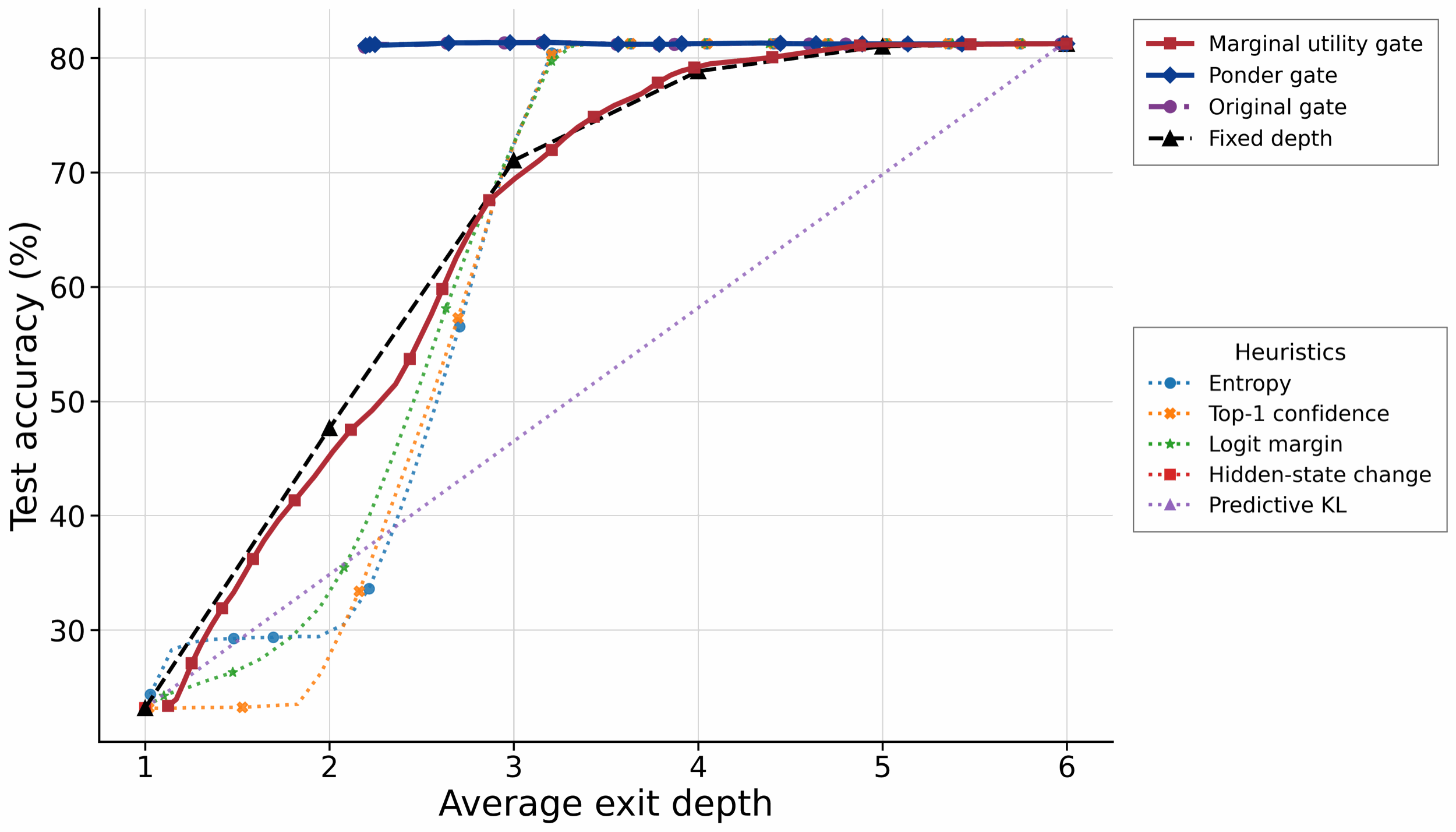}
    \caption{Uniform MLP-gate trajectory, $\beta=0.1$, MLP post-hoc gate.}
\end{subfigure}

\vspace{0.3em}
\begin{subfigure}[t]{0.42\linewidth}
    \centering
    \includegraphics[width=\linewidth,height=1.5in,keepaspectratio]{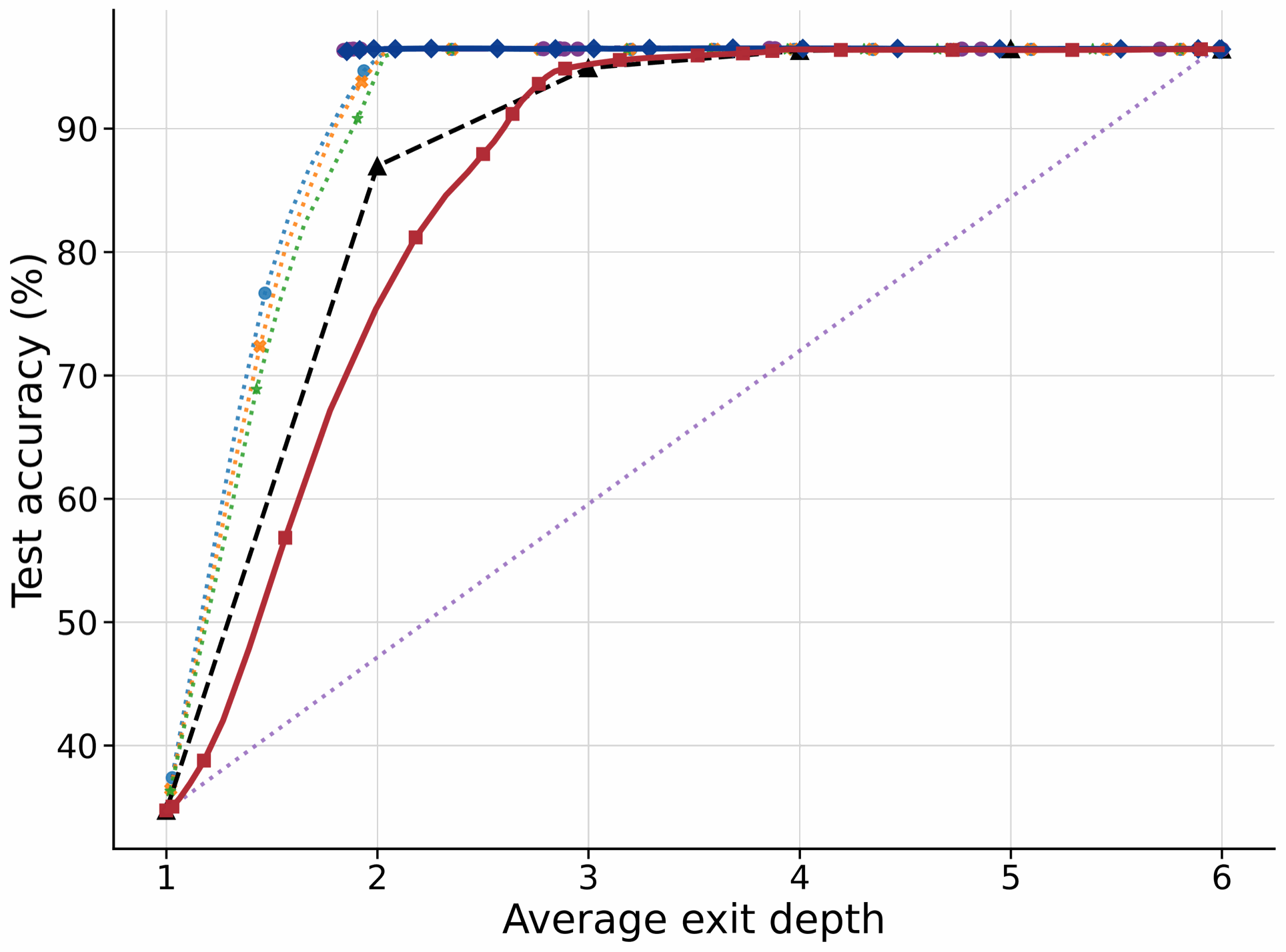}
    \caption{Geometric MLP-gate trajectory, $\lambda=0.3$, $\beta=0.1$, linear post-hoc gate.}
\end{subfigure}
\hfill
\begin{subfigure}[t]{0.545\linewidth}
    \centering
    \includegraphics[width=\linewidth,height=1.5in,keepaspectratio]{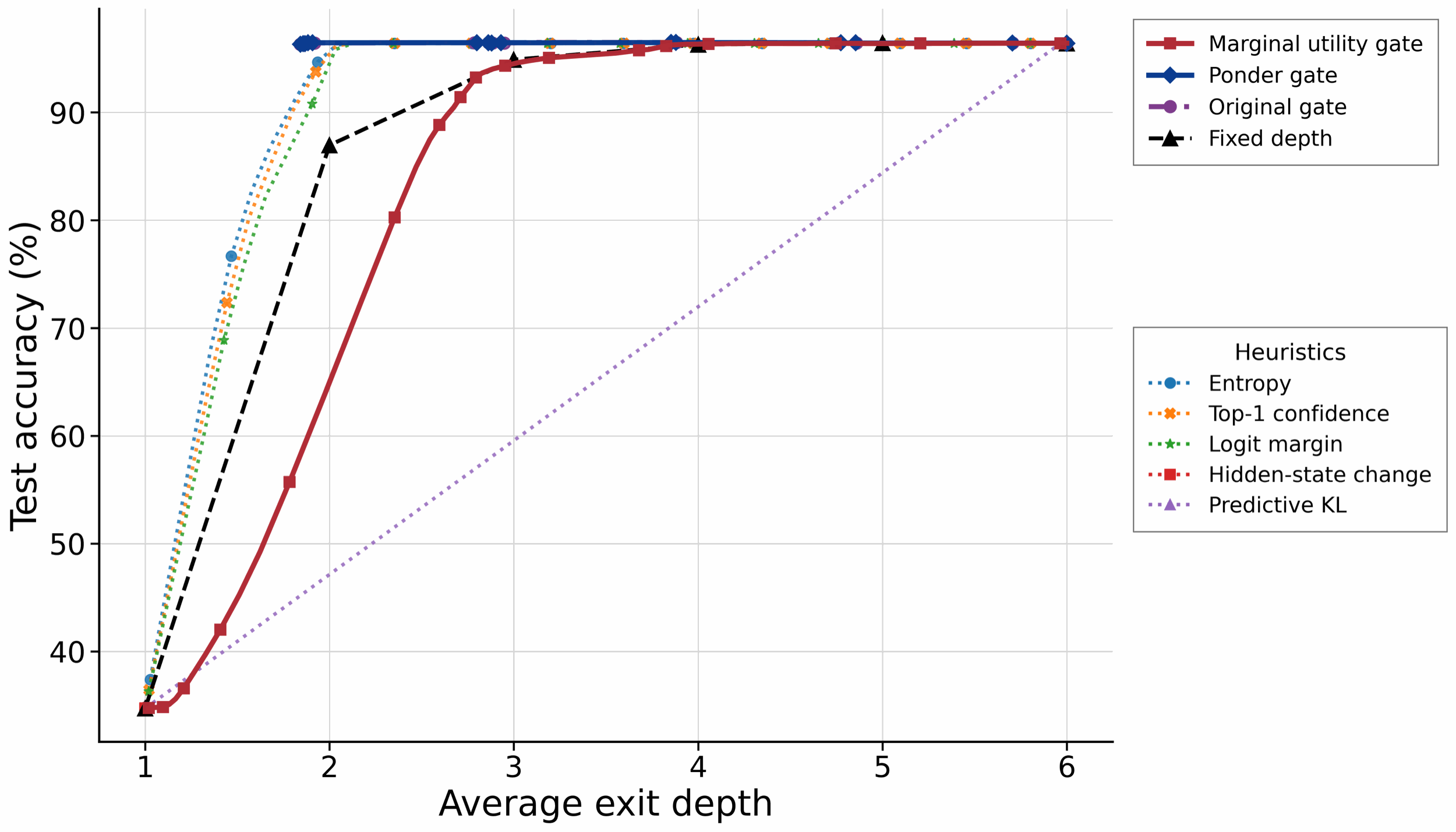}
    \caption{Geometric MLP-gate trajectory, $\lambda=0.3$, $\beta=0.1$, MLP post-hoc gate.}
\end{subfigure}
\caption{
Post-hoc gate training on frozen MANO trajectories originally trained with an MLP learned gate.
Across both uniform and geometric priors, post-hoc gates test whether a newly trained readout can improve on the gate learned jointly with the trajectory.
}
\label{fig:posthoc_gate_mlp_learned_appendix}
\end{figure}

\endgroup

Figure~\ref{fig:posthoc_gate_fixed_appendix} shows post-hoc gates trained on frozen fixed-prior trajectories. Here the post-hoc gates recover strong early-exit behavior, reaching near-final-loop accuracy at much lower average depth, which demonstrates that the gate class is capable of representing useful stopping policies when trajectory formation is separated from readout learning. The simple confidence readouts remain competitive with the fitted gates, indicating that much of the stopping signal is already exposed in the prediction trajectory and does not require a learned readout to extract.

Figures~\ref{fig:posthoc_gate_linear_learned_appendix} and~\ref{fig:posthoc_gate_mlp_learned_appendix} show post-hoc gates trained on trajectories that were themselves produced by joint learned-gate training, for the linear and MLP native gates respectively. As in Section~\ref{sec:posthoc-gate-fitting}, fitting a new post-hoc gate on these trajectories does not improve the frontier: the newly trained gates largely track the native gate and remain bounded by the quality of the underlying trajectory regardless of gate class, and the PonderNet-style objective again outperforms the marginal-utility objective, which exits later without improving the frontier.

Taken together, these frozen-trajectory results confirm the diagnosis from Section~\ref{sec:posthoc-gate-fitting} across both gate classes: the limiting factor is not gate expressivity or post-hoc fitting, since the same gate class recovers strong frontiers on fixed-prior trajectories, but the trajectory produced by joint training itself.

\section{Reasoning-Tuned Ouro}
\label{app:ouro_thinking}

The Ouro paper reports that post-SFT RLVR attempts did not bring significant gains, and identifies the dynamic early-exit mechanism as a central difficulty for RL training. Fast rollout systems assume a fixed execution path, but Ouro uses dynamic depth computation. This makes the gate an important interface between latent recurrent computation and optimization. We therefore evaluate whether trajectory readouts can recover competitive compute--quality tradeoffs on the reasoning-tuned Ouro-Thinking checkpoints.

We evaluate Ouro-1.4B-Thinking and Ouro-2.6B-Thinking on the same six benchmarks used in our main Ouro experiments. This keeps the task distribution and scoring protocol fixed, allowing us to isolate whether reasoning tuning changes the readout behavior of recurrent trajectories. Table~\ref{tab:ouro_thinking_diagnostic} reports the best validation-selected points for each model and dataset. The same qualitative pattern persists in this experiment. The pretrained gate is a useful readout, but it is not necessarily the best. In 7 of the 12 settings, the best readout is a simple post-hoc heuristic rather than the pretrained gate. Thus, reasoning tuning does not remove the main readout effect. The pretrained gate remains useful, but in many settings a simple confidence or convergence readout gives the best performance-compute tradeoff. This suggests that even reasoning-specialized looped models expose stopping information outside the gate.

\begin{table}[htbp]
\centering
\caption{
Ouro-Thinking checkpoint diagnostic.
We report the best validation-selected operating points on the held-out evaluation split.
}
\label{tab:ouro_thinking_diagnostic}
\begin{tabular}{lllrr}
\toprule
Model & Dataset & Readout & Accuracy (\%) & Avg. loops \\
\midrule
Ouro-1.4B-Thinking & ARC-C & Hidden L2 & 61.9 & 3.31 \\
Ouro-1.4B-Thinking & ARC-E & Gate & 84.4 & 2.57 \\
Ouro-1.4B-Thinking & CSQA & Entropy & 73.9 & 2.78 \\
Ouro-1.4B-Thinking & HellaSwag & Hidden cosine & 68.9 & 2.23 \\
Ouro-1.4B-Thinking & MMLU & Max prob. & 67.1 & 3.20 \\
Ouro-1.4B-Thinking & OpenBookQA & Max prob. & 40.4 & 2.15 \\
\midrule
Ouro-2.6B-Thinking & ARC-C & Hidden L2 & 66.0 & 2.57 \\
Ouro-2.6B-Thinking & ARC-E & Gate & 87.6 & 2.28 \\
Ouro-2.6B-Thinking & CSQA & Gate & 79.5 & 2.30 \\
Ouro-2.6B-Thinking & HellaSwag & Gate & 73.7 & 2.95 \\
Ouro-2.6B-Thinking & MMLU & Max prob. & 75.7 & 2.89 \\
Ouro-2.6B-Thinking & OpenBookQA & Max prob. & 40.6 & 2.72 \\
\bottomrule
\end{tabular}
\end{table}

\end{document}

%% file: math_commands.tex
\usepackage{amsmath,amsfonts,bm}

\def\eqref#1{equation~\ref{#1}}

\def\1{\bm{1}}

\DeclareMathAlphabet{\mathsfit}{\encodingdefault}{\sfdefault}{m}{sl}
\SetMathAlphabet{\mathsfit}{bold}{\encodingdefault}{\sfdefault}{bx}{n}

